\pdfoutput=1

\documentclass[11pt]{article}
\usepackage{colortbl}

\usepackage[preprint]{acl}

\usepackage{times}
\usepackage{latexsym}
\usepackage{multirow} 
\usepackage{graphicx}
\usepackage[T1]{fontenc}

\usepackage[utf8]{inputenc}

\usepackage{microtype}
\usepackage{makecell}
\usepackage{booktabs}
\usepackage{subcaption}
\usepackage{bm}
\usepackage{inconsolata}

\PassOptionsToPackage{activate={true,nocompatibility},final,tracking,protrusion,expansion,factor=1100,stretch=10,shrink=10}{microtype}

\usepackage{bigdelim,dcolumn}  

\usepackage{amsmath}  
\usepackage{cleveref}  
\crefformat{section}{\S#2#1#3} 
\crefformat{subsection}{\S#2#1#3}
\crefformat{subsubsection}{\S#2#1#3}

\usepackage{enumitem}  
\setlist[1]{topsep=0.3em, itemsep=0.125em, parsep=0.05em}
\usepackage{relsize}  

\usepackage{nicematrix}

\usepackage[outline]{contour}
\usepackage{ulem}
\newcommand{\myul}[4]{%
  \begingroup%
  \renewcommand \ULdepth {#1}%
  \renewcommand \ULthickness {#2}%
  \contourlength{#3}%
  \uline{\phantom{#4}}%
  \llap{\contour{white}{#4}}%
  \endgroup%
}
\newcommand{\ul}[1]{\myul{2pt}{0.5pt}{1pt}{#1}}
\newcommand \myulc[5]{
  \begingroup%
  \renewcommand \ULdepth {#1}%
  \renewcommand \ULthickness {#2}%
  \contourlength{#3}%
  \textcolor{#5}{
    \uline{\textcolor{black}{\phantom{#4}}}
  }\llap{\contour{white}{#4}}
  \endgroup%
}
\newcommand{\ulc}[2]{\myulc{2pt}{0.5pt}{1pt}{#2}{#1}}
\usepackage{etoolbox}
\robustify{\ul}
\robustify{\ulc}

\definecolor{gGreen}{HTML}{6aa84f}
\definecolor{gGreenLight}{HTML}{d9ead3}
\definecolor{gRed}{HTML}{cc0000}
\definecolor{gDarkRed}{HTML}{990000}
\definecolor{gPurple}{HTML}{9900ff}
\definecolor{gBlue}{HTML}{1155cc}

\definecolor{gLightGrey}{HTML}{cccccc}
\definecolor{gDarkBlue}{HTML}{1c4587}
\definecolor{gDarkCyan}{HTML}{0c343d}

\definecolor{gThemeBlue}{HTML}{4285f4}
\definecolor{gThemeCyan}{HTML}{46bdc6}
\definecolor{gThemeOrange}{HTML}{ff6d01}
\definecolor{gThemeRed}{HTML}{ea4335}
\definecolor{gThemeGreen}{HTML}{34a853}

\colorlet{lightCyan}{cyan!10}  
\renewcommand{\ng}[1]{\textcolor{gGreen}{\textscale{0.85}{#1}}}
\newcommand{\nr}[1]{\textcolor{gRed}{\textscale{0.85}{#1}}}
\newcommand{\n}[1]{{\textscale{0.85}{#1}}}

\newcommand{\nCrt}[1]{\textcolor{gDarkRed}{#1}}  
\newcommand{\na}{---}  

\newcommand{\tplVar}[1]{\texttt{\textcolor{gDarkBlue}{\{#1\}}}}  
\newcommand{\tplVal}[1]{\begingroup\textcolor{gDarkBlue}{#1}\endgroup}
\newcommand{\nl}{\textcolor{gDarkCyan}{\textbackslash n}}
\newcommand{\pptSep}{\hspace{0.04em}-{}-{}-}  

\usepackage{scrextend}  

\newlength\tindent
\newcommand{\prompt}[1]{
    \setlength\parindent{0pt}
    {#1}
    \setlength\parindent{\tindent}  
}

\newcommand{\Prompt}{\noindent \emph{\normalsize Prompt to ChatGPT}}
\newcommand{\Response}{\noindent \emph{\normalsize Example Response from ChatGPT}}
\newcommand{\promptNResponse}[2]{  
    \Prompt

    \prompt{#1}

    ~\par

    \Response

    \prompt{#2}
}

\newcommand{\pptHline}{
    \arrayrulecolor{gLightGrey}\addlinespace[0.125em]
        \hline
    \addlinespace[0.125em]\arrayrulecolor{black}
}
\newcommand{\crtCline}{
    \arrayrulecolor{gLightGrey}\addlinespace[0.125em]
        \cline{2-4}
    \addlinespace[0.125em]\arrayrulecolor{black}
}
\newcommand{\lightCline}[3]{
    \arrayrulecolor{gLightGrey}\addlinespace[#2]
        \cline{#1}
    \addlinespace[#3]\arrayrulecolor{black}
}

\newcommand{\txtRule}{  
    \vspace{-0.2em}
        \noindent\textcolor{gLightGrey}{\rule[0.3em]{\linewidth}{0.4pt}}
    \vspace{-1.5em}
}

\newcommand{\tplNER}[2]{  
    Sentence: "\tplVal{#1}"
    
    Named Entities: [\tplVal{#2}]\nl
}
\newcommand{\tplNERq}[2]{  
    Query: "\tplVal{#1}"
    
    Named Entities: [\tplVal{#2}]\nl
}
\newcommand{\tplCrt}[3]{  
    Sentence: "\tplVal{#1}"
    
    Text Span: "\tplVal{#2}"

    Label: \tplVal{#3}\nl
}
\newcommand{\tplCrtS}[2]{  
    Sentence: "\tplVal{#1}"
    
    Text Span: "\tplVal{#2}"\nl
}

\newcommand{\emphSpan}[1]{\{\{#1\}\}}  

\let\oldtexttt\texttt  
\renewcommand{\texttt}[1]{{\textscale{0.95}{\oldtexttt{#1}}}}

\newcommand{\crtSpan}[1]{\textcolor{gThemeGreen}{#1}}
\newcommand{\crtType}[1]{\textcolor{gThemeBlue}{#1}} 
\newcommand{\crtNA}[1]{\textcolor{gThemeOrange}{#1}} 
\newcommand{\lbSpan}{\crtSpan{Span}}
\newcommand{\lbType}{\crtType{Type}}
\newcommand{\lbNA}{\crtNA{NA}}

\usepackage{amssymb}
\usepackage{pifont}
\newcommand{\cmark}{\ding{51}}%
\newcommand{\xmark}{\ding{55}}%

\newcommand{\error}[1]{\textcolor{gRed}{#1}} 
\newcommand{\correct}[1]{\textcolor{gGreen}{#1}} 

\newcommand{\guideA}[1]{\textcolor{gBlue}{#1}}
\newcommand{\guideB}[1]{\textcolor{gPurple}{#1}}

\newcommand{\lastRefer}[1]{\vspace{0.5em} \noindent #1}

\newcommand{\authorSep}{\hspace{1pc}}%

\title{ProgGen: Generating Named Entity Recognition Datasets Step-by-step \\ with Self-Reflexive Large Language Models}

\author{%
    Yuzhao Heng \authorSep Chunyuan Deng \authorSep Yitong Li \\
    {\bf Yue Yu \authorSep Yinghao Li \authorSep Rongzhi Zhang \authorSep Chao Zhang} \\
    Georgia Institute of Technology \\
    \texttt{\{stefan.heng,cdeng73,yli3277,yueyu,yinghaoli,rongzhi.zhang,chaozhang\}@gatech.edu}
}

\begin{document}
\maketitle
\begin{abstract}
    Although Large Language Models (LLMs) exhibit remarkable adaptability across domains, these models often fall short in structured knowledge extraction tasks such as named entity recognition (NER).
    This paper explores an innovative, cost-efficient strategy to harness LLMs with modest NER capabilities for producing superior NER datasets.
    Our approach diverges from the basic class-conditional prompts by instructing LLMs to self-reflect on the specific domain, thereby generating domain-relevant attributes (such as category and emotions for movie reviews), which are utilized for creating attribute-rich training data. 
    Furthermore, we preemptively generate entity terms and then develop NER context data around these entities, effectively bypassing the LLMs' challenges with complex structures.
    Our experiments across both general and niche domains reveal significant performance enhancements over conventional data generation methods while being more cost-effective than existing alternatives.\footnote{Our code, prompts, and generated datasets are available at \url{https://github.com/StefanHeng/ProgGen} for reproducing our results.}
\end{abstract}

\section{Introduction}

Recently, Large Language Models (LLMs) have showcased their impressive capabilities across various NLP downstream tasks, including question answering, mathematical reasoning, and code generation~\citep{touvron2023llama,palm2,openai2023gpt4}. Beyond these tasks, there is also a growing interest in exploiting LLMs' strong capabilities and extensive parametric knowledge for synthetic data generation. While previous research has predominantly focused on generating datasets for instruction tuning~\citep{wang2023selfinstruct,xu2023baize,li2023camel} or text classification tasks~\citep{supergen,meng2023tuning,ye-etal-2022-zerogen,ye-etal-2022-progen,gupta2023targen,li-etal-2023-synthetic}, there has been limited focus on producing high-quality datasets for information extraction (IE). This gap can be attributed to the deficiency of LLMs in handling the complex formats and relationships between entities~\cite{Li2023EvaluatingCI,Wei2023ZeroShotIE}.

\begin{figure*}[tb]
    \centering
    \includegraphics[width=\textwidth]{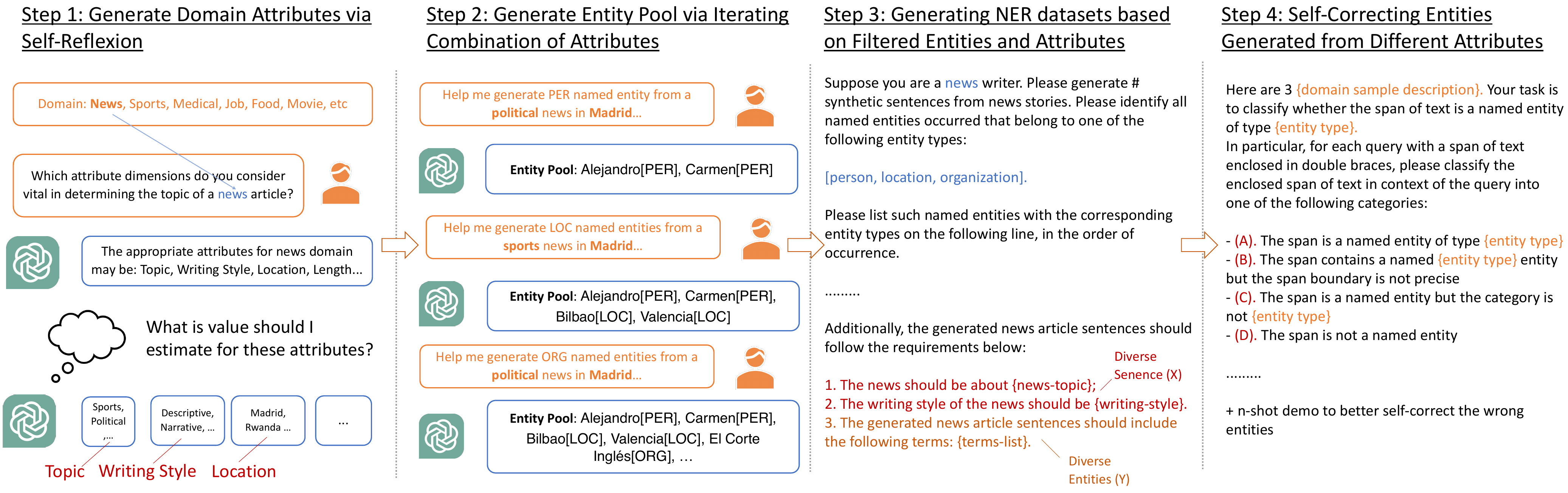} \vspace{-1.2pc}
    \caption{ProgGen NER data generation pipeline. 
        Given a dataset domain, a set of interested entity classes with definitions, and a few demo samples, 
        we prompt an LLM step-by-step to generate diverse NER samples. 
        We leverage the generated samples to train a small model for NER.
    }
    \label{fig:pipeline}
\end{figure*}

Stemming from the limited capabilities of IE, LLMs also demonstrate inadequate performance in the domain of Named Entity Recognition (NER) \citep{Xie.2023.empirical, universal-ner, li2024gno}.
Achieving high performance in NER typically necessitates a supervised model trained on in-domain data, which is a requirement hard to meet in real-world scenarios.
Leveraging the vast knowledge of LLMs to enhance the capabilities of small supervised models remains a challenging task.
Several previous works~\citep{few-shot-ner,zhou-etal-2022-melm} augment training data through the masked language modeling (MLM) task.
For example, \citet{ge-ner} employ a retrieval system to generate NER samples from open-domain QA datasets; 
\citet{universal-ner} utilize LLMs to \emph{annotate} a pre-existing corpus for supervised fine-tuning, a process that necessitates domain-specific corpora for application.

We introduce ProgGen, a new method that leverages the self-reflexive capabilities of LLMs to progressively synthesize NER datasets.
Our method, in contrast to traditional approaches that directly generate synthetic NER datasets, operates through a sequential process involving multiple steps.
To overcome the challenges posed by LLMs' limited ability to handle complex structures, we initially focus on generating entity terms, which are then employed for constructing the NER dataset.
This step-by-step generation process allows for more effective dataset creation in new domains.

In addition to progressively generating NER datasets, we follow several recent studies to leverage LLMs' ability to self-reflect~\cite{reflexion, self-refine} from two perspectives. First, instead of employing simple class-conditional prompts, we leverage LLMs to generate domain-related attributes. For each attribute, we enumerate several values, iterating through combinations of these values to potentially generate a diverse and attributed dataset. Second, we adopt the self-reflection capability to examine the correctness of the entities it generates. These methods expand our usage from solely relying on the parametric knowledge of LLMs to leveraging their cognitive abilities to generate high-quality NER data.

Our findings reveal that a small supervised model, fine-tuned on our dataset, demonstrates robust performance across both general-domain and new-domain datasets in a low-resource context. Additionally, our approach exhibits cost-efficiency advantages over traditional prompt methods or LLMs employing in-context learning. We summarize our key contributions as follows:

\begin{enumerate}
    \item \emph{Novel Progressive Generation Method} - 
    We deploy a step-wise method starting from generating entities for subsequent NER sample generation, 
    where we also leverage LLMs' cognitive ability to self-flex on attributes and self-correct wrong NER samples. 
    \item \emph{Low-Resource Utilization} -  
    Our approach minimizes reliance on human-annotated samples, substantially smaller compared to the datasets required by prior models 
    and are cheap in terms of inference cost,
    making our approach particularly advantageous in scenarios where large annotated datasets are unavailable or impractical to compile.
    \item \emph{Enhanced Efficiency and Performance} - Our downstream model are efficient compared to LLM inference and exhibits superior performance in specialized domains,  
    surpassing the standard data generation baseline by 8\% in F1 score in the best case. 
\end{enumerate}

\section{Related Work}

LLMs have opened new avenues for generating synthetic data, as highlighted by recent studies~\citep{supergen,meng2023tuning,ye-etal-2022-zerogen,ye-etal-2022-progen,yu-etal-2023-regen,gupta2023targen,li-etal-2023-synthetic}, mainly for text classification tasks. 
However, the use of overly simplistic and generic prompts in these methods often yields suboptimal results, as the representativeness and the diversity of the synthetic data. 
To overcome this, some studies have adopted interactive learning strategies to enhance instance generation~\citep{liu-etal-2022-wanli,zhang2022adaptive,attr-prompt,chung-etal-2023-increasing}, albeit at the expense of increased human intervention.


  


Generating synthetic data for IE is challenging due to the need for detailed consideration of entities and their relationships. To address this, some studies incorporate additional rules~
\citep{zhang2022prboost} or knowledge graphs~\citep{chia-etal-2022-relationprompt,josifoski-etal-2023-exploiting} to assist IE synthetic data generation.
\citet{xu-etal-2023-s2ynre} propose a two-stage training pipeline to alternately learn from synthetic and golden data together. 
For the NER task studied in this work, \citet{prompt-da,zhou-etal-2022-melm} used masked language modeling to augment the training data, GeNER \citep{ge-ner} leverages an entity retrieval model to create entity labels. Recently, several works studied to generate synthetic text with LLMs, but focus on general-domain or biomedical NER with fixed entity type~\citep{guo-roth-2021-constrained,hiebel-etal-2023-synthetic,tang2023does,xu2023knowledge}.
These approaches, however, do not fully accommodate the diversity of entity types encountered in broader, open-world scenarios. 
\citet{universal-ner} also studied creating NER instruction tuning data, but they focused on using LLMs to \emph{label} the corpus to create data for supervised finetuning, which diverges significantly from the synthetic data generation focus of our study.





\section{Method}

\begin{table}[tb]
    \centering
    \footnotesize
        



    \begin{tabular}{ m{0.95\columnwidth} }
        \toprule \hspace{-0.75em} \textbf{Simple Prompt} \\ 
        \emph{Instruction}: Suppose you are \tplVal{a news writer}. Please generate \tplVar{\#generate} synthetic \tplVal{sentences} from \tplVal{news stories}. Please identify all named entities occurred that belong to one of the following entity types:

\tplVal{[person, location, organization]}.

Please list such named entities with the corresponding entity types on the following line, in the order of occurrence.

If no entity is found in the generated sentence, leave the brackets empty. \\
        \pptHline
        \emph{Examples}: Here are some example sentences and annotations for your reference. Please follow this format.\nl

Examples:\nl



\tplNER{Saudi Arabia executes Pakistani man.}{Saudi Arabia (location)}

\tplVal{...}
 \\

        \addlinespace[0.125em] \hline \addlinespace[0.25em] \hspace{-0.75em} \textbf{Diversify X} \\ \tplVar{Simple Prompt}\nl


Additionally, the generated news article sentences should follow the requirements below:

1. The news should be about \tplVar{news-topic};

2. The writing style of the news should be \tplVar{writing-style}.
 \\

        \addlinespace[0.125em] \hline \addlinespace[0.25em] \hspace{-0.75em} \textbf{Diversify Y} \\ 
\tplVar{Simple Prompt}\nl


Additionally, in the generated sentences, include the following terms: [\tplVar{terms-list}].
 \\

        \addlinespace[0.125em] \hline \addlinespace[0.25em] \hspace{-0.75em} \textbf{Diversify X + Y} \\ \tplVar{Simple Prompt}\nl


Additionally, the generated news article sentences should follow the requirements below:

1. The news should be about \tplVar{news-topic};

2. The writing style of the news should be \tplVar{writing-style}.

3. The generated news article sentences should include the following terms: [\tplVar{terms-list}].
 \\
        \bottomrule
    \end{tabular}



    \caption{NER Sample Generation prompt templates for the CoNLL-2003 dataset. We append explicit diversity requirements to the standard sample-generation prompt. } \vspace{-0.5pc}
    \label{tbl:prompt}
\end{table}
Our workflow is succinctly depicted in Figure~\ref{fig:pipeline}. At the outset, we engage LLMs in a sequential process to generate a variety of NER samples.
Unlike traditional methods that rely on domain-specific, unlabeled corpora, our approach is more pragmatic and resource-efficient. We utilize only a handful of examples to prompt LLMs to produce sentences along with their corresponding named entity annotations. This method is exemplified in Table~\ref{tbl:prompt}, which showcases a standard few-shot prompt designed for NER sample generation\footnote{NER sample representation were searched over 4 choices as detailed in Appendix \ref{apd:sample-format-search}.}. 
Our procedure involves requesting LLMs to produce multiple samples within a single completion to enhance efficiency. Subsequently, we carefully select a subset of entity annotations identified by LLMs as particularly challenging. These annotations are re-annotated by the same LLM, this time with the aid of additional, manually crafted annotation guidelines and a selection of representative examples specifically chosen for this task. To perform NER tasks, we utilize DeBERTa~\cite{he2021deberta} trained on the generated dataset, demonstrating the practicality and effectiveness of our low-resource, LLM-based approach for NER sample generation.\footnote{See Appendix \ref{apd:templates-n-examples} for additional prompts and generated datasets used in our study.}

\subsection{Problem Setup}
Our setup is formalized as follows. 
We are given a brief task domain description $d$, target named entity classes $\mathcal{Y}$ with their definitions,
and a few demo samples $\mathcal{D}_{\mathrm{demo}} = \{(x_i, y_i)\}_{i=1}^n$ 
where $n < 10$. 
Each sample consists of a sentence $x_i$ paired with an entity list $y_i={(e,c)}^{n_i}$, consisting of contiguous text spans in $x_i$ and their corresponding entity classes.
Utilizing these information, 
we prompt LLMs in a step-wise fashion and process the LLMs' responses into NER samples $\mathcal{D}_{\mathrm{gen}}$.
The aggregated dataset $\mathcal{D_{\mathrm{gen}}}$, combined with the demo samples $\mathcal{D_{\mathrm{demo}}}$, is used to train a small model for NER. 
The model is then evaluated on an unseen test set.





\begin{figure*}[tb]
  \centering
    \includegraphics[width=0.9\textwidth]{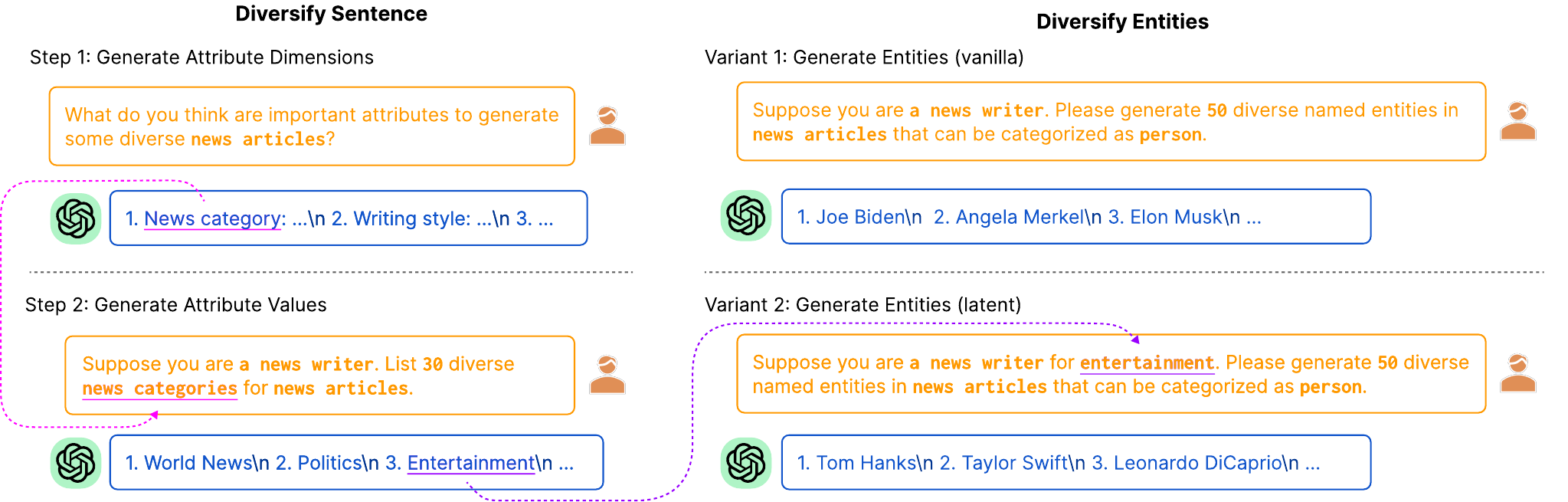} 
    \caption{Diversity Requirement Generation workflow. 
        \textbf{(Left)} Diversify Sentence:
        LLMs are prompted to generate attribute dimensions first and then attribute values given each dimension. 
        \textbf{(Right)} Diversify Entities: 
        LLMs are prompted to generate named entities for entity class, optionally conditioned on a domain-specific ``topic'' category. 
    }    
    \label{fig:diverse-x&y}
\end{figure*}




\subsection{Sample Diversity}
\label{sec:diversity}
To enhance the diversity of our generated NER samples, we augment a standard NER-sample-generation prompt (Simple Prompt) 
by incorporating explicit diversity requirements for the generated samples through configuration sampling. 
We propose 4 variants aiming to enrich sentence semantics and entity variety (examples in Table \ref{tbl:prompt}). 
Processed diversity requirement pool with statistics are available at Appendix \ref{apd:diversity-stats}. 

\subsubsection{Diversify Sentence}
\label{sec:diverse-x}

We study imposing explicit requirements on various semantic aspects of the generated sentences (Diversify X).
Inspired by several previous work to generate diverse data~\citep{attr-prompt,chung-etal-2023-increasing}, we generate attribute dimensions and values given the dataset domain in a \emph{human-AI collaboration} fashion (illustrated in Figure \ref{fig:diverse-x&y}). 

\paragraph{Generate Attribute Dimension}
We first prompt LLMs to generate diverse attribute dimensions given the dataset domain. 
Given that this step consumes relatively less tokens and thus less cost, we leverage the more capable GPT-4 model. 
We manually review and select suitable attribute dimensions for subsequent steps. 

For each dataset, we also include a ``sample topic/category'' dimension , such as ``news topic'' and ``movie query category'', 
leveraged for subsequent diversity variants (\cref{sec:diversify-Y} \& \cref{sec:diversify-X+Y}). 

\paragraph{Generate Attribute Values}
In this step, LLMs are prompted to generate potential values for each selected attribute dimension, 
where we examine and filter out low-quality values. 
We utilize GPT-3.5 for its cost-effectiveness.

\paragraph{Construct Diversity Requirement}
We sample each generated attribute dimension at a pre-determined probability such that the expected number of requirements in a sample-generation prompt $R_x$ is small ($\in [1, 2]$). 
We then sample an attribute value for each sampled attribute dimension. 
The diversity requirement templated from the sampled dimension-value pairs.

\subsubsection{Diversify Entities} 
\label{sec:diversify-Y}
We study explicit requirements on inclusion of named entities in the generated sentences (workflow illustrated in Figure \ref{fig:diverse-x&y}). 

\paragraph{Generate Entities} 
We generate named entities by prompting GPT3.5 given dataset domain (Diversify Y). 
Similarly to the Diversify Sentence process, we filter out low-quality generated entities based on manual inspection. 

We explore two variants, (1) \textit{vanilla} where we generate $M$ entities for each entity class, 
and (2) \textit{latent} where we generate $M$ entities conditioned on each entity class and additionally on each domain-specific topic value (via the dataset-independent ``topic'' attribute). 



\paragraph{Construct Diversity Requirement}
We sample from the generated entity pool with an expected number of exampled entity $R_y$. 
We then construct the diversity requirement from the sampled named entities via a template. 
We omit the entity type corresponding to each sampled entity so that LLMs have more freedom in entity type annotations\footnote{We found that this lead to performance gains in preliminary experiments. Additional discussion in Appendix \ref{apd:data-gen-diverse-y-setup}.}. 

\subsubsection{Diversify Both Sentence and Entities} 
\label{sec:diversify-X+Y}
We provide explicit requirements on both sentence semantic and entity inclusion (Diversify X + Y). 
We first construct the sentence requirements, then we use the sampled ``topic'' attribute value to sample named entities 
from the corresponding \textit{latent} entity pool such that the diversity requirements are semantically consistent. 



\begin{table}[tb]
    \centering 
    \footnotesize
    \begin{tabular}{ m{0.95\columnwidth} }
        \toprule

        \emph{Instruction}: Here are \tplVar{\#correct} \tplVar{domain sample description}. Your task is to classify whether the span of text is a named entity of type \tplVar{entity type}.

In particular, for each query with a span of text enclosed in double braces, please classify the enclosed span of text in context of the query into one of the following categories:

\hspace{0.05em}- (A). The span is a named entity of type \tplVar{entity type}

\hspace{0.05em}- (B). The span contains a named \tplVar{entity type} entity but the \crtSpan{span boundary is not precise}

\hspace{0.05em}- (C). The span is a named entity but the \crtType{category is not} \tplVar{entity type}

\hspace{0.05em}- (D). The span is \crtNA{not a named entity}

If the label is (B), you should provide the \crtSpan{correct span boundary}.

If the label is (C), you should provide the \crtType{correct entity type}. Pick an entity type from the following list:

[\tplVar{remaining entity types in the dataset}, other].\nl \\
        \pptHline
        \tplVar{entity-type-specific annotation instruction}\nl \\
        \pptHline
        \emph{Examples}: Here are some example span classifications for your reference. Please follow this format.\nl

Examples:\nl 

\tplVar{(sentence, span, correction) demo examples}\nl

\pptSep \\
        \pptHline
        Please classify the following \tplVar{\#correct} spans of text:\nl

\tplVar{(sentence, entity span) samples to correct} \\
        \bottomrule
    \end{tabular}
    \caption{
        LLM Self-Correction prompt template. 
        For each entity annotation prompted, LLMs are instructed to optionally (1) \crtSpan{ correct the entity span}, (2) \crtType{correct the entity type}, or 3) \crtNA{drop the annotation}.}
    \label{tbl:correction-prompt}
\end{table}

\subsection{LLM Self-Correction}
\label{sec:self-correction}

Creating NER datasets critically involves defining entities and providing annotation guidelines for edge cases to ensure clarity \citep{universal-ner}. 
These guidelines often vary across datasets, even for the same entity types (e.g., Location in news vs in restaurant queries). 
Given LLMs' ability to learn from feedback and correct their own mistakes \citep{reflexion, self-refine}, 
we tailor instructions and examples for each dataset and entity class, directing LLMs to refine their annotations,
aiming to align more closely with dataset-specific nuances. 
Further, to enhance cost efficiency, only a subset of annotations that LLMs find ``challenging'' are selected for self-correction.

\subsubsection{Annotation Uncertainty Ranking}
\label{sec:annotation-uncertainty-ranking}
We identify challenging samples using a scoring function that averages the log-probability of tokens within each entity annotation, i.e. \texttt{entity span (entity type)}, 
where a lower score means higher uncertainty and thus more challenging. 
Preliminary experiment compares this function with averaged token-wise loss values from a downstream model trained on the generated data 
(Appendix \ref{apd:compare-annotation-uncertainty-ranking-function}), 
and shows that self-correction with the former scoring function yields better downstream performance.

\subsubsection{Prompt Construction}
We stratify all annotations to self-correct by the entity type annotated, allowing for more precise annotation instructions 
given the same token budget. 
Conditioned on the entity class, 
an LLMs are given (context, entity) pairs and are instructed to optionally improve the annotation by
(1) \crtSpan{correcting the entity span boundary}, (2) \crtType{correcting the entity type}, and (3) \crtNA{dropping the entity annotation}
(template shown in Table \ref{tbl:correction-prompt}). 
We batch correction samples for cost efficiency. 


We construct the entity-type-specific prompts manually by alternating between the following 2 steps. 
(1) We examine top-uncertain entity annotations for each generated dataset to identify common error cases in LLMs' annotations. 
We write correction instructions including common error cases. 
We hand-picked diverse demo samples from the generated datasets that represent the most common failure cases. 
(2) In case of uncertain annotation or class definition ambiguities (especially a problem for specific-domain datasets), 
since we were not able to find detailed annotation schemas for the datasets we studied, 
we inspected the original samples in the dataset to infer the annotation guidelines. 
We then manually write the entity definitions and annotation instructions. 


\vspace{0.5em}\noindent Further analysis of ProgGen design choices are available in Appendix \ref{apd:experiments}.

\begin{table}[tb]
    \resizebox{0.47\textwidth}{!}{
    \centering 
    \begin{tabular}{ l l c c c}
        \toprule \thead{Dataset} &       \thead{Domain} &           \thead{\#sample} &  \thead{C} & \thead{D} \\
        \midrule
            CoNLL-2003* & News & 15K / 3.5K / 3.7K &              3 &             4 \\
            WikiGold* &   Wikipedia  &              1.7K &              3 &              3 \\
             MIT Movie &        Movie &       9.7K / 2.4K &             12 &             7 \\
            MIT Restaurant &   Restaurant &       7.7K / 1.5K &              8 &             5 \\
        \bottomrule
    \end{tabular}}
    \caption{
        Dataset Statistics. \#sample is provided for the original train/dev/test or train/test splits respectively if available. 
        C and D refer to the number of entity classes and the number of demo samples we use respectively. 
    }
    \label{tbl:dataset-stats}
\end{table}

\begin{table*}[tb]
    \centering 
      \resizebox{0.95\linewidth}{!}{
    \begin{NiceTabular}{@{\extracolsep{6pt}} l l *{10}{c}}[colortbl-like]
        \toprule
        \multicolumn{2}{c}{\multirow{2}{*}[-0.2em]{\thead{Method}}} 
            & \multicolumn{2}{c}{\thead{CoNLL-2003}} & \multicolumn{2}{c}{\thead{WikiGold}} & \multicolumn{2}{c}{\thead{MIT-Movie}} & \multicolumn{2}{c}{\thead{MIT-Restaurant}} & \multicolumn{2}{c}{\thead{Average}} \\
        \addlinespace[0.2em] \cline{3-4} \cline{5-6} \cline{7-8} \cline{9-10} \cline{11-12} \addlinespace[0.2em] 
        & & \thead{F1} & \thead{Price} & \thead{F1} & \thead{Price} & \thead{F1} & \thead{Price} & \thead{F1} & \thead{Price} & \thead{F1} & \thead{Price} \\
        
        \midrule 
        \multirow{3}{*}[-0em]{Few-Shot} 
            & Few Shot-FT \citep{huang-etal-2021-shot} & 61.4 &  \na & 64.0 & \na & 38.0  & \na &  44.1 & \na & 51.9 & \na \\
            & SpanNER \citep{wang-etal-2021-learning-language-description} & 71.1 &  --- & \na & \na & 65.4  & \na & 49.1 & \na & \na & \na \\
            & Ent-LM \citep{few-shot-ner} & 66.8 & \na & \na & \na & 59.4  & \na & \na & \na & \na & \na \\
        \midrule 
        \multirow{3}{*}[-0em]{Data Gen.} 
            & GeNER \citep{ge-ner} & 71.0 & \na & 72.5 & \na & \na & \na & \na & \na & \na & \na \\
            & LLM-DA \citep{llm-da} & 54.6 & \na & \na & \na & 58.9 & \na & \na & \na & \na & \na \\
            & Simple Prompt (ours) & 72.0 & 0.1 & 75.8 & 0.16 & 64.8 & 0.13 & 58.4 & 0.13 & 67.8 & 0.13 \\
        
        \midrule
        \multirow{8}{*}[-0em]{\makecell{ProgGen\\(ours)}} 
            & Diversify X 
                & 69.2 \nr{-2.8} & 0.24 & 75.5 \nr{-0.3} & 0.33 & 65.0 \ng{+0.2} & 0.36 & 54.7 \nr{-3.7} & 0.31 & 66.1 \nr{-1.7} & 0.31\\
            \rowcolor{lightCyan} \cellcolor{white} & \quad + Self-Correction 
                & 71.9 \nr{-0.1} & 0.33 & 77.3 \ng{+1.5} &0.44 & \nCrt{63.5} \nr{-1.3} & 0.50 & 56.1 \nr{-2.3} & 0.55 & 67.2 \nr{-0.6} & 0.46 \\ 
            
            & Diversify Y (vanilla) 
                & 71.2 \nr{-0.8} & 0.24 & 76.2 \ng{+0.4} & 0.31 & 73.0 \ng{+8.2} & 0.35 & 61.5 \ng{+3.1} & 0.30 & 70.5 \ng{+2.7} & 0.30 \\
            \rowcolor{lightCyan} \cellcolor{white} & \quad + Self-Correction &
                71.4 \nr{-0.6} & 0.32 & \textbf{78.2 \ng{+2.4}} & 0.47 & \textbf{73.3 \ng{+8.5}} & 0.51 & \textbf{62.8 \ng{+4.4}} & 0.52 & 71.4 \ng{+3.6} & 0.46 \\ 
            
            & Diversify Y (latent) 
                & 75.6 \ng{+3.6} & 0.24 & 75.7 \nr{-0.1} &0.32 & 71.8 \ng{+7.0} & 0.35  & 60.9 \ng{+2.5} & 0.30 & 71.0 \ng{+3.2} & 0.30 \\
            \rowcolor{lightCyan} \cellcolor{white} & \quad + Self-Correction & 
                \textbf{77.5 \ng{+5.5}} & 0.33 & 76.4 \ng{+0.6} & 0.47 & 72.9 \ng{+8.1} & 0.52 & 61.4 \ng{+3.0} & 0.52 & \textbf{72.1 \ng{+4.3}} & 0.46 \\
            
            & Diversify X + Y 
                & 73.7 \ng{+1.7} & 0.27 & 76.1 \ng{+0.3} & 0.35 & 70.6 \ng{+5.8} & 0.39 & 59.5 \ng{+1.1} & 0.33 & 70.0 \ng{+2.2} & 0.34 \\
            \rowcolor{lightCyan} \cellcolor{white} & \quad + Self-Correction &
                74.3 \ng{+2.3} & 0.35 & \nCrt{74.6} \nr{-1.2} & 0.48 &  72.2 \ng{+7.4} &  0.54 & 59.8 \ng{+1.4} & 0.59 & 70.2 \ng{+2.4} & 0.50 \\
        
        \midrule
        \multicolumn{2}{c}{Few-shot ICL} & 71.5 & 0.39 & 74.5 & 0.49 & 74.2 & 0.61 & 67.0 & 0.48 & 71.8 & 0.49 \\
        \bottomrule
    \end{NiceTabular}
    }
    \caption{
        ProgGen Performance Comparison. 
        Price refers to the API querying cost (\$). The prices are for 1) Data Generation: Generating around 1.5K samples added by the cost for self-correction,
        and 2) Inference: few-shot learning prediction for 1.5K samples. 
        F1 \textcolor{gGreen}{gains}/\textcolor{gRed}{drops} are w.r.t Simple Prompt. 
        Relative performance \nCrt{drops} after Self-Correction are colored. 
    }
    \label{tbl:main-results}
\end{table*}

\section{Experiments}

We assess the effectiveness of our framework via downstream model performance on 
generated NER datasets.

\subsection{Datasets}
\label{sec:ori-dataset}
We examine datasets from various domains and with different number of entity classes, including: 
\begin{itemize}

    \item \textbf{CoNLL-2003} \citep{conll-2003} features news articles from Reuters with 
    three broadly-defined entity types: person, location and organization.
    
    \item \textbf{WikiGold} \citep{wiki-gold} comprises Wikipedia articles. It's a small dataset with the same set of entity types as CoNLL-2003. 
    
    \item \textbf{MIT-Movie} and \textbf{MIT-Restaurant} \citep{asgard} contain dialog system queries related to movies and restaurants, respectively. 
    Both datasets incorporate a wide range of entity types, such as movie titles, genres, characters, and names of restaurants and cuisines.
    
\end{itemize}

The dataset statistics\footnote{
    Following prior studies \citep{drop-misc, ge-ner}, 
    the miscellaneous (MISC) class is omitted due to its lack of a precise definition. 
    Datasets originally containing the MISC entity class are denoted with asterisks*.
} are summarized in Table \ref{tbl:dataset-stats}. 
For datasets that originally contain over 1\% negative samples, i.e. samples with no entity annotations, 
we include a negative sample in the set of demo samples to enhance LLMs learning what's \emph{not} a relevant named entity.

\subsection{Experiment Setup}
\label{sec:experiment-setup}

\paragraph{Data Generation} 
We use 1-shot demo as described in \citet{few-shot-ner}, wherein we include one entity annotation for each entity class. 
In the attribute dimension generation step for \emph{Diversify X}, we prompted the ChatGPT4 web App 3 times. 
All other generations are from ChatGPT3.5 through OpenAI API\footnote{
We used the \texttt{gpt-3.5-turbo-1106} checkpoint. 
}.  
We generate \textasciitilde 1.5K training samples for each dataset and for each method. 
We drop duplicates and filter out malformed samples (details in Appendix \ref{apd:ner-sample-processing}).  
See Appendix \ref{apd:data-stats} and \ref{apd:templates-n-examples} for generated datasets samples and statistics. 
We select all generated entity annotations with log-probability below $P_c = -2e^{-2}$ for LLM self-correction, capped at $T_c = 20\%$ of all entity annotations.

\paragraph{Downstream Model Training} 
For all setups, we train a DeBERTaV3 \citep{deberta-v3} model for 16 epochs with batch size 24, utilizing the \texttt{microsoft/deberta-v3-base} checkpoint from HuggingFace.
We train with the AdamW optimizer \citep{adam-w}, a learning rate of $4e^{-5}$, a weight decay of $1e^{-4}$, and a linear scheduler with warmup over the first 200 steps. 
The 1-shot demo samples are integrated into training with a weight of 5. 
Following prior work \cite{few-shot-ner,attr-prompt}, we don't use a validation set and directly evaluate the final trained model with micro-average F1.
For our approaches, we also report precision and recall. 

\paragraph{Baselines} We compare our approach with the existing few-shot baselines: (1) Few-shot FT~\citep{huang-etal-2021-shot}, which directly finetunes a pretrained language model on few-shot examples.
(2) SpanNER~\citep{wang-etal-2021-learning-language-description}, which learns from natural language supervision to identify entity classes. 
(3) Ent-LM~\citep{few-shot-ner}, which leverages masked language modeling to augment the few-shot examples. 
We also consider the following data generation baselines:
(4) GeNER~\citep{ge-ner}, which uses an entity retrieval module to retrieve from unlabeled corpus to create NER training data.
(5) LLM-DA~\citep{llm-da}, which prompts GPT-3.5 to rewrite the context and entities of few-shot samples to augment training data. 
(6) Our standard data generation prompt (Simple Prompt). 
Besides, we provide few-shot in-context learning (ICL) on the same set of demonstration samples for reference. 
We use reported results from prior work. 

\lastRefer{See Appendix \ref{apd:implementation} for additional implementation details. }

\begin{table*}[tb]
    \centering 
    \resizebox{0.95\linewidth}{!}{
    \begin{NiceTabular}{ l l *{5}{c}}[colortbl-like]
        \toprule
        \multicolumn{2}{c}{\thead{Method}} & \thead{CoNLL-2003} & \thead{WikiGold} & \thead{MIT-Movie} & \thead{MIT-Restaurant} & \thead{Average} \\
        
        \midrule
        \multicolumn{2}{c}{Simple Prompt} & 67.2 / 77.4 & 69.2 / 83.7 & 62.3 / 67.4 & 55.4 / 61.8 & 63.5 / 72.6 \\
        
        \midrule
        \multirow{8}{*}[-0.25em]{ProgGen} 
            & Diversify X 
                & 63.8 \nr{-3.4} / 75.6 \nr{-2.8} & 71.0 \ng{+1.8} / 80.6 \nr{-3.1} & 63.4 \ng{+1.1} / 66.6 \nr{-0.8} & 52.1 \nr{-3.3} / 57.6 \nr{-4.2} & 62.6 \nr{-0.9} / 70.1 \nr{-2.5} \\
            \rowcolor{lightCyan} \cellcolor{white} & \quad + Self-Correction 
                & 67.5 \ng{+0.3} / 77.0 \nr{-0.4} & 73.6 \ng{+4.4} / 81.4 \nr{-2.3} & \nCrt{62.4} \ng{+0.1} / \nCrt{64.5} \ng{-2.9} & 53.9 \nr{-1.5} / 58.5 \nr{-3.3} & 64.4 \ng{+0.9} / 70.4 \nr{-2.2} \\ 
            
            & Diversify Y (vanilla) 
                & 68.2 \ng{+1.0} / 74.5 \nr{-2.9} & 73.8 \ng{+4.6} / 78.7 \nr{-5.0} & 71.7 \ng{+9.4} / \textbf{74.4 \ng{+7.0}} & 60.4 \ng{+5.0} / 62.7 \ng{+0.9} & 68.5 \ng{+5.0} / 72.6 \n{+0.0} \\
            \rowcolor{lightCyan} \cellcolor{white} & \quad + Self-Correction
                & 69.2 \ng{+2.0} / \nCrt{73.7} \nr{-3.7} & \textbf{77.9 \ng{+8.7}} / \nCrt{78.5} \nr{-5.2} & \textbf{73.0 \ng{+10.7}} / \nCrt{73.7} \ng{+6.3} & \textbf{61.9 \ng{+6.5}} / \textbf{63.4 \ng{+1.6}} & 70.5 \ng{+7.0} / 72.3 \nr{-0.3} \\ 
            
            & Diversify Y (latent) 
                & 72.7 \ng{+5.5} / 78.8 \ng{+1.4} & 74.8 \ng{+5.6} / 76.7 \nr{-7.0} & 70.0 \ng{+7.7} / 73.7 \ng{+6.3}  & 59.9 \ng{+4.5} / 61.9 \ng{+0.1} & 69.4 \ng{+5.9} / 72.8 \ng{+0.2} \\
            \rowcolor{lightCyan} \cellcolor{white} & \quad + Self-Correction 
                & \textbf{75.0 \ng{+7.8}} / \textbf{80.2 \ng{+2.8}} & 75.8 \ng{+6.6} / 77.1 \nr{-6.6} & 71.9 \ng{+9.6} / 74.1 \ng{+6.7} & 60.5 \ng{+5.1} / 62.2 \ng{+0.4} & \textbf{70.8 \ng{+7.3}} / \textbf{73.4 \ng{+0.8}} \\
            
            & Diversify X + Y 
                & 69.2 \ng{+2.0} / 78.8 \ng{+1.4} & 75.0 \ng{+5.8} / 77.3 \nr{-6.4} & 69.3 \ng{+7.0} / 71.9 \ng{+4.5} & 58.5 \ng{+3.1} / 60.4 \nr{-1.4} & 68.0 \ng{+4.5} / 72.1 \nr{-0.5} \\
            \rowcolor{lightCyan} \cellcolor{white} & \quad + Self-Correction &
                71.8 \ng{+4.6} / \nCrt{76.9} \nr{-0.5} & \nCrt{74.0} \ng{+4.8} / \nCrt{75.2} \nr{-8.5} & 72.0 \ng{+9.7} / 72.4 \ng{+5.0} & 59.3 \ng{+3.9} / \nCrt{60.3} \nr{-1.5} & 69.8 \ng{+5.3} / \nCrt{71.2} \nr{-1.4} \\
        
        \midrule
        \multicolumn{2}{c}{Few-shot ICL} & 65.8 / 78.3 & 70.2 / 79.3 & 73.2 / 75.3 & 64.8 / 69.4 & 68.5 / 75.6 \\
        \bottomrule
    \end{NiceTabular}
    }
    \caption{
        ProgGen Performance Comparison. We show precision and recall separated by slash (precision / recall). 
    }
    \label{tbl:main-results-pr}
\end{table*}

\subsection{Results}
\label{sec:experiment-results}

Tables \ref{tbl:main-results} and \ref{tbl:main-results-pr} display our main results. 
We first note that the Simple Prompt baseline is strong and consistently out-perform prior works in few-shot NER,
stemmed from the strong base knowledge of LLMs. 
Also, few-shot ICL have recall values higher than most data generation approaches, 
reflecting loss of world knowledge when distilling LLMs to generated data. 

We further note that (1) our best-performing setup matches or outperforms ICL in 3 out of 4 datasets we study. 
In general domains (News, Wikipedia), even Simple Prompt out-performs ICL, 
indicating a small pretrained BERT model can learn from noisy LLM annotations. 
(2) Batching sample- and correction-generation achieves roughly the same cost w.r.t. ICL inference. 
Data generation cost is one-time (see additional cost analysis in Appendix \ref{apd:api-token-cost-analysis}), 
whereas few-shot ICL incurs inference API querying cost for each new prediction sample. 
Combined with the fact that (3) a small, task-specific downstream model has negligible inference overhead (c.f. LLMs), 
we highlight the efficacy of data generation for NER tasks.

\subsubsection{Sample Diversity} 
\label{sec:sample-diversity-results}
Incorporating diversity notably improves model performance, outperforming Simple Prompt across all datasets in most cases. 
Diversify Y variants are the best-performing, yielding at least 2\% F1 gains on average and up to 8\% improvement for nuanced domains. 
The \emph{latent} variant is more robust with higher average performance and more consistent gains over Simple Prompt, 
which may benefit from the implicit diversity in sentence semantics rooted from the latent ``topic'' dimension. 
Conversely, Diversify X tends to lower performance, 
a sign that sentence diversity may not benefit NER tasks, relative to entity diversity. 
Further manual inspection finds that data generated from Diversify X variants have lower-quality entity annotations (Appendix \ref{apd:analysis-correction-result-with-stats}), 
which may be the root cause of low performance. 
In particular, such annotations are more frequently incorrect and the wrong annotations encompass more disparate failure types. 
We hypothesize that explicit semantic requirements in Diversify X result in less-frequent sentences and smoother token distributions, 
thus creating room for additional LLM annotation errors. 
Combining sentence and entity diversity (Diversify X+Y) yields intermediate gains, 
likely still affected by the shortcomings of Diversify X.

\paragraph{Diversity have diminishing gains. } 
Comparing the \textit{vanilla} and \textit{latent} variants of Diversify Y, 
while the performances in general domain are mixed, 
vanilla Diversify Y wins in specific domains. 
We hypothesize that specific-domain entity classes are already narrowly focused (e.g. Director in movie queries vs. Person in news) 
such that the \emph{vanilla} setting generates sufficiently diverse datasets. 
The additional ``topic'' attribute value for the latent setting may only marginally increase diversity and 
at times introduce noise
(e.g. generate \emph{Genre} entities for movie queries about \emph{Cast and crew}; Table \ref{tbl:diverse-config-sample-values-mit-movie}; Appendix \ref{apd:diverse-y-latent-samples}).

\subsubsection{Self-Correction} 
LLM self-correction leads to additional 1-2\% F1 improvements for most cases. 
Manual inspection on LLM self-correction responses further show that most corrections are valid (i.e., turns originally wrong entity annotations into correct ones), 
indicating LLMs can learn and generalize from annotation feedback. 
LLM self-correction mostly preserves the relative trend among diversity variants as opposed to closing the gap, 
underscoring the critical role of sample diversity and the initial quality of annotations. 
Precision scores generally lag behind recall, suggesting a propensity for over-identification of named entities by LLMs, leading to more false positives \citep{gpt-ner}. 
Further, 
self-correction tends to enhance precision more than recall,
correlating with the distribution of LLM correction kinds (Appendix \ref{apd:analysis-correction-result-with-stats}). 



Below, we discuss and show examples for two most common classes of LLMs' systematic biases that yield false positives, which are the focus of annotation instructions in self-correction prompts. 
See Appendix \ref{apd:representative-annotatino-errors} for more error classes and representative samples.

\paragraph{LLMs are overly confident in what is a named entity. } 
LLMs tend to falsely annotate generic, non-named references and categorical references as named entities. 

\begin{addmargin}[\parindent]{\parindent}
    \setlength\parindent{0pt}
    \small \vspace{0.5em}
    Span: \texttt{The \error{\textbf{CEO}} of Apple Inc. announces a new product launch event.}

    Entity Type: person

    Span: \texttt{\error{\textbf{global organization}} announces plans for climate change summit in Reykjavik.}

    Entity Type: organization
\end{addmargin}

\paragraph{LLMs have overly broad definitions of entity classes. }
We found that LLMs may associate named entities of irrelevant types into one of the entity classes in prompt, 
such as counting events and awards as organization entities. 

\begin{addmargin}[\parindent]{\parindent}
    \setlength\parindent{0pt}
    \small \vspace{0.5em}

    
    Span: \texttt{Zhang Yimou, a renowned \textbf{Chinese} film director, has won numerous awards...}

    Entity Type: \error{Location} (Correct type: Demonym)

    Span: \texttt{Show me a trailer for the movie that has a song performed by \textbf{Elton John}.}

    Entity Type: \error{Song} (Correct type: Artist)
\end{addmargin}

\subsubsection{Specific Domain}
\label{sec:specific-domain-results}
In specific domains like Movies and Restaurants, data generation shows reduced effectiveness compared to general domains (News, Wikipedia), 
with most ProgGen variants lagging in F1 scores behind ICL.  
This shows that distilling LLMs' knowledge via data generation is especially challenging for more nuanced domains, 
with presumably smaller proportion of pre-training corpus on the given domain and respective entity classes. 
Notably, recall values are lower for specific domains, 
suggesting that the generated dataset might not fully capture the requisite domain-specific knowledge.
However, gains for Diversify Entity variants roughly double for specific domains compared to general domains, 
effectively narrowing the performance gap with ICL in these challenging areas.



We further note that, in specific-domain datasets, 
certain entity classes are especially hard to define and separate due to ambiguity, yielding more wrong LLM annotations 
(detailed in Appendices \ref{apd:analysis-correction-result-with-stats} and \ref{apd:entity-class-ambiguous}). 
Manual inspection and partial F1 scores reveal that 
while LLMs correctly identify many entities, inaccuracies primarily arise from imprecise span boundaries and misclassified entity types
(Appendices \ref{apd:partial-f1-eval} and \ref{apd:representative-annotatino-errors}). 

\lastRefer{See Appendix \ref{apd:experiments} for additional experiments. }

\begin{table}[tb]
    \centering
    \resizebox{1\linewidth}{!}{
        
        
    \begin{tabular}{ l c l c c c c}
        \toprule 
        \thead{Method} & \thead{Dataset\\Difference} & \thead{CoNLL\\-2003} & \thead{Wiki\\Gold} & \thead{MIT-\\Movie} & \thead{MIT-\\Rest.} \\
        \midrule
        Supervised Learning & \na & 90.3 & 88.0 & 84.8 & 76.5 \\
        
        \addlinespace[0.125em] \hline \addlinespace[0.25em] 
        LLM Annot. & Annot. quality & 75.9 \textbf{\nr{-14.4}} &  75.0 \textbf{\nr{-13.0}} &  72.6 \textbf{\nr{-12.2}} & 64.0 \textbf{\nr{-12.5}} \\
        LLM Reph. \& Annot. & Syntax & 75.8 \nr{-0.1} & 77.4 \ng{+2.4} & 71.7 \nr{-0.9} & 63.0 \nr{-1.0} \\
        
        \addlinespace[0.125em] \hline \addlinespace[0.25em] 
        LLM Generated & Semantics & 75.6 \nr{-0.2} & 76.2 \nr{-0.8} & 73.0 \ng{+1.3} & 61.5 \nr{-1.5} \\ 
        \bottomrule
    \end{tabular}}
    
    \caption{Sentence distribution and LLM annotation accuracy ablation. Dataset Difference and \textcolor{gRed}{Performance Drop values} are w.r.t. the row above. }
    \label{tbl:sentence-n-annotation-ablation}
\end{table}

\begin{figure}[tb]
  \centering
    \includegraphics[width=1\linewidth]{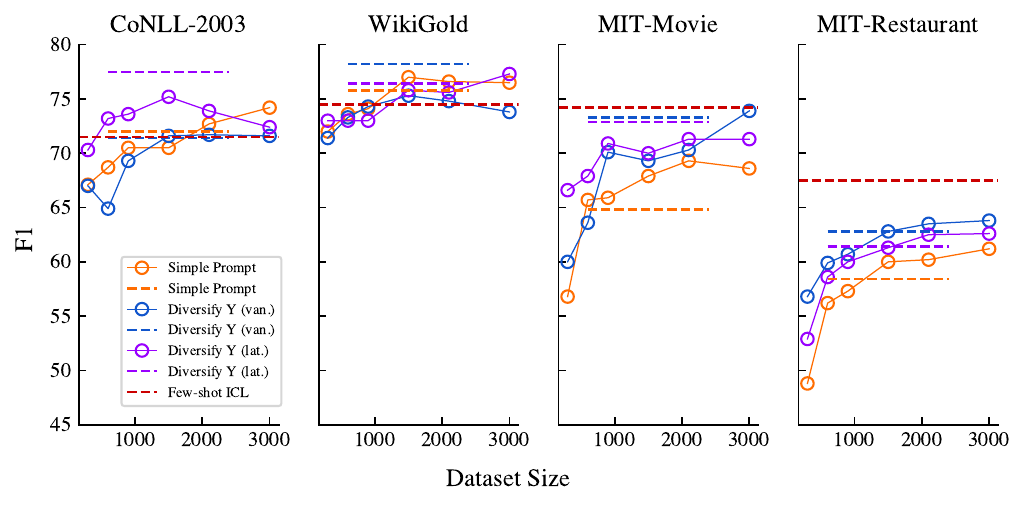} 
    \vspace{-1.8pc}
    \caption{Sample Diversity Scaling plot. Corresponding F1 from the main results (Table \ref{tbl:main-results}) are shown in dashed lines.}
    \label{fig:diversity-scaling}
\end{figure}

\section{Discussion}

\paragraph{Scaling diversity alone plateaus} 

Using generated dataset size as a rough proxy of sample diversity, we plot its relationship with downstream model performance in Figure \ref{fig:diversity-scaling}. We compare Simple Prompt with the top-performing Diversify Y variants where we re-generate 3K valid samples. We take 6 subsets of sample sizes [300, 600, 900, 1.5K, 2.1K, 3K] where we enforce the smaller subsets are contained in the larger subsets.
We then train NER models as in \cref{sec:experiment-setup}. 
We observe that substantially increasing dataset size doesn't lead to much performance gains across the board, confirming our speculation that sample diversity alone has diminishing returns (\cref{sec:sample-diversity-results}).

\paragraph{Entity annotation accuracy is performance bottleneck}
We conduct an ablation experiment to evaluate the impact of sentence distribution and LLM annotation accuracy on model performance, 
using modified datasets described below:
\begin{itemize}
    \item \textit{Supervised Learning}. A 90/10 train/dev split on 1.5K dataset samples as an upperbound\footnote{For WikiGold, we use the entire training dataset with just 1.1K samples.}. 
    \item \textit{LLM Annotated}. A 1.5K-sample subset of the training set containing the 1,350 training samples above, with entity annotations from ChatGPT.
    \item \textit{LLM Rephrased \& Annotated}. Sentences from the samples above, first rephrased and then annotated by ChatGPT. 
    \item \textit{LLM Generated}. The best-performing diversity setup using 1.5K LLMs-generated data for reference (from Table \ref{tbl:main-results}). 
\end{itemize}

We then train NER models as in \cref{sec:experiment-setup}. For the \textit{Supervised Learning} setup, we use the model with best dev loss. The results are shown in Table \ref{tbl:sentence-n-annotation-ablation}. 
We note that differences in sentence syntactic and semantic distributions hardly influence downstream performance, where as
the largest drop in performance is attributed to the lower-quality of LLMs-generated annotations compared to the original. 
On the other hand, top-performing data-generation approaches achieve similar performances, indicating that 
with sufficient diversity, LLM-generated sentences matches a domain-relevant unlabeled corpus. 
Thus, annotating accurately w.r.t entity class definitions becomes the performance bottleneck. 

\paragraph{Step-wise sample generation} 

Considering the limited instruction following capabilities of GPT-3.5 \cite{instruct-gpt} may influence performance, 
we explored generating NER samples step-wise (from sentences, entity spans, to entity types) 
with manually-written Chain-of-Thought \cite{cot} demos for entity annotation. 
We found no noticeable performance gains, despite the higher token querying costs due to re-prompting partially generated samples.
See Appendix \ref{apd:step-wise-gen} for additional discussion. 


\section{Conclusion}


We present ProgGen, an innovative framework that
leverages LLMs to generate 
diverse and accurate task-specific datasets 
via requirement sampling and self-reflection
for low-resource NER.
Experimental results on 4 datasets encompassing various domains demonstrate the advantages of our data generation approach compared to LLM few-shot in-context learning 
both in downstream model performance and cost-efficiency.
Our research highlights the critical role of entity diversity in improving dataset quality and demonstrates the effectiveness of incorporating LLM self-correction alongside manual feedback to enhance annotation accuracy. 
Further analysis on the generated datasets reveals 
persistent biases in LLM entity classification
and that annotating accurately is a major bottleneck 
to close the gap with traditional supervised learning methods.

\section{Limitation \& Future Work}
\paragraph{More Comprehensive Experiments}
Our investigation focused on widely recognized NER benchmarks. 
Although we did not find signs of LLMs regurgitating 
training samples\footnote{For example, in CoNLL-2003, we found that LLMs-generated news articles that are more recent, as opposed to news around 2003. } \citep{llm-privacy}, 
experiments on less-widely-used datasets such as datasets in the medical domain or datasets released after LLMs' pretraining cutoff times would yield more reliable findings,
as such datasets are less likely to be contaminated. 
Additionally, our work can be viewed as data augmentation which may particularly benefit lower-resource domains. 

Furthermore, 
our study was constrained to utilizing ChatGPT3.5, primarily due to budgetary limitations. 
Expanding future research to a broader range of LLMs could provide more insights, as more advanced LLMs may follow instructions and annotate entities better. 



\paragraph{Enhancing Automation}
Selecting representative demos for LLM Self-Correction can get labor-intensive since it involves examining large number of entity annotations
for each generated dataset. 
It's worth assessing annotation instructions and demos shared for a given task domain. 
Advancing automatic feedback generation leveraging stronger LLMs such as GPT-4 can help, namely in (1) crafting annotation instructions and (2) evaluating annotations from weaker LLMs. 

Further, the hyperparameter choices for data generation in this work is setup ad hoc whereas they may influence downstream performance noticeably. 
Future work can explore how to optimize these hyperparameters under the few-shot setup. 

\paragraph{Better Annotations} We found that LLMs may generate different/conflicting annotations on the same sentence. Such uncertainty signals can be leveraged to identify challenging instances and improve entity annotations. 
See Appendix \ref{apd:future-work} for additional discussions.

\section*{Ethics Statement}

We acknowledge the potential biases of texts generated by language models which 
may not be factual and appropriately represent the included entities (including persons, locations, organizations, movies, restaurants) in the generated datasets. 
In our studies, we did not find noticeable cases of toxic language in the generated data as OpenAI API have content moderation policies in place. 
Nonetheless, harmful synthetic sentences may be present in the processed datasets. 

Our research utilized publicly accessible NER benchmarks and was designed to be computationally efficient, 
with each training session on a generated dataset requiring no more than 10 minutes on a NVIDIA GeForce RTX 2080 Ti GPU.

\section*{Acknowledgements}
This work was supported in part by NSF CAREER IIS-2144338, ONR MURI N00014-17-1-2656, and research grant from ADP.
We appreciate the compute resources provided by Chao Zhang's group. 
We also thank Junmo Kang, Alan Ritter and the anonymous reviewers for their insightful feedback on the project.


\bibliography{custom,anthology}

\begin{thebibliography}{58}
\expandafter\ifx\csname natexlab\endcsname\relax\def\natexlab#1{#1}\fi

\bibitem[{Anil et~al.(2023)Anil, Dai, Firat, Johnson, Lepikhin, Passos, Shakeri, Taropa, Bailey, Chen et~al.}]{palm2}
Rohan Anil, Andrew~M Dai, Orhan Firat, Melvin Johnson, Dmitry Lepikhin, Alexandre Passos, Siamak Shakeri, Emanuel Taropa, Paige Bailey, Zhifeng Chen, et~al. 2023.
\newblock \href {https://arxiv.org/abs/2305.10403} {Palm 2 technical report}.
\newblock \emph{ArXiv preprint}.

\bibitem[{Balasuriya et~al.(2009)Balasuriya, Ringland, Nothman, Murphy, and Curran}]{wiki-gold}
Dominic Balasuriya, Nicky Ringland, Joel Nothman, Tara Murphy, and James~R. Curran. 2009.
\newblock \href {https://aclanthology.org/W09-3302} {Named entity recognition in {W}ikipedia}.
\newblock In \emph{Proceedings of the 2009 Workshop on The People{'}s Web Meets {NLP}: Collaboratively Constructed Semantic Resources (People{'}s Web)}, pages 10--18.

\bibitem[{Carlini(2020)}]{llm-privacy}
Nicholas Carlini. 2020.
\newblock Privacy considerations in large language models.
\newblock \url{https://blog.research.google/2020/12/privacy-considerations-in-large.html}.

\bibitem[{Chia et~al.(2022)Chia, Bing, Poria, and Si}]{chia-etal-2022-relationprompt}
Yew~Ken Chia, Lidong Bing, Soujanya Poria, and Luo Si. 2022.
\newblock \href {https://doi.org/10.18653/v1/2022.findings-acl.5} {{R}elation{P}rompt: Leveraging prompts to generate synthetic data for zero-shot relation triplet extraction}.
\newblock In \emph{Findings of the Association for Computational Linguistics: ACL 2022}, pages 45--57, Dublin, Ireland. Association for Computational Linguistics.

\bibitem[{Chung et~al.(2023)Chung, Kamar, and Amershi}]{chung-etal-2023-increasing}
John Chung, Ece Kamar, and Saleema Amershi. 2023.
\newblock \href {https://doi.org/10.18653/v1/2023.acl-long.34} {Increasing diversity while maintaining accuracy: Text data generation with large language models and human interventions}.
\newblock In \emph{Proceedings of the 61st Annual Meeting of the Association for Computational Linguistics (Volume 1: Long Papers)}, pages 575--593, Toronto, Canada. Association for Computational Linguistics.

\bibitem[{Devlin et~al.(2019)Devlin, Chang, Lee, and Toutanova}]{bert}
Jacob Devlin, Ming-Wei Chang, Kenton Lee, and Kristina Toutanova. 2019.
\newblock \href {https://doi.org/10.18653/v1/N19-1423} {{BERT}: Pre-training of deep bidirectional transformers for language understanding}.
\newblock In \emph{Proceedings of the 2019 Conference of the North {A}merican Chapter of the Association for Computational Linguistics: Human Language Technologies, Volume 1 (Long and Short Papers)}, pages 4171--4186.

\bibitem[{Guo and Roth(2021)}]{guo-roth-2021-constrained}
Ruohao Guo and Dan Roth. 2021.
\newblock \href {https://doi.org/10.18653/v1/2021.findings-acl.396} {Constrained labeled data generation for low-resource named entity recognition}.
\newblock In \emph{Findings of the Association for Computational Linguistics: ACL-IJCNLP 2021}, pages 4519--4533, Online. Association for Computational Linguistics.

\bibitem[{Gupta et~al.(2023)Gupta, Scaria, Anantheswaran, Verma, Parmar, Sawant, Mishra, and Baral}]{gupta2023targen}
Himanshu Gupta, Kevin Scaria, Ujjwala Anantheswaran, Shreyas Verma, Mihir Parmar, Saurabh~Arjun Sawant, Swaroop Mishra, and Chitta Baral. 2023.
\newblock \href {https://arxiv.org/abs/2310.17876} {Tar{GEN}: Targeted data generation with large language models}.
\newblock \emph{ArXiv preprint}.

\bibitem[{He et~al.(2023)He, Gao, and Chen}]{deberta-v3}
Pengcheng He, Jianfeng Gao, and Weizhu Chen. 2023.
\newblock \href {https://openreview.net/forum?id=sE7-XhLxHA} {De{BERT}av3: Improving {D}e{BERT}a using {ELECTRA}-style pre-training with gradient-disentangled embedding sharing}.
\newblock In \emph{The Eleventh International Conference on Learning Representations}.

\bibitem[{He et~al.(2021)He, Liu, Gao, and Chen}]{he2021deberta}
Pengcheng He, Xiaodong Liu, Jianfeng Gao, and Weizhu Chen. 2021.
\newblock \href {https://openreview.net/forum?id=XPZIaotutsD} {Deberta: decoding-enhanced bert with disentangled attention}.
\newblock In \emph{9th International Conference on Learning Representations, {ICLR} 2021, Virtual Event, Austria, May 3-7, 2021}.

\bibitem[{Hiebel et~al.(2023)Hiebel, Ferret, Fort, and N{\'e}v{\'e}ol}]{hiebel-etal-2023-synthetic}
Nicolas Hiebel, Olivier Ferret, Karen Fort, and Aur{\'e}lie N{\'e}v{\'e}ol. 2023.
\newblock \href {https://doi.org/10.18653/v1/2023.eacl-main.170} {Can synthetic text help clinical named entity recognition? a study of electronic health records in {F}rench}.
\newblock In \emph{Proceedings of the 17th Conference of the European Chapter of the Association for Computational Linguistics}, pages 2320--2338, Dubrovnik, Croatia. Association for Computational Linguistics.

\bibitem[{Huang et~al.(2021)Huang, Li, Subudhi, Jose, Balakrishnan, Chen, Peng, Gao, and Han}]{huang-etal-2021-shot}
Jiaxin Huang, Chunyuan Li, Krishan Subudhi, Damien Jose, Shobana Balakrishnan, Weizhu Chen, Baolin Peng, Jianfeng Gao, and Jiawei Han. 2021.
\newblock \href {https://doi.org/10.18653/v1/2021.emnlp-main.813} {Few-shot named entity recognition: An empirical baseline study}.
\newblock In \emph{Proceedings of the 2021 Conference on Empirical Methods in Natural Language Processing}, pages 10408--10423, Online and Punta Cana, Dominican Republic. Association for Computational Linguistics.

\bibitem[{Josifoski et~al.(2023)Josifoski, Sakota, Peyrard, and West}]{josifoski-etal-2023-exploiting}
Martin Josifoski, Marija Sakota, Maxime Peyrard, and Robert West. 2023.
\newblock \href {https://doi.org/10.18653/v1/2023.emnlp-main.96} {Exploiting asymmetry for synthetic training data generation: {S}ynth{IE} and the case of information extraction}.
\newblock In \emph{Proceedings of the 2023 Conference on Empirical Methods in Natural Language Processing}, pages 1555--1574.

\bibitem[{Kim et~al.(2022)Kim, Yoo, Yoon, Lee, and Kang}]{ge-ner}
Hyunjae Kim, Jaehyo Yoo, Seunghyun Yoon, Jinhyuk Lee, and Jaewoo Kang. 2022.
\newblock \href {https://aclanthology.org/2022.emnlp-main.417} {Simple questions generate named entity recognition datasets}.
\newblock In \emph{Proceedings of the 2022 Conference on Empirical Methods in Natural Language Processing}, pages 6220--6236.

\bibitem[{Li et~al.(2023{\natexlab{a}})Li, Fang, Yang, Wang, Ye, Zhao, and Zhang}]{Li2023EvaluatingCI}
Bo~Li, Gexiang Fang, Yang Yang, Quansen Wang, Wei Ye, Wen Zhao, and Shikun Zhang. 2023{\natexlab{a}}.
\newblock \href {https://arxiv.org/abs/2304.11633} {Evaluating chatgpt's information extraction capabilities: An assessment of performance, explainability, calibration, and faithfulness}.
\newblock \emph{ArXiv preprint}.

\bibitem[{Li et~al.(2023{\natexlab{b}})Li, Hammoud, Itani, Khizbullin, and Ghanem}]{li2023camel}
Guohao Li, Hasan Abed Al~Kader Hammoud, Hani Itani, Dmitrii Khizbullin, and Bernard Ghanem. 2023{\natexlab{b}}.
\newblock \href {http://arxiv.org/abs/2303.17760} {Camel: Communicative agents for "mind" exploration of large language model society}.

\bibitem[{Li et~al.(2021)Li, Ding, Shang, McAuley, and Feng}]{drop-misc}
Jiacheng Li, Haibo Ding, Jingbo Shang, Julian McAuley, and Zhe Feng. 2021.
\newblock \href {https://doi.org/10.18653/v1/2021.acl-long.352} {Weakly supervised named entity tagging with learnable logical rules}.
\newblock In \emph{Proceedings of the 59th Annual Meeting of the Association for Computational Linguistics and the 11th International Joint Conference on Natural Language Processing (Volume 1: Long Papers)}, pages 4568--4581.

\bibitem[{Li et~al.(2024)Li, Ramprasad, and Zhang}]{li2024gno}
Yinghao Li, Rampi Ramprasad, and Chao Zhang. 2024.
\newblock \href {http://arxiv.org/abs/2402.13364} {A simple but effective approach to improve structured language model output for information extraction}.

\bibitem[{Li et~al.(2023{\natexlab{c}})Li, Zhu, Lu, and Yin}]{li-etal-2023-synthetic}
Zhuoyan Li, Hangxiao Zhu, Zhuoran Lu, and Ming Yin. 2023{\natexlab{c}}.
\newblock \href {https://doi.org/10.18653/v1/2023.emnlp-main.647} {Synthetic data generation with large language models for text classification: Potential and limitations}.
\newblock In \emph{Proceedings of the 2023 Conference on Empirical Methods in Natural Language Processing}, pages 10443--10461.

\bibitem[{Liu et~al.(2022{\natexlab{a}})Liu, Swayamdipta, Smith, and Choi}]{liu-etal-2022-wanli}
Alisa Liu, Swabha Swayamdipta, Noah~A. Smith, and Yejin Choi. 2022{\natexlab{a}}.
\newblock \href {https://doi.org/10.18653/v1/2022.findings-emnlp.508} {{WANLI}: Worker and {AI} collaboration for natural language inference dataset creation}.
\newblock In \emph{Findings of the Association for Computational Linguistics: EMNLP 2022}, pages 6826--6847, Abu Dhabi, United Arab Emirates. Association for Computational Linguistics.

\bibitem[{Liu et~al.(2022{\natexlab{b}})Liu, Chen, and Xu}]{prompt-da}
Jian Liu, Yufeng Chen, and Jinan Xu. 2022{\natexlab{b}}.
\newblock \href {https://doi.org/10.24963/ijcai.2022/590} {Low-resource ner by data augmentation with prompting}.
\newblock In \emph{Proceedings of the Thirty-First International Joint Conference on Artificial Intelligence, {IJCAI-22}}, pages 4252--4258. International Joint Conferences on Artificial Intelligence Organization.
\newblock Main Track.

\bibitem[{Liu et~al.(2013)Liu, Pasupat, Cyphers, and Glass}]{asgard}
Jingjing Liu, Panupong Pasupat, Scott Cyphers, and Jim Glass. 2013.
\newblock \href {https://doi.org/10.1109/ICASSP.2013.6639301} {Asgard: A portable architecture for multilingual dialogue systems}.
\newblock In \emph{2013 IEEE International Conference on Acoustics, Speech and Signal Processing}, pages 8386--8390.

\bibitem[{Liu et~al.(2019)Liu, Ott, Goyal, Du, Joshi, Chen, Levy, Lewis, Zettlemoyer, and Stoyanov}]{roberta}
Yinhan Liu, Myle Ott, Naman Goyal, Jingfei Du, Mandar Joshi, Danqi Chen, Omer Levy, Mike Lewis, Luke Zettlemoyer, and Veselin Stoyanov. 2019.
\newblock \href {http://arxiv.org/abs/1907.11692} {Roberta: A robustly optimized bert pretraining approach}.

\bibitem[{Loshchilov and Hutter(2019)}]{adam-w}
Ilya Loshchilov and Frank Hutter. 2019.
\newblock \href {https://openreview.net/forum?id=Bkg6RiCqY7} {Decoupled weight decay regularization}.
\newblock In \emph{7th International Conference on Learning Representations, {ICLR} 2019, New Orleans, LA, USA, May 6-9, 2019}.

\bibitem[{Ma et~al.(2022)Ma, Zhou, Gui, Tan, Li, Zhang, and Huang}]{few-shot-ner}
Ruotian Ma, Xin Zhou, Tao Gui, Yiding Tan, Linyang Li, Qi~Zhang, and Xuanjing Huang. 2022.
\newblock \href {https://doi.org/10.18653/v1/2022.naacl-main.420} {Template-free prompt tuning for few-shot {NER}}.
\newblock In \emph{Proceedings of the 2022 Conference of the North American Chapter of the Association for Computational Linguistics: Human Language Technologies}, pages 5721--5732.

\bibitem[{Madaan et~al.(2023)Madaan, Tandon, Gupta, Hallinan, Gao, Wiegreffe, Alon, Dziri, Prabhumoye, Yang, Gupta, Majumder, Hermann, Welleck, Yazdanbakhsh, and Clark}]{self-refine}
Aman Madaan, Niket Tandon, Prakhar Gupta, Skyler Hallinan, Luyu Gao, Sarah Wiegreffe, Uri Alon, Nouha Dziri, Shrimai Prabhumoye, Yiming Yang, Shashank Gupta, Bodhisattwa~Prasad Majumder, Katherine Hermann, Sean Welleck, Amir Yazdanbakhsh, and Peter Clark. 2023.
\newblock \href {https://openreview.net/forum?id=S37hOerQLB} {Self-refine: Iterative refinement with self-feedback}.
\newblock In \emph{Thirty-seventh Conference on Neural Information Processing Systems}.

\bibitem[{Manning(2006)}]{dont-optimize-f1}
Chris Manning. 2006.
\newblock Doing named entity recognition? don't optimize for f1.
\newblock \url{https://nlpers.blogspot.com/2006/08/doing-named-entity-recognition-dont.html}.
\newblock Blog post.

\bibitem[{Meng et~al.(2022)Meng, Huang, Zhang, and Han}]{supergen}
Yu~Meng, Jiaxin Huang, Yu~Zhang, and Jiawei Han. 2022.
\newblock \href {https://openreview.net/forum?id=4G1Sfp_1sz7} {Generating training data with language models: Towards zero-shot language understanding}.
\newblock In \emph{Advances in Neural Information Processing Systems}.

\bibitem[{Meng et~al.(2023)Meng, Michalski, Huang, Zhang, Abdelzaher, and Han}]{meng2023tuning}
Yu~Meng, Martin Michalski, Jiaxin Huang, Yu~Zhang, Tarek Abdelzaher, and Jiawei Han. 2023.
\newblock \href {https://proceedings.mlr.press/v202/meng23b.html} {Tuning language models as training data generators for augmentation-enhanced few-shot learning}.
\newblock In \emph{Proceedings of the 40th International Conference on Machine Learning}, volume 202 of \emph{Proceedings of Machine Learning Research}, pages 24457--24477. PMLR.

\bibitem[{Meng et~al.(2021)Meng, Zhang, Huang, Wang, Zhang, Ji, and Han}]{wikigold-split}
Yu~Meng, Yunyi Zhang, Jiaxin Huang, Xuan Wang, Yu~Zhang, Heng Ji, and Jiawei Han. 2021.
\newblock \href {https://doi.org/10.18653/v1/2021.emnlp-main.810} {Distantly-supervised named entity recognition with noise-robust learning and language model augmented self-training}.
\newblock In \emph{Proceedings of the 2021 Conference on Empirical Methods in Natural Language Processing}, pages 10367--10378, Online and Punta Cana, Dominican Republic. Association for Computational Linguistics.

\bibitem[{OpenAI(2023)}]{openai2023gpt4}
OpenAI. 2023.
\newblock \href {http://arxiv.org/abs/2303.08774} {Gpt-4 technical report}.

\bibitem[{Ouyang et~al.(2022)Ouyang, Wu, Jiang, Almeida, Wainwright, Mishkin, Zhang, Agarwal, Slama, Ray, Schulman, Hilton, Kelton, Miller, Simens, Askell, Welinder, Christiano, Leike, and Lowe}]{instruct-gpt}
Long Ouyang, Jeffrey Wu, Xu~Jiang, Diogo Almeida, Carroll Wainwright, Pamela Mishkin, Chong Zhang, Sandhini Agarwal, Katarina Slama, Alex Ray, John Schulman, Jacob Hilton, Fraser Kelton, Luke Miller, Maddie Simens, Amanda Askell, Peter Welinder, Paul~F Christiano, Jan Leike, and Ryan Lowe. 2022.
\newblock \href {https://proceedings.neurips.cc/paper_files/paper/2022/file/b1efde53be364a73914f58805a001731-Paper-Conference.pdf} {Training language models to follow instructions with human feedback}.
\newblock In \emph{Advances in Neural Information Processing Systems}, pages 27730--27744.

\bibitem[{Segura-Bedmar et~al.(2013)Segura-Bedmar, Mart{\'\i}nez, and Herrero-Zazo}]{semeval-2013}
Isabel Segura-Bedmar, Paloma Mart{\'\i}nez, and Mar{\'\i}a Herrero-Zazo. 2013.
\newblock \href {https://aclanthology.org/S13-2056} {{S}em{E}val-2013 task 9 : Extraction of drug-drug interactions from biomedical texts ({DDIE}xtraction 2013)}.
\newblock In \emph{Second Joint Conference on Lexical and Computational Semantics (*{SEM}), Volume 2: Proceedings of the Seventh International Workshop on Semantic Evaluation ({S}em{E}val 2013)}, pages 341--350, Atlanta, Georgia, USA. Association for Computational Linguistics.

\bibitem[{Shinn et~al.(2023)Shinn, Cassano, Gopinath, Narasimhan, and Yao}]{reflexion}
Noah Shinn, Federico Cassano, Ashwin Gopinath, Karthik~R Narasimhan, and Shunyu Yao. 2023.
\newblock \href {https://openreview.net/forum?id=vAElhFcKW6} {Reflexion: language agents with verbal reinforcement learning}.
\newblock In \emph{Thirty-seventh Conference on Neural Information Processing Systems}.

\bibitem[{Tang et~al.(2023)Tang, Han, Jiang, and Hu}]{tang2023does}
Ruixiang Tang, Xiaotian Han, Xiaoqian Jiang, and Xia Hu. 2023.
\newblock \href {https://arxiv.org/abs/2303.04360} {Does synthetic data generation of llms help clinical text mining?}
\newblock \emph{ArXiv preprint}.

\bibitem[{Thakkar et~al.(2023)Thakkar, Bolukbasi, Ganapathy, Vashishth, Chandar, and Talukdar}]{self-influence}
Megh Thakkar, Tolga Bolukbasi, Sriram Ganapathy, Shikhar Vashishth, Sarath Chandar, and Partha Talukdar. 2023.
\newblock \href {https://doi.org/10.18653/v1/2023.emnlp-main.125} {Self-influence guided data reweighting for language model pre-training}.
\newblock In \emph{Proceedings of the 2023 Conference on Empirical Methods in Natural Language Processing}, pages 2033--2045, Singapore. Association for Computational Linguistics.

\bibitem[{Tjong Kim~Sang and De~Meulder(2003)}]{conll-2003}
Erik~F. Tjong Kim~Sang and Fien De~Meulder. 2003.
\newblock \href {https://aclanthology.org/W03-0419} {Introduction to the {C}o{NLL}-2003 shared task: Language-independent named entity recognition}.
\newblock In \emph{Proceedings of the Seventh Conference on Natural Language Learning at {HLT}-{NAACL} 2003}, pages 142--147.

\bibitem[{Touvron et~al.(2023)Touvron, Lavril, Izacard, Martinet, Lachaux, Lacroix, Rozi{\`e}re, Goyal, Hambro, Azhar et~al.}]{touvron2023llama}
Hugo Touvron, Thibaut Lavril, Gautier Izacard, Xavier Martinet, Marie-Anne Lachaux, Timoth{\'e}e Lacroix, Baptiste Rozi{\`e}re, Naman Goyal, Eric Hambro, Faisal Azhar, et~al. 2023.
\newblock \href {https://arxiv.org/abs/2302.13971} {Llama: Open and efficient foundation language models}.
\newblock \emph{ArXiv preprint}.

\bibitem[{Wang et~al.(2023{\natexlab{a}})Wang, Sun, Li, Ouyang, Wu, Zhang, Li, and Wang}]{gpt-ner}
Shuhe Wang, Xiaofei Sun, Xiaoya Li, Rongbin Ouyang, Fei Wu, Tianwei Zhang, Jiwei Li, and Guoyin Wang. 2023{\natexlab{a}}.
\newblock \href {http://arxiv.org/abs/2304.10428} {{GPT}-{NER}: Named entity recognition via large language models}.

\bibitem[{Wang et~al.(2023{\natexlab{b}})Wang, Wei, Schuurmans, Le, Chi, Narang, Chowdhery, and Zhou}]{self-consistency}
Xuezhi Wang, Jason Wei, Dale Schuurmans, Quoc~V Le, Ed~H. Chi, Sharan Narang, Aakanksha Chowdhery, and Denny Zhou. 2023{\natexlab{b}}.
\newblock \href {https://openreview.net/forum?id=1PL1NIMMrw} {Self-consistency improves chain of thought reasoning in language models}.
\newblock In \emph{The Eleventh International Conference on Learning Representations}.

\bibitem[{Wang et~al.(2021)Wang, Chu, Zhang, and Gao}]{wang-etal-2021-learning-language-description}
Yaqing Wang, Haoda Chu, Chao Zhang, and Jing Gao. 2021.
\newblock \href {https://doi.org/10.18653/v1/2021.findings-emnlp.139} {Learning from language description: Low-shot named entity recognition via decomposed framework}.
\newblock In \emph{Findings of the Association for Computational Linguistics: EMNLP 2021}, pages 1618--1630, Punta Cana, Dominican Republic. Association for Computational Linguistics.

\bibitem[{Wang et~al.(2023{\natexlab{c}})Wang, Kordi, Mishra, Liu, Smith, Khashabi, and Hajishirzi}]{wang2023selfinstruct}
Yizhong Wang, Yeganeh Kordi, Swaroop Mishra, Alisa Liu, Noah~A. Smith, Daniel Khashabi, and Hannaneh Hajishirzi. 2023{\natexlab{c}}.
\newblock \href {http://arxiv.org/abs/2212.10560} {Self-instruct: Aligning language models with self-generated instructions}.

\bibitem[{Wei et~al.(2022)Wei, Wang, Schuurmans, Bosma, brian ichter, Xia, Chi, Le, and Zhou}]{cot}
Jason Wei, Xuezhi Wang, Dale Schuurmans, Maarten Bosma, brian ichter, Fei Xia, Ed~H. Chi, Quoc~V Le, and Denny Zhou. 2022.
\newblock \href {https://openreview.net/forum?id=_VjQlMeSB_J} {Chain of thought prompting elicits reasoning in large language models}.
\newblock In \emph{Advances in Neural Information Processing Systems}.

\bibitem[{Wei et~al.(2023)Wei, Cui, Cheng, Wang, Zhang, Huang, Xie, Xu, Chen, Zhang, Jiang, and Han}]{Wei2023ZeroShotIE}
Xiang Wei, Xingyu Cui, Ning Cheng, Xiaobin Wang, Xin Zhang, Shen Huang, Pengjun Xie, Jinan Xu, Yufeng Chen, Meishan Zhang, Yong Jiang, and Wenjuan Han. 2023.
\newblock \href {https://arxiv.org/abs/2302.10205} {Zero-shot information extraction via chatting with chatgpt}.
\newblock \emph{ArXiv preprint}.

\bibitem[{Xie et~al.(2023)Xie, Li, Zhang, Zhang, Liu, and Wang}]{Xie.2023.empirical}
Tingyu Xie, Qi~Li, Jian Zhang, Yan Zhang, Zuozhu Liu, and Hongwei Wang. 2023.
\newblock \href {https://aclanthology.org/2023.emnlp-main.493} {Empirical study of zero-shot {NER} with chatgpt}.
\newblock In \emph{Proceedings of the 2023 Conference on Empirical Methods in Natural Language Processing, {EMNLP} 2023, Singapore, December 6-10, 2023}, pages 7935--7956. Association for Computational Linguistics.

\bibitem[{Xu et~al.(2023{\natexlab{a}})Xu, Wang, Lyu, Dai, Zhang, and Mao}]{xu-etal-2023-s2ynre}
Benfeng Xu, Quan Wang, Yajuan Lyu, Dai Dai, Yongdong Zhang, and Zhendong Mao. 2023{\natexlab{a}}.
\newblock \href {https://doi.org/10.18653/v1/2023.acl-long.455} {{S}2yn{RE}: Two-stage self-training with synthetic data for low-resource relation extraction}.
\newblock In \emph{Proceedings of the 61st Annual Meeting of the Association for Computational Linguistics (Volume 1: Long Papers)}, pages 8186--8207, Toronto, Canada. Association for Computational Linguistics.

\bibitem[{Xu et~al.(2023{\natexlab{b}})Xu, Guo, Duan, and McAuley}]{xu2023baize}
Canwen Xu, Daya Guo, Nan Duan, and Julian McAuley. 2023{\natexlab{b}}.
\newblock \href {http://arxiv.org/abs/2304.01196} {Baize: An open-source chat model with parameter-efficient tuning on self-chat data}.

\bibitem[{Xu et~al.(2023{\natexlab{c}})Xu, Cui, Yu, Kan, Shi, Zhuang, Jin, Ho, and Yang}]{xu2023knowledge}
Ran Xu, Hejie Cui, Yue Yu, Xuan Kan, Wenqi Shi, Yuchen Zhuang, Wei Jin, Joyce Ho, and Carl Yang. 2023{\natexlab{c}}.
\newblock \href {https://arxiv.org/abs/2311.00287} {Knowledge-infused prompting: Assessing and advancing clinical text data generation with large language models}.
\newblock \emph{ArXiv preprint}.

\bibitem[{Ye et~al.(2022{\natexlab{a}})Ye, Gao, Li, Xu, Feng, Wu, Yu, and Kong}]{ye-etal-2022-zerogen}
Jiacheng Ye, Jiahui Gao, Qintong Li, Hang Xu, Jiangtao Feng, Zhiyong Wu, Tao Yu, and Lingpeng Kong. 2022{\natexlab{a}}.
\newblock \href {https://doi.org/10.18653/v1/2022.emnlp-main.801} {{Z}ero{G}en: Efficient zero-shot learning via dataset generation}.
\newblock In \emph{Proceedings of the 2022 Conference on Empirical Methods in Natural Language Processing}, pages 11653--11669, Abu Dhabi, United Arab Emirates. Association for Computational Linguistics.

\bibitem[{Ye et~al.(2022{\natexlab{b}})Ye, Gao, Wu, Feng, Yu, and Kong}]{ye-etal-2022-progen}
Jiacheng Ye, Jiahui Gao, Zhiyong Wu, Jiangtao Feng, Tao Yu, and Lingpeng Kong. 2022{\natexlab{b}}.
\newblock \href {https://doi.org/10.18653/v1/2022.findings-emnlp.269} {{P}ro{G}en: Progressive zero-shot dataset generation via in-context feedback}.
\newblock In \emph{Findings of the Association for Computational Linguistics: EMNLP 2022}, pages 3671--3683, Abu Dhabi, United Arab Emirates. Association for Computational Linguistics.

\bibitem[{Ye et~al.(2024)Ye, Xu, Wang, Zhou, Zhang, Gui, and Huang}]{llm-da}
Junjie Ye, Nuo Xu, Yikun Wang, Jie Zhou, Qi~Zhang, Tao Gui, and Xuanjing Huang. 2024.
\newblock \href {http://arxiv.org/abs/2402.14568} {Llm-da: Data augmentation via large language models for few-shot named entity recognition}.

\bibitem[{Yu et~al.(2023{\natexlab{a}})Yu, Zhuang, Zhang, Meng, Ratner, Krishna, Shen, and Zhang}]{attr-prompt}
Yue Yu, Yuchen Zhuang, Jieyu Zhang, Yu~Meng, Alexander Ratner, Ranjay Krishna, Jiaming Shen, and Chao Zhang. 2023{\natexlab{a}}.
\newblock \href {https://openreview.net/forum?id=6hZIfAY9GD} {Large language model as attributed training data generator: A tale of diversity and bias}.
\newblock In \emph{Thirty-seventh Conference on Neural Information Processing Systems Datasets and Benchmarks Track}.

\bibitem[{Yu et~al.(2023{\natexlab{b}})Yu, Zhuang, Zhang, Meng, Shen, and Zhang}]{yu-etal-2023-regen}
Yue Yu, Yuchen Zhuang, Rongzhi Zhang, Yu~Meng, Jiaming Shen, and Chao Zhang. 2023{\natexlab{b}}.
\newblock \href {https://doi.org/10.18653/v1/2023.findings-acl.748} {{R}e{G}en: Zero-shot text classification via training data generation with progressive dense retrieval}.
\newblock In \emph{Findings of the Association for Computational Linguistics: ACL 2023}, pages 11782--11805, Toronto, Canada. Association for Computational Linguistics.

\bibitem[{Zhang et~al.(2022{\natexlab{a}})Zhang, West, Cui, and Zhang}]{zhang2022adaptive}
Rongzhi Zhang, Rebecca West, Xiquan Cui, and Chao Zhang. 2022{\natexlab{a}}.
\newblock Adaptive multi-view rule discovery for weakly-supervised compatible products prediction.
\newblock In \emph{Proceedings of the 28th ACM SIGKDD Conference on Knowledge Discovery and Data Mining}, pages 4521--4529.

\bibitem[{Zhang et~al.(2022{\natexlab{b}})Zhang, Yu, Shetty, Song, and Zhang}]{zhang2022prboost}
Rongzhi Zhang, Yue Yu, Pranav Shetty, Le~Song, and Chao Zhang. 2022{\natexlab{b}}.
\newblock Prboost: Prompt-based rule discovery and boosting for interactive weakly-supervised learning.
\newblock \emph{arXiv preprint arXiv:2203.09735}.

\bibitem[{Zhou et~al.(2023)Zhou, Sch{\"a}rli, Hou, Wei, Scales, Wang, Schuurmans, Cui, Bousquet, Le, and Chi}]{least2most}
Denny Zhou, Nathanael Sch{\"a}rli, Le~Hou, Jason Wei, Nathan Scales, Xuezhi Wang, Dale Schuurmans, Claire Cui, Olivier Bousquet, Quoc~V Le, and Ed~H. Chi. 2023.
\newblock \href {https://openreview.net/forum?id=WZH7099tgfM} {Least-to-most prompting enables complex reasoning in large language models}.
\newblock In \emph{The Eleventh International Conference on Learning Representations}.

\bibitem[{Zhou et~al.(2022)Zhou, Li, He, Bing, Cambria, Si, and Miao}]{zhou-etal-2022-melm}
Ran Zhou, Xin Li, Ruidan He, Lidong Bing, Erik Cambria, Luo Si, and Chunyan Miao. 2022.
\newblock \href {https://doi.org/10.18653/v1/2022.acl-long.160} {{MELM}: Data augmentation with masked entity language modeling for low-resource {NER}}.
\newblock In \emph{Proceedings of the 60th Annual Meeting of the Association for Computational Linguistics (Volume 1: Long Papers)}, pages 2251--2262, Dublin, Ireland. Association for Computational Linguistics.

\bibitem[{Zhou et~al.(2024)Zhou, Zhang, Gu, Chen, and Poon}]{universal-ner}
Wenxuan Zhou, Sheng Zhang, Yu~Gu, Muhao Chen, and Hoifung Poon. 2024.
\newblock \href {https://openreview.net/forum?id=r65xfUb76p} {Universal{NER}: Targeted distillation from large language models for open named entity recognition}.
\newblock In \emph{The Twelfth International Conference on Learning Representations}.

\end{thebibliography}

\newpage
\clearpage
\appendix
\label{sec:appendix}

\section{Additional Analysis}
\label{apd:analysis}

\begin{table*}[tb]
    \centering 
    \resizebox{0.70\linewidth}{!}{
    \begin{NiceTabular}{ l l *{5}{c}}[colortbl-like]
        \toprule
        \multicolumn{2}{c}{\thead{Method}} & \thead{CoNLL\\-2003} & \thead{Wiki\\Gold} & \thead{MIT-\\Movie} & \thead{MIT-\\Restaurant} & \thead{Average} \\
        
        \midrule
        \multicolumn{2}{c}{Simple Prompt} & 72.0 / 75.6 & 75.8 / 77.9 & 64.8 / 76.7 & 58.4 / 73.5 & 67.8 / 75.9 \\
        
        \midrule
        \multirow{8}{*}[-0.25em]{ProgGen} 
            & Diversify X 
                & 69.2 / 72.5 & 75.5 / 80.2 & 65.0 / 73.9 & 54.7 / 72.3 & 66.1 / 74.7 \\
            \rowcolor{lightCyan} \cellcolor{white} & \quad + Self-Correction 
                & 71.9 / 74.6 &77.3 / 80.3 & \nCrt{63.5} / \nCrt{73.5} & 56.1 / 73.1 & 67.2 / 75.4 \\ 
            
            & Diversify Y (vanilla) 
                & 71.2 / 73.7 & 76.2 / 81.4 & 73.0 / 80.9 & 61.5 / 78.2 & 70.5 / 78.6 \\
            \rowcolor{lightCyan} \cellcolor{white} & \quad + Self-Correction
                & 71.4 / 74.3 & \textbf{78.2} / \textbf{82.1} & \textbf{73.3} / \textbf{81.2} & \textbf{62.8} / \textbf{78.8} & 71.4 / 79.1 \\ 
            
            & Diversify Y (latent) 
                & 75.6 / 78.1 & 75.7 / 79.2 & 71.8 / 80.0  & 60.9 / 78.2 & 71.0 / 78.9 \\
            \rowcolor{lightCyan} \cellcolor{white} & \quad + Self-Correction 
                & \textbf{77.5} / \textbf{79.7} & 76.4 / 80.1 & 72.9 / \textbf{81.2} & 61.4 / 78.0 & \textbf{72.1} / \textbf{79.8} \\
            
            & Diversify X + Y 
                & 73.7 / 75.4 & 76.1 / 80.2 & 70.6 / 79.9 & 59.5 / 76.5 & 70.0 / 78.0 \\
            \rowcolor{lightCyan} \cellcolor{white} & \quad + Self-Correction
                & 74.3 / 76.3 & \nCrt{74.6} / \nCrt{78.7} & 72.2 / 79.9 & 59.8 / 76.9 & 70.2 / 78.0 \\
        
        \midrule
        \multicolumn{2}{c}{Few-shot ICL} & 71.5 / 73.8 & 74.5 / 79.2 & 74.2 / 84.9 & 67.0 / 81.6 & 71.8 / 79.9 \\
        \bottomrule
    \end{NiceTabular}
    }
    \caption{
        ProgGen Performance Comparison. We show F1 and partial F1 score separated by slash (F1 / partial F1). 
    }
    \label{tbl:main-results-partial}
\end{table*}

\subsection{Partial F1 Evaluation}
\label{apd:partial-f1-eval}

We note that f1 only reward exact boundary matches of identified named entities \citep{dont-optimize-f1} 
whereas partial matches may still benefit users of NER systems. 
As such, we also report the partial f1 scores following prior work \citep{universal-ner}.
We use the setup as described in \citet{semeval-2013}, where 
predicted named entities that overlaps with ground truth named entities of the same type are counted as 
0.5 true positive\footnote{We used the implementation from the \texttt{nervaluate} package available at \url{https://github.com/davidsbatista/NER-Evaluation}}. 
We report partial F1 scores for the main experiments (\cref{sec:experiment-results}) in Table \ref{tbl:main-results-partial}. 
We observe that F1 and partial F1 scores share similar trends where we can make similar conclusions as in \cref{sec:experiment-results}. 
We further note that 
(1) the gap between exact F1 and partial F1 scores are greater for specific domains (\textasciitilde{}10\% vs \textasciitilde{}3\% for general domains), 
and thus
(2) partial F1 closes the performance gap between general-domain- and specific-domain-datasets, 
indicating that annotating the precise entity span boundaries is more challenging in specific domains,
corroborating with manual inspection (Appendix \ref{apd:representative-annotatino-errors}). 

\begin{table}[tb]
    \centering 
    \resizebox{0.75\linewidth}{!}{
    \begin{tabular}{ *5{c} }
        \toprule
        \thead{CoNLL\\-2003} & \thead{Wiki\\Gold} & \thead{MIT-\\Movie} & \thead{MIT-\\Restaurant} & \thead{Average} \\
        
        \midrule
        94.2 & 88.0 & 87.8 & 79.9 & 88.6 \\

        \bottomrule
    \end{tabular}
    }
    \caption{Supervised Learning results (F1).}
    \label{tbl:supervised-learning}
\end{table}

\subsection{Full Supervised Learning}
\label{apd:supervised-learning}
As reference, we show standard supervised learning results using the same training pipeline as in \cref{sec:experiment-setup} and our de-duplicated version of the original NER datasets (Appendix \ref{apd:ner-sample-processing}) in Table \ref{tbl:supervised-learning}.
We report F1 on the model with best dev loss.

\begin{table}[tb]
    \centering 
    \resizebox{1\columnwidth}{!}{

        
        
            
            
            


    \begin{tabular}{ l *5{c} }
        \toprule
        \thead{Method} & \thead{CoNLL\\-2003} & \thead{Wiki\\Gold} & \thead{MIT-\\Movie} & \thead{MIT-\\Rest.} & \thead{Avg.} \\
        
        \midrule
        Simple Prompt & 69.5 & 72.2 & 67.0 & 56.1 & 66.2 \\
        
        \midrule
            Diversify X & 73.2 \ng{+3.7} & 76.4 \ng{+4.2} & 70.2 \ng{+3.2} & 55.9 \nr{-0.2} & 68.9 \ng{+2.7}  \\
            
            Diversify Y (vanilla) & 72.1 \ng{+2.6} & 76.1 \ng{+3.9} & \textbf{71.6 \ng{+4.6}} & \textbf{61.5 \ng{+5.4}} & 70.3 \ng{+4.1} \\
            
            Diversify Y (latent) 
                & 73.4 \ng{+3.9} & \textbf{77.9 \ng{+5.7}} & 70.5 \ng{+3.5}  & 60.6 \ng{+4.5} & \textbf{70.6 \ng{+4.4}} \\
            
            Diversify X + Y 
                & \textbf{76.4 \ng{+6.9}} & 72.9 \ng{+0.7} & 66.2 \nr{-0.8} & 58.7 \ng{+2.6} & 68.6 \ng{+2.4} \\

        \bottomrule
    \end{tabular}
    }
    \caption{Early Data Generation Performance Comparison (F1).}
    \label{tbl:1-stage-early-results}
\end{table}

\begin{table}[tb]
    \centering 
    \resizebox{0.85\linewidth}{!}{
    \begin{tabular}{ c *5{c} }
        \toprule
        \thead{$L$} & \thead{CoNLL\\-2003} & \thead{Wiki\\Gold} & \thead{MIT-\\Movie} & \thead{MIT-\\Restaurant} & \thead{Average} \\
        
        \midrule
        10 & \textbf{74.4} $\pm$ 0.5 & \textbf{76.7} $\pm$ 1.0 & 61.9 $\pm$ 3.0 & \textbf{58.1} $\pm$ 1.9 & \textbf{67.8} $\pm$ 1.6 \\
        3 & 71.2 $\pm$ 0.8 & 74.6 $\pm$ 0.6 & \textbf{66.7} $\pm$ 0.5 & 57.8 $\pm$ 0.9 & 67.6 $\pm$ 1.0 \\

        \bottomrule
    \end{tabular}
    }
    \caption{Simple Prompt stability comparison (F1). $L$ refers to the number of NER samples LLMs are instructed to generate in each prompt.}
    \label{tbl:sim-prompt-seed}
\end{table}

\subsection{OpenAI API Stability}
\label{apd:unstable-api-rerun}
We note that data generation via OpenAI's Chat Completion API may yield unstable results, with minor data generation pipeline edits. 
Unfortunately, OpenAI API did not support generation output seeding until Nov. 2023, where even seeding yield different results sometimes. 
Earlier NER data generation results 
yield slightly different F1 scores compared to that we report in \cref{sec:experiment-results}, as shown in Table \ref{tbl:1-stage-early-results}.
We attribute the discrepancy to OpenAI server changes (\texttt{system\_fingerprint})\footnote{\url{https://cookbook.openai.com/examples/reproducible_outputs_with_the_seed_parameter}}. 
We can observe similar trends except that in earlier experiments, Diversify X works better more frequently. 
The final conclusion remains the same: 
(1) diversity improves performance, where entity diversity contributes the most, and
(2) the \emph{latent} variant of diverse entity yield most robust gains across datasets. 

To further assess whether unstable OpenAI completions influence our findings, 
we re-run Simple Prompt 3 times with different random seeds (42, 43, 44) for 
prompt construction, OpenAI completion and downstream model training. 
We note that GPT3.5 frequently fail to generate $L = 50$ samples as instructed in prompt (may stop generation around 20 samples; see Appendix \ref{apd:data-gen-sample-gen-setup}), 
Thus, for the ease of seeding, we generated NER samples with $L \in [3, 10]$. 
We report the average F1 and standard deviation in Table \ref{tbl:sim-prompt-seed}. 
We observe that different $L$'s influence dataset-specific performances noticeably with fluctuations, 
while for both $L$ values, the average F1 scores stays around the same
w.r.t. the results we report (Tables \ref{tbl:main-results} and \ref{tbl:1-stage-early-results}). 
This confirms our findings that that ProgGen leads to noticeable gains over Simple Prompt.

\begin{table}[tb]
    \centering 
    \resizebox{0.95\columnwidth}{!}{
    \begin{tabular}{ m{0.33\columnwidth} l *{5}{c}}
        \toprule
        \thead{Method} & \thead{Gen.\\Kind} & \thead{CoNLL\\-2003} & \thead{Wiki\\Gold} & \thead{MIT-\\Movie} & \thead{MIT-\\Rest.} & \thead{Avg.} \\
        
        \midrule
        Simple Prompt & NER & 0.10 & 0.16 & 0.13 & 0.13 & 0.13 \\
        
        \midrule
        \multirow{5}{*}[-0em]{Diversify X}
            & Attr. Dim. & 0.09 & 0.14 & 0.11 & 0.13 & 0.12 \\
            & Attr. Val. & 0.00 & 0.00 & 0.00 & 0.00 & 0.00 \\ 
            & NER        & 0.24 & 0.33 & 0.36 & 0.31 & 0.31 \\
            & Correct.   & 0.09 & 0.11 & 0.14 & 0.24 & 0.15 \\
            \rowcolor{lightCyan} \cellcolor{white} 
            & \emph{Total} & 0.42 & 0.48 & 0.61 & 0.68 & 0.57 \\
        
        \midrule
        \multirow{4}{*}[-0em]{\makecell{Diversify Y\\(vanilla)}}
            & Entity     & 0.01 & 0.02 & 0.07 & 0.04 & 0.04 \\
            & NER        & 0.24 & 0.31 & 0.35 & 0.30 & 0.30 \\
            & Correct.   & 0.08 & 0.16 & 0.16 & 0,22 & 0.16 \\ 
            \rowcolor{lightCyan} \cellcolor{white} 
            & \emph{Total} & 0.33 & 0.49 & 0.58 & 0.56 & 0.49 \\
        
        \midrule
        \multirow{4}{*}[-0em]{\makecell{Diversify Y\\(latent)}}
            & Entity     & 0.18 & 0.23 & 0.27 & 0.08 & 0.19 \\
            & NER        & 0.24 & 0.32 & 0.35 & 0.30 & 0.30 \\
            & Correct.   & 0.09 & 0.15 & 0.17 & 0.22 & 0.16 \\ 
            \rowcolor{lightCyan} \cellcolor{white} 
            & \emph{Total} & 0.51 & 0.70 & 0.58 & 0.56 & 0.49 \\
            
        \midrule
        \multirow{4}{*}[-0em]{Diversify X + Y}
            & Divers.    & 0.27 & 0.37 & 0.38 & 0,21 & 0.31 \\
            & NER        & 0.27 & 0.35 & 0.39 & 0.33 & 0.34 \\
            & Correct.   & 0.08 & 0.13 & 0.15 & 0.26 & 0.16 \\ 
            \rowcolor{lightCyan} \cellcolor{white} 
            & \emph{Total} & 0.62 & 0.85 & 0.92 & 0.80 & 0.80 \\
        
        \midrule
        Few-shot ICL 
            & Pred. & 0.39 & 0.49 & 0.61 & 0.48 & 0.49 \\
        \bottomrule
    \end{tabular}
    }
    \caption{
        ProgGen API Token Querying Cost Comparison. 
        \emph{Gen. Kind} refers to the type of data generated and are ordered by the ProgGen steps performed, 
        where \emph{Attr. Dim.}, \emph{Attr. Val.}, \emph{NER} and \emph{Correct.} refer to generating attribute dimensions, attribute values, the entity pool, NER samples and entity corrections respectively. 
        For \emph{Diversify X + Y}, we sum the cost of all diversity requirement generation in \emph{Divers.} (values taken from \emph{Diversify X} and \emph{Diversify Y (latent)} rows). 
    }
    \label{tbl:token-query-cost}
\end{table}

\subsection{API Querying Cost}
\label{apd:api-token-cost-analysis}
In this section, we show the token querying cost\footnote{We used GPT3.5 input \$0.001, output \$0.002 and GPT4 input \$0.03, output \$0.06 for 1K tokens (cost at the time of experiments).}
for each step of data generation for the main experiments (\cref{sec:experiment-results}) in Table \ref{tbl:token-query-cost}. 
We make the following observations: 
(1) Simple Prompt costs 4X-5X less than ProgGen, since it contains only NER sample generation with significantly less number of prompt tokens (Appendix \ref{apd:data-gen-sample-gen-setup}). 
(2) The cost for Diversify X are primarily attributed to the high price of querying GPT4. 
(3) For a given dataset, costs for NER sample generation and entity correction are roughly the same across all diversity variants, 
since the only differences are the length of the diversity requirements and number of generated entity annotations, respectively.

\subsection{LLM Self-Correction}

\subsubsection{Prompt Quality}

We found that LLMs are especially biased in certain cases. 
For example, LLMs frequently consider demonyms (e.g. Chinese, American) as location entities. 
We found that more precise categorization of wrongly-annotated terms in the annotation instruction leads to more valid LLM self-corrections 
(e.g. ``Demonyms'' vs ``Adjectives'', ``Ambiguous identifiers'' vs ``General references''). 
Instructing LLMs that certain terms are \emph{not} named entities seems less effective than suggesting such terms are named entities of ``other'' type. 

\subsubsection{Performance Drops}
\label{apd:self-correction-score-drop-reasons}

In the main results (Tables \ref{tbl:main-results} and \ref{tbl:main-results-pr}), we observe that less frequently, 
self-corrections leads to performance drops or insignificant performance differences. 
We hypothesize the reason is unstable training, as a result of 3 scenarios: 
(1) Since average log-probability is imperfect in identifying wrong annotations, 
\textit{valid LLM corrections may not significantly influence downstream model learning}
if the majority of same-class annotation errors are in the remaining 80\% un-selected annotations, and thus not corrected 
(for example, cursory manual inspection found only 6 out of 40 occurrences of \texttt{directed (Director)} selected);
(2) \textit{LLMs may (less frequently) falsely correct previously valid entity annotations}, such as introducing noise into the training data by misclassifying entities;
(3) Certain training samples have a larger influence on model learning~\citep{self-influence}. Changes to the annotations of these samples can lead to significant shifts in results.
Appendix \ref{apd:uncertainty-ranking-experiment} can serve as an evidence for points (2) and (3). 

We notice that occasionally, LLM Self-Correction trades improved precision with lower recalls, indicating LLMs may drop valid named entity annotations.


\subsubsection{Correlation with Correction Statistics}
\label{apd:analysis-correction-result-with-stats}
We make a \emph{loose} assumption that the probability of LLM corrections that are invalid is about the same across diversity variants and regardless of the class of annotation error. 

\paragraph{Downstream Performance}
As such, we can correlate the distribution of LLM correction types (Appendix \ref{apd:correction-stats}) with downstream performance gains (Tables \ref{tbl:main-results} and \ref{tbl:main-results-pr}). 

We note that by definition of F1 score for NER evaluation, (1) valid \crtSpan{entity span corrections} would improve true positives and reduce false positives, 
(2) valid \crtType{entity type corrections} would improve true positives, and 
(3) valid \crtNA{entity drops} would reduce false positives. 
More true positives lead to improvements in precision and recall, whereas less false positives lead to higher recall. 

We make the following observations: 
(1) For all datasets except MIT-Restaurant, we observe high ratio of \crtNA{entity drops} relative to \crtSpan{entity span} and \crtType{type corrections}, indicating high occurrence of false positives (Tables \ref{tbl:correction-dist-by-setup} and \ref{tbl:correction-dist-by-entity-type}), 
which can explain mostly higher precision gains from LLM self-correction. 
(2) For all diversity variants with explicit semantic requirements (\emph{Diversify X} and \emph{Diversify X+Y}), 
we observe noticeably higher ratios of annotations LLM corrected (Table \ref{tbl:correction-dist-by-setup}), 
reflecting the low-quality annotations related to sentence diversity. 

\paragraph{Annotation Difficulty}
With the same assumption, we can also get insight into which classes are more challenging for LLMs to annotate. 
In Table \ref{tbl:correction-dist-by-entity-type}, 
we observe that the difficulty for annotating entities in specific domains is greater and more varied across classes 
(e.g. In MIT-Movie, LLMs corrected from 10+\% to \textasciitilde 60\% selected entities). 

Specifically, based on the distribution of entity type corrections (Figure \ref{fig:type-correction-counts}),
we can probe into how well LLMs separate relevant entity classes. 
Notably, (1) in CoNLL-2003 and WikiGold, Location entities and Organization entities are more interleaved. 
(2) In MIT-Movie, Genre and Plot entities, and Viewers' Rating and Review entities, are more mixed up. 
(3) LLMs especially struggle with identifying the correct type in MIT-Restaurant, 
with Amenity, Cuisine and Dish entities mixed together and many Amenity and Hours entities falsely labeled.

\begin{table*}[tb]
    \centering
    \resizebox{1\textwidth}{!}{
    \footnotesize
    \begin{tabular}{ l m{0.85\textwidth} c m{0.22\textwidth} }
        \toprule 
        \thead{Type} & \thead{Span} & \thead{Issue} & \thead{Desc.} \\
        
        \midrule
        \multicolumn{4}{l}{\hspace{-0.75em} \emph{CoNLL-2003}} \\
        \midrule
        \multirow{3}{*}[-0.8em]{person}
            & Renowned economist \textbf{Dr. \ul{John Smith}} predicts a downturn in the stock market. & \lbSpan & Starting title \\ \crtCline
            & Local \textbf{woman} raises funds for homeless shelter by knitting blankets. & \lbNA & Generic reference \\ \crtCline 
            & \textbf{CEO of \ul{General Motors}} predicts electric vehicles will account for 40\% of company's sales by 2025. 
                & \makecell{\lbNA/\lbType/\\\lbSpan\&\lbType} & {Non-name reference / Title / \par Contains named org. entity} \\
        
        \addlinespace[0.125em] \hline \addlinespace[0.25em]
        \multirow{5}{*}[-1.1em]{location}
            & \textbf{Japanese} prime minister announces economic reforms to boost the country's GDP. & \lbType & Denonym  \\ \crtCline
            & \textbf{Hurricane Katrina} devastates New Orleans, Louisiana. & \lbType & Natural disaster \\ \crtCline
            & Celebrities flock to \textbf{\ul{Paris} Fashion Week} for the latest trends. 
                & \makecell{\lbType/\\\lbSpan} & {Event / \par Contains named loc. entity} \\ \crtCline
            & New study shows increase in pollution levels in \textbf{major cities}. & \lbNA & Categorical reference \\ \crtCline
            & New mural depicting local history to be commissioned in \textbf{downtown arts district}. & \lbNA & Non-name reference \\
        
        \addlinespace[0.125em] \hline \addlinespace[0.25em]
        \multirow{5}{*}[-0.5em]{organization}
            & \textbf{Chinese} President Xi Jinping meets with Indian Prime Minister Narendra Modi to discuss trade relations. & \lbType & Denonym  \\ \crtCline
            & WHO issues new guidelines for \textbf{COVID-19} prevention. & \lbType & Virus \\ \crtCline
            & Apple announces new \textbf{iPhone} release date. & \lbType & Product \\ \crtCline
            & New startup \textbf{company} in Edinburgh secures funding for expansion. & \lbNA & Generic reference \\ \crtCline
            & Women starts \textbf{non-profit organization} to provide free meals for the homeless in downtown Los Angeles. & \lbNA & Categorical reference \\
        
        \midrule
        \multicolumn{4}{l}{\hspace{-0.75em} \emph{WikiGold}} \\
        \midrule
        \multirow{5}{*}[-1.5em]{person}
            & \textbf{Dr. James E. Webb} was instrumental in the development and success of the Apollo lunar landing program. & \lbSpan & Starting title \\ \crtCline
            & \textbf{President of the \ul{United States}} is the head of state and head of government of the United States of America. 
                & \makecell{\lbNA/\lbType/\\\lbSpan\&\lbType} & {Non-name reference / Title / \par Contains named org. entity} \\ \crtCline 
            & In Greek mythology, \textbf{Apollo} was the god of music, poetry, art, oracles, archery, plague, medicine, sun, light, and knowledge. & \lbType & Fictional figure \\ \crtCline 
            & The \textbf{Mona Lisa} is a famous painting created by Leonardo da Vinci. & \lbType & Work of art \\ \crtCline 
            & She is a well-known \textbf{entrepreneur} who has founded several successful companies in the tech industry. & \lbType & Non-name reference \\
        
        \addlinespace[0.125em] \hline \addlinespace[0.25em]
        \multirow{3}{*}[-1.1em]{location}
            & The \textbf{African} elephant is the largest living terrestrial animal. & \lbType & Denonym  \\ \crtCline
            & According to the National Weather Service, \textbf{Hurricane Katrina} was the costliest natural disaster in the history of the United States. & \lbType & Natural disaster \\ \crtCline
            & The new business park, located in the heart of \textbf{the city}, aims to attract both local and international companies looking to expand their presence in the region. & \lbNA & Non-name reference \\

        \addlinespace[0.125em] \hline \addlinespace[0.25em]
        \multirow{9}{*}[-2.4em]{organization}
            & Sylvia Earle, an \textbf{American} marine biologist, has dedicated her life to the exploration and conservation of the world's oceans. & \lbType & Denonym  \\ \crtCline
            & The director, Steven Spielberg, won an \textbf{Academy Award} for Best Director for his work on the film. & \lbType & Award \\ \crtCline
            & The film premiered at the \textbf{\ul{Cannes} Film Festival}.
                & \makecell{\lbType/\\\lbSpan\&\lbType} & {Event / \par Contains named loc. entity} \\ \crtCline 
            & The 2020 Summer Olympics was postponed to 2021 due to the \textbf{COVID-19} pandemic. & \lbType & Virus \\ \crtCline
            & NASA's \textbf{Artemis program} aims to return humans to the Moon by 2024, with the goal of establishing a sustainable human presence on the lunar surface. 
                & \lbType & Landmark project \\ \crtCline
            & The \textbf{Harry Potter} series, produced by Warner Bros., has been a massive success worldwide. & \lbNA & Work of art \\ \crtCline
            & The Battle of Helm's Deep, a fictional battle in the novel \textbf{The Two Towers} by J.R.R. Tolkien, is one of the most iconic scenes in the book and its film adaptation. 
                & \lbType & Literature \\ \crtCline
            & The Battle of Gettysburg was a significant turning point in the \textbf{American Civil War}. & \lbType & Historical period \\ \crtCline
            & The \textbf{band} released a music video for their latest single on YouTube, which quickly gained millions of views. & \lbNA & Non-name reference \\

        \bottomrule
    \end{tabular}
    }
    \caption{
        Representative LLM annotation errors for CoNLL-2003 and WikiGold. 
        In the \emph{Span} and \emph{Type} columns, we show the entity spans annotated by LLMs in \textbf{bold}, within the sentences as context, and the corresponding annotated types. 
        In \emph{Issue}, we label the wrong annotation as one of (1) \crtSpan{wrong span (Span)}, (2) \crtType{wrong type (Type)}, and (3) \crtNA{not a named entity (NA)}. 
        Spans that can be considered as named entities outside the relevant classes in the respective dataset are also labeled as \crtType{Type}. 
        Multiple valid entity interpretations are separated by slash. 
        Candidate entity spans are \protect\ul{underlined}. 
        We summarize the class of annotation error in the \emph{Desc.} column. 
    }
    \label{tbl:annotation-errors-general-domain}
\end{table*}

\begin{table*}[tb]
    \centering
    \resizebox{0.92\textwidth}{!}{
    \footnotesize
    \begin{tabular}{ l m{0.65\textwidth} c m{0.22\textwidth} }
        \toprule 
        \thead{Type} & \thead{Span} & \thead{Issue} & \thead{Desc.} \\
        
        \midrule
        \multicolumn{4}{l}{\hspace{-0.75em} \emph{MIT-Movie}} \\
        \midrule
        \multirow{2}{*}[-0em]{Title}
            & Can I find tickets for the \textbf{new \ul{James Bond} movie}? & \makecell{\lbSpan/\\\lbSpan\&\lbType} & Starting \& trailing desc. \par Character \\ \crtCline
            & What is the plot of the \textbf{newest movie} starring Owen Wilson? & \lbNA & Generic reference \\

        \addlinespace[0.125em] \hline \addlinespace[0.25em]
        \multirow{2}{*}[-0.1em]{Viewers' Rating}
            & Please show me a movie with a \ul{\textbf{high} viewers' rating}. & \lbSpan & Incomplete phrase \\ \crtCline
            & What's the \textbf{rating} for the latest Marvel movie? & \lbNA & Generic reference \\

        \addlinespace[0.125em] \hline \addlinespace[0.25em]
        \multirow{2}{*}[-0.1em]{Year}
            & When was "The Godfather" \textbf{released}? & \lbNA & Time indicator \\ \crtCline
            & Can you show me the trailer for the \textbf{latest} Harry Potter film? & \lbNA & Temporal descriptor \\

        \addlinespace[0.125em] \hline \addlinespace[0.25em]
        \multirow{2}{*}[-0.1em]{Genre}
            & Are there any \textbf{good \ul{action} movies} playing today? & \lbSpan & Starting \& trailing desc. \\ \crtCline
            & What movie features a \textbf{Dance competition} and compelling plot twists & \lbType & Movie element (Plot) \\

        \addlinespace[0.125em] \hline \addlinespace[0.25em]
        \multirow{2}{*}[-1.3em]{Director}
            & Who \textbf{directed} the highest-rated horror movie of the year? & \lbNA & Director indicator \\ \crtCline
            & I'm open to watching a movie directed by a \textbf{female filmmaker} that received critical acclaim. Do you have any recommendations? & \lbNA & Non-name reference \\ \crtCline
            & Can I see a list of theaters showing the \textbf{new \ul{Christopher Nolan} film}? & \lbSpan & Starting \& trailing desc. \\

        \addlinespace[0.125em] \hline \addlinespace[0.25em]
        \multirow{2}{*}[-0.1em]{MPAA Rating}
            & I want to see a trailer for The Wolf of Wall Street, \textbf{rated \ul{R}} & \lbSpan & Starting desc. \\ \crtCline
            & Is John McClane in any \textbf{parental guidance} films that are worth watching? & \lbType & Rating interpretation \\

        \addlinespace[0.125em] \hline \addlinespace[0.25em]
        \multirow{2}{*}[-0.1em]{Plot}
            & Which director has released a new film with a \textbf{thrilling plot} that I can get a sneak peek of? & \lbNA & Plot desc. \\ \crtCline
            & could you tell me the \textbf{plot} of the latest marvel film? & \lbNA & Generic reference \\

        \addlinespace[0.125em] \hline \addlinespace[0.25em]
        Actor & Who are the \textbf{main actors} in the latest sci-fi movie directed by Christopher Nolan? & \lbNA & Generic reference \\

        \addlinespace[0.125em] \hline \addlinespace[0.25em]
        Trailer & Can you play a \textbf{Movie \ul{clip}} from Inception starring Marion Cotillard? & \lbSpan & Starting desc. \\

        \addlinespace[0.125em] \hline \addlinespace[0.25em]
        \multirow{2}{*}[-1.2em]{Song}
            & Can you recommend a movie with \textbf{spectacular music} from 1971? & \lbNA & Song attribute \\ \crtCline
            & Show me a movie from 1999 with a \textbf{soundtrack} that includes dance music. & \lbNA & Generic reference \\ \crtCline
            & I want to watch a movie with a song by \textbf{Adele}. Show me the options. & \lbType & Artist \\

        \addlinespace[0.125em] \hline \addlinespace[0.25em]
        \multirow{2}{*}[-0.1em]{Review}
            & What movie with a mind-bending plot has the \textbf{best viewers' rating}? & \lbType & Viewers' Rating \\ \crtCline
            & I'm looking for a \textbf{superb} movie directed by Morpheus. Can you help me find one? & \lbType & Viewers' Rating \\

        \addlinespace[0.125em] \hline \addlinespace[0.25em]
        Character & Show me a movie with a \textbf{strong female lead} character. & \lbNA & Categorical reference \\
        
        \midrule
        \multicolumn{4}{l}{\hspace{-0.75em} \emph{MIT-Restaurant}} \\
        \midrule
        Restaurant Name & What are the hours of operation for the \textbf{steakhouse} downtown? & \lbType & Cuisine \\

        \midrule
        \multirow{2}{*}[-0.1em]{Amenity}
            & Recommend a family-friendly diner that serves \textbf{breakfast} all day. & \lbType & Hours \\ \crtCline
            & Where is the closest cafe for a good \textbf{Cafe} Mocha? & \lbType & Cuisine \\

        \midrule
        \multirow{2}{*}[-0.1em]{Cuisine}
            & What time does the \textbf{\ul{Italian} restaurant} on Main Street close tonight? & \lbSpan & Trailing desc. \\ \crtCline
            & Is there a 24-hour restaurant nearby that offers a \textbf{breakfast} buffet? & \lbType & Hours \\ 

        \midrule
        \multirow{2}{*}[-0.1em]{Dish}
            & I'm craving some fast food, where can I find a good \textbf{burger joint}? & \lbType & Cuisine \\ \crtCline
            & I want to try a new seafood \textbf{dish}, any suggestions? & \lbNA & Generic reference \\

        \midrule
        Hours & What are the \textbf{operating hours} for the BBQ food truck on 5th Avenue? & \lbNA & Hours indicator \\

        \midrule
        Location & I'm looking for a well-recommended seafood restaurant \ul{in this \textbf{area}}. & \lbSpan & Whole descriptive phrase \\

        \midrule
        \multirow{2}{*}[-0.1em]{Price}
            & Which restaurant in this area serves Italian cuisine at a \textbf{\ul{reasonable} price}? & \lbSpan & Trailing desc. \\ \crtCline
            & What is the \textbf{price range} for italian restaurants in this area? & \lbNA & Price indicator \\

        \midrule
        Rating & I'm looking for a high-end restaurant with a \ul{\textbf{5-star} rating} and a location in downtown Manhattan & \lbSpan & Whole descriptive phrase \\

        \bottomrule
    \end{tabular}
    }
    \caption{
        Representative LLM annotation errors for MIT-Movie and MIT-Restaurant. 
    }
    \label{tbl:annotation-errors-specific-domain}
\end{table*}

\subsection{Representative Annotation Errors}
\label{apd:representative-annotatino-errors}

\normalsize

In this section, we select and show representative entity annotation errors for each dataset in Tables \ref{tbl:annotation-errors-general-domain} and \ref{tbl:annotation-errors-specific-domain} 
to highlight systematic biases in LLMs. 
See the exact and illustrative annotation instructions that covers broader classes of LLM annotation errors 
repository\footnote{\url{https://github.com/StefanHeng/ProgGen/reproduce/correction}}.

We note that, for general domain datasets (CoNLL-2003 and WikiGold), 
the main problem in LLM annotations is false positives, including 
(1) annotating generic and non-name references (e.g. CEO as person entity) and 
(2) categorizing irrelevant named entities (e.g. mythological figures as person entities). 
For specific domain datasets (MIT-Movie and MIT-Restaurant), LLMs mainly suffer from 
(1) imprecise spans annotated (e.g. ``action movie'' as opposed to ``action''),
and (2) entity type confusion (e.g. Viewers' Rating vs Review). 
In MIT-Restaurant, we also observed that GPT-3.5 may drop many valid named entities (more frequently than other datasets). 

\begin{table*}[tb]
    \centering
    \resizebox{0.8\textwidth}{!}{
    \footnotesize
    \begin{tabular}{ c m{0.65\textwidth} l }
        \toprule 
        \thead{Correct?} & \thead{Span} & \thead{Type} \\
        
        \midrule
        \multicolumn{3}{l}{\hspace{-0.75em} \emph{MIT-Movie}} \\
        \midrule
        \multirow{3}{*}[-0em]{\cmark}
            & what is the \textbf{best} viewer rated vampire film & \multirow{3}{*}[-0em]{Viewers' Rating} \\
            & what are some \textbf{good} kids movies starring adrian pasdar & \\
            & list the romance films directed by james cameron rated \textbf{must see} \\
        
        \pptHline 
        \multirow{3}{*}{\xmark}
            & what was the \textbf{{best}} rated stanley kubrick film & \multirow{3}{*}[-0em]{\error{Review}} \\
            & what are some \textbf{good} animated films \\
            & are there any meg ryan romantic comedy movies that are considered \textbf{must see} \\
        
        \midrule
        \multicolumn{3}{l}{\hspace{-0.75em} \emph{MIT-Restaurant}} \\
        \midrule
        \multirow{3}{*}[-0em]{\cmark}
            & is there a \textbf{breakfast} place that has valet parking & \multirow{3}{*}[-0em]{Hours} \\
            & find me a southwestern restaurant that serves \textbf{breakfast} and is located nearby & \\
            & can you help me get to a restaurant where i can get \textbf{lunch} for under 10 & \\
        
        \pptHline 
        \multirow{4}{*}[-0.1em]{\xmark}
            & id like to find a \textbf{breakfast} place & \multirow{3}{*}[-0em]{\error{Cuisine}} \\
            & i need to know of a place that serves \textbf{breakfast} beginning as early as 5 30 \\
            & is there anywhere near here that is open 24 hours and serves \textbf{breakfast} \\
        
        \arrayrulecolor{gLightGrey}\addlinespace[0.125em] \cline{2-3} \addlinespace[0.125em]\arrayrulecolor{black}
         & can you help me find a high end restaurant where i can have \textbf{lunch} & \error{Amenity} \\

        \bottomrule
    \end{tabular}
    }
    \caption{
        Selected original NER samples that showcase ambiguous class separation in MIT-Movie and MIT-Restaurant. 
        For each group (same expected entity type), we show correct (\cmark) annotations on top and \error{incorrect} annotations (\xmark) on bottom (w.r.t. our inferred entity definitions). 
        All samples are original samples in the test sets with entity spans in query contexts highlighted in \textbf{bold} (\emph{Span}) and the original type annotations (\emph{Type}). 
    }
    \label{tbl:original-dataset-ambiguous-sep}
\end{table*}

\begin{table*}[tb]
    \centering
    \resizebox{0.65\textwidth}{!}{
    \footnotesize
    \begin{tabular}{ m{0.5\textwidth} c c }
        \toprule 
        
        

        \thead{Span} & \multicolumn{2}{c}{\thead{Type}} \\
        
        \midrule
        \multicolumn{2}{l}{\hspace{-0.75em} \emph{MIT-Movie}} \\
        \midrule
        is there a \textbf{\guideA{trailer}} out for advengers yet & \guideA{Trailer} & \guideB{\emph{NA}} \\
        is billy wilder the \guideA{\ul{director}} of inception & \guideA{Director} & \guideB{\emph{NA}} \\
        show me the latest \textbf{\guideA{trailer}} for the avengers & \guideA{Trailer} & \guideB{\emph{NA}} \\
        what is the \guideA{\ul{mpaa rating}} for star wars episode 5 & \guideA{MPAA Rating} & \guideB{\emph{NA}} \\
        what is the \guideA{\ul{vuewers rating}} for heidi & \guideA{Viewers' Rating} & \guideB{\emph{NA}} \\
        
        \midrule
        \multicolumn{2}{l}{\hspace{-0.75em} \emph{MIT-Restaurant}} \\
        \midrule
        is there a restaurant that has a \textbf{\guideA{\ulc{gPurple}{5 star} rating}} for it that people like & \guideA{Rating} & \guideB{Rating} \\
        find a clean place to eat that has \ulc{gBlue}{\textbf{\guideB{reasonable}} prices} & \guideA{Price} & \guideB{Price} \\
        find me a thai restaurant with a \textbf{\guideA{\ulc{gPurple}{great} rating}} & \guideA{Rating} & \guideB{Rating} \\
        any good place to get a pie at an \ulc{gBlue}{\textbf{\guideB{affordable}} price} & \guideA{Price} & \guideB{Price} \\

        \bottomrule
    \end{tabular}
    }
    \caption{
        Selected original samples that showcase conflicting annotation guidelines in MIT-Movie and MIT-Restaurant. 
        In each group (similar context), we show original samples in the test sets with different high-level guidelines (\guideA{A} and \guideB{B}). 
        We highlight originally annotated entity spans in \textbf{bold} and \ul{underline} alternative entity spans under the opposing guideline (\emph{Span}). 
        In \emph{Type}, we show the corresponding entity type under each guideline. 
    }
    \label{tbl:original-dataset-conflict-guide}
\end{table*}

\subsection{Entity Class Ambiguities}
\label{apd:entity-class-ambiguous}
We note that, espcially for datasets 
in the specific domain (MIT-Movie and MIT-Restaurant), 
certain entity classes are hard to define just given its label (e.g. Viewers' Rating vs Review). 
Without access to entity annotation schemas, there can be multiple valid interpretations leading to different ``correct'' annotations. 
Further, inferring the annotation schemas is challenging even for humans since even original samples in the test set are noisy. 
We show three significant (highly-occurring) classes of ambiguities encountered in our study below. 
See Appendix \ref{apd:representative-annotatino-errors} for representative LLM annotation errors that reflect these challenges. 
In particular, we define the entities with our judgement from reviewing samples in the dataset (mostly determined via majority vote on similar entity contexts) 
and show representative ambiguous samples in the test set. 

\subsubsection{Ambiguous Class Separation}


\paragraph{MIT Movie: Viewers' Rating vs Review}

We separate Viewers' Rating entities and Review entities based on level of detail, 
where Viewers' Rating entities are shorter such as ``good''
and Review entities are more detailed such as ``incredible visual effects''. 
By this definition, a large fraction of Review entities in the test set should have been Viewers' Rating entities. 



\paragraph{MIT Restaurant: Cuisine vs Hours} 

It's unclear whether meal times such as ``breakfast'' and ``lunch'' are Cuisine entities or Hours entities, 
since (1) both types share similar query contexts for the same entity span and (2) the occurrence counts of the same entity span can be close to each other (e.g. 8 vs 9). 
Thus, we found no better solution than just declaring all such terms as Hours entities (for slightly more annotations). 

\vspace{0.5em} \noindent We show samples in Table \ref{tbl:original-dataset-ambiguous-sep}.

\subsubsection{Conflicting Annotation Guideline} 
We note that 
higher-order annotation guidelines are counter-intuitively conflicting in certain cases, where 
annotations need to follow opposite rules for different entity classes\footnote{
    One reason can be that the tasks for MIT-Movie and MIT-Restaurant are more precisely slot filling as opposed to named entity recognition, 
where the relevant information to extract may differ.}.

\paragraph{MIT Movie: Trailer vs \emph{rest}} 
We note that annotating the Trailer class is not precisely NER, where vast majority of the annotated spans 
simply indicate querying about trailer information (\guideA{A}). 
However, other entity types don't following the same rule, i.e. querying about other entity types (e.g. director information) are not highlighted (\guideB{B}). 

\paragraph{MIT Restaurant: Rating vs Price} 
Rating entities frequently include "rating" in the phrase, such as "4-star rating" as opposed to "4-star" (\guideA{A}), where as Price entities often omit trailing "price" terms (\guideB{B}). 


\vspace{0.5em} \noindent We show samples in Table \ref{tbl:original-dataset-conflict-guide}. 

\subsubsection{Unseen Edge Cases} 
We note that LLMs-generated samples are more diverse (Appendix \ref{apd:gen-data-diversity-stats}) and 
LLMs may generate edge-case spans unlike any cases in the original entities annotated. 
For example, in WikiGold, LLMs may generate places in mythology as location entities where as the original dataset samples don't contain sentences about mythology. 
In such ambiguous cases, we use our judgement to determine the correct annotation. 
For example, 
(1) in WikiGold, we consider places in mythology such as "underworld" as Location entities; 
(2) In MIT-Movie, we consider popular movie studios such as "Marvel" and "Disney" as Genre entities, 
and we ignore label interpretations (e.g. "Parental Guidance" as opposed to "PG") as MPAA Rating entities. 







\section{Additional Experiments}
\label{apd:experiments}
In this section, we discuss preliminary experiments that guide the ProgGen design choices. 

\begin{table}[tb]
    \centering \scriptsize
    \begin{tabular}{ c m{0.65\columnwidth} c}
        \toprule \thead{Sample\\Format} & \thead{Example} & F1  \\

        \midrule \multicolumn{3}{l}{\hspace{-0.75em} \emph{Natural language}} \\ \addlinespace[0.125em] 
        \midrule Pair & Sentence: ``\tplVal{EU rejects German call to boycott British lamb.}''

Named Entities: [\tplVal{EU} (\tplVal{organization}), \tplVal{German} (\tplVal{miscellaneous}), \tplVal{British} (\tplVal{miscellaneous})]
 & \textbf{59.8} \\

        \addlinespace[0.125em] \hline \addlinespace[0.25em] Inline & Sentence: ``\tplVal{EU rejects German call to boycott British lamb.}''

Annotated Sentence: ``**\tplVal{EU}** (\tplVal{organization}) \tplVal{rejects} **\tplVal{German}** (\tplVal{miscellaneous}) \tplVal{call to boycott} **\tplVal{British}** (\tplVal{miscellaneous}) \tplVal{lamb.}''
 & 58.4 \\

        \midrule \multicolumn{3}{l}{\hspace{-0.75em} \emph{Token-wise}} \\ \addlinespace[0.125em] 
        \addlinespace[0.125em] \hline \addlinespace[0.25em] List & Sentence: ``\tplVal{EU rejects German call to boycott British lamb.}''

Tokens: [\tplVal{EU}, \tplVal{rejects}, \tplVal{German}, \tplVal{call}, \tplVal{to}, \tplVal{boycott}, \tplVal{British}, \tplVal{lamb}, \tplVal{.}]

NER tags: [\tplVal{B-organization}, \tplVal{O}, \tplVal{B-miscellaneous}, \tplVal{O}, \tplVal{O}, \tplVal{O}, \tplVal{B-miscellaneous}, \tplVal{O}, \tplVal{O}]
 & 52.1 \\ 

        \addlinespace[0.125em] \hline \addlinespace[0.25em] Lines & sentence:

\tplVal{EU}, \tplVal{B-organization}

\tplVal{rejects}, \tplVal{O}

\tplVal{German}, \tplVal{B-miscellaneous}

\tplVal{call}, \tplVal{O}

\tplVal{to}, \tplVal{O}

\tplVal{boycott}, \tplVal{O}

\tplVal{British}, \tplVal{B-miscellaneous}

\tplVal{lamb}, \tplVal{O}

\tplVal{.}, \tplVal{O}
 & - \\
        
        \bottomrule
    \end{tabular}
    \caption{NER Sample Format examples and downstream performance on CoNLL-2003. } \vspace{-0.5pc}
    \label{tbl:ner-sample-formats}
\end{table}

\subsection{Sample Format Search}
\label{apd:sample-format-search}
Cursory experiments explored 4 different formats for representing NER samples (examples in Table \ref{tbl:ner-sample-formats}) in two directions: 
(1) generating complete sentences (\emph{natural language}), and 
(2) generating token by token (\emph{token-wise}): 
\begin{enumerate}
    \item \emph{Natural Pair}. Generate a sentence and the list of (entity name, entity type) pairs. 
    \item \emph{Natural Inline}. Generate a sentence and then generate the sentence again with inline entity class annotations. 
    \item \emph{Token List}. Generate a sentence, then generate the tokens in the sentence, and finally the BIO tags corresponding to each token. 
    \item \emph{Token Lines}. Generate (token, BIO tag) line by line, in CoNLL-2003 style. 
\end{enumerate}

We note that each sample formatting are equally expressive. 
Namely, LLMs have the option to annotate multi-occurring text spans with potentially different types. 



We conduct experiments on the CoNLL-2003 dataset and select the top-performing sample format for the rest of our experiments in this study. 

We generate around 1K NER samples\footnote{We used the GPT3.5 \texttt{gpt-3.5-turbo-0613} checkpoint.}\footnote{Experiments were done between Sep. 12th - 17th, 2023.} using a prompt template like Simple Prompt (Table \ref{tbl:prompt}). 
We use the first 5 samples in the CoNLL-2003 training set as demo examples. 
we observed that the \emph{Token Lines} format had repetition issues and generated exact samples in the original CoNLL-2003 dataset. Thus, we did not investigate this sample format further. 
A BERT \cite{bert} model was trained on other sample formats with learning rate $2e^{-5}$. 
Other training hyperparameters remain the same as the main experiments (\cref{sec:experiment-setup} and Appendix \ref{apd:implementation}). 
We used the \texttt{bert-base-cased} checkpoint from HuggingFace. 
We compare the best test set F1 of the BERT model from epoch-wise evaluation (results in Table \ref{tbl:ner-sample-formats}). 
The model trained on datasets generated via the \emph{Natural Pair} format performs the best.

\begin{figure}[tb]
    \centering

    \begin{subfigure}[b]{1\columnwidth}
        \centering
        \hspace{-0.8pc}\includegraphics[width=1.03\columnwidth]{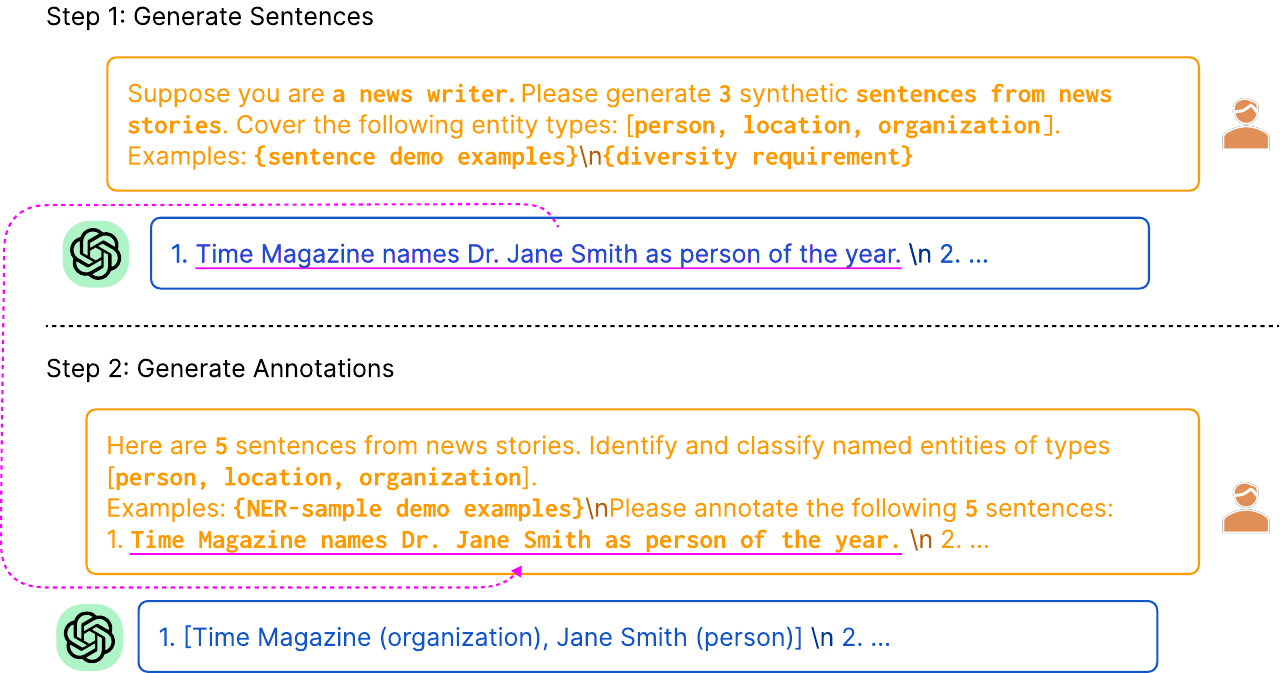}  
        \caption{
            2-Stage NER Sample Generation pipeline. 
            We first prompt LLMs to generate sentences. The LLMs are prompted again to annotate the generated sentences.
        }
        \label{fig:2-stage-pipeline}
    \end{subfigure}
    
    \vspace{0.5pc}
    
    \begin{subfigure}[b]{1\columnwidth}
        \hspace{-0.7pc}\includegraphics[width=1.04\columnwidth]{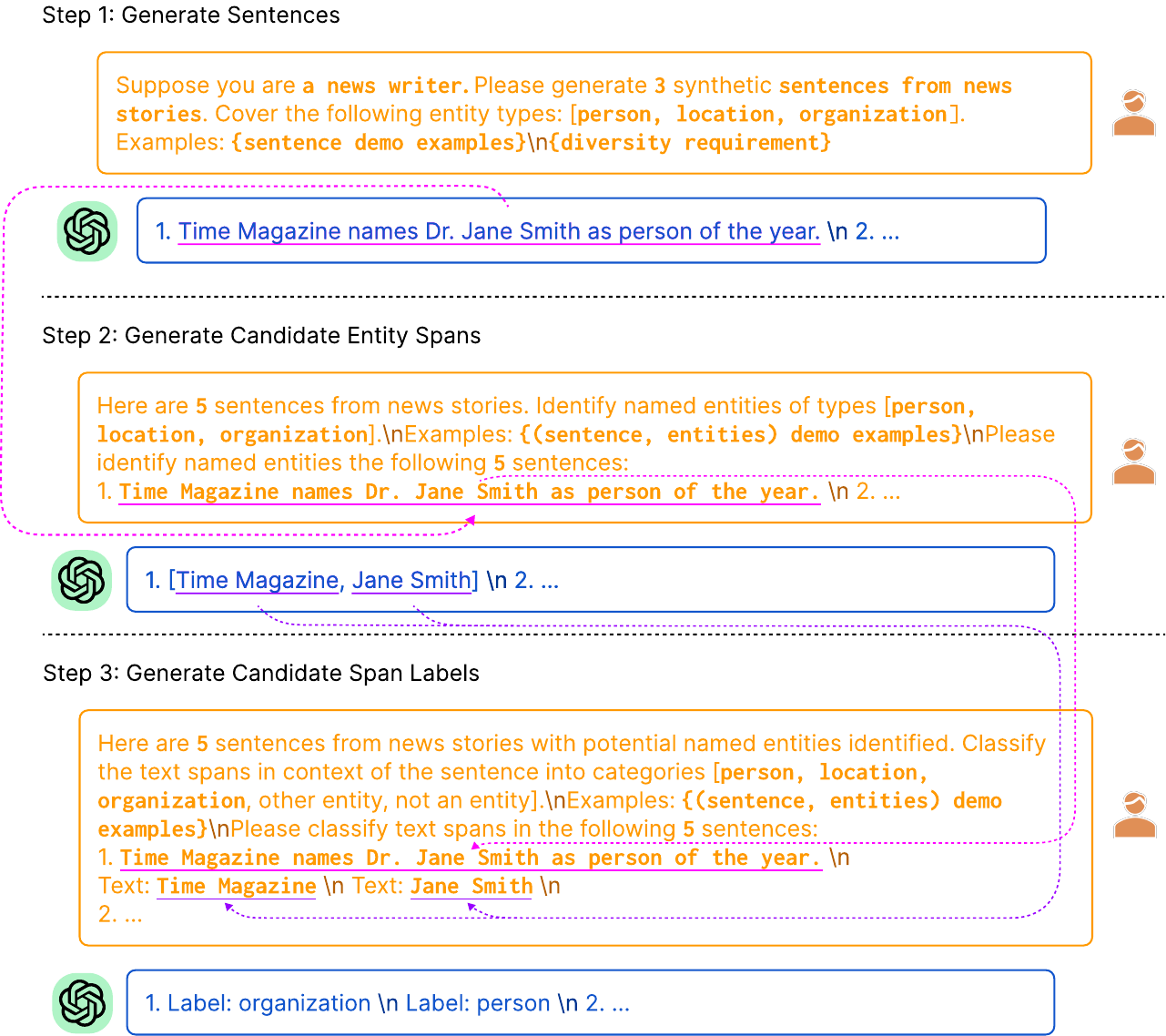}
        \caption{
            3-Stage NER Sample Generation pipeline. 
            We first prompt LLMs to generate sentences. Then, LLMs are prompted again to identify potential named entities in the generated sentences.
            Finally, LLMs are prompted to classify the type of each named entity identified in context of the generated sentences, with the option to drop the named entity. 
        }
        \label{fig:3-stage-pipeline}
    \end{subfigure}
    \vspace{-1.2pc}
    \caption{Step-wise NER Sample Generation pipeline.}
    \label{fig:step-wise-pipeline}
\end{figure}

\begin{table}[tb]
    \centering 
    \resizebox{1\columnwidth}{!}{
        
        
            
            
            

    
    \begin{tabular}{ l *5{c} }
        \toprule
        \thead{Method} & \thead{CoNLL\\-2003} & \thead{Wiki\\Gold} & \thead{MIT-\\Movie} & \thead{MIT-\\Restaurant} & \thead{Average} \\
        
        \midrule
        Simple Prompt & 71.3 & 75.6 & 65.5 & 54.2 & 66.7 \\
        
        \midrule
            Diversify X & 72.7 \ng{+1.4} & 76.8 \ng{+1.2} & 62.7 \nr{-2.8} & 57.2 \ng{+3.0} & 67.4 \ng{+0.7}  \\
            
            Diversify Y (vanilla) & 73.6 \ng{+2.3} & \textbf{77.4 \ng{+1.8}} & \textbf{70.6 \ng{+5.1}} & \textbf{61.3 \ng{+7.1}} & \textbf{70.7 \ng{+4.0}} \\
            
            Diversify Y (latent) 
                & \textbf{74.9 \ng{+3.6}} & 76.7 \ng{+1.1} & 65.9 \ng{+0.4}  & 59.2 \ng{+5.0} & 69.2 \ng{+2.5} \\
            
            Diversify X + Y 
                & \textbf{74.9 \ng{+3.6}} & 76.7 \ng{+1.1} & 65.8 \ng{+0.3} & 57.9 \ng{+3.7} & 68.8 \ng{+2.1} \\

        \bottomrule
    \end{tabular}
    }
    \caption{2-Stage Data Generation Performance Comparison (F1).}
    \label{tbl:2-stage-results}
\end{table}

        

\begin{table}[tb]
    \centering 
    \tiny
    \begin{tabular}{ m{0.94\columnwidth} }
        \toprule

        \emph{Instruction}: Here are \tplVal{5 sentences from news stories}. Please analyze each sentence and identify all named entities occurred that belong to one of the following entity types:

[\tplVal{person, location, organization}].

Show your reasoning in bullet points.

Please list such named entities with the corresponding entity types on the following line, in the order of occurrence.

If no entity is found in the generated sentence, leave the brackets empty. \\
        \pptHline
        \emph{Examples}: Here are some example reasonings and annotations for your reference. Please follow this format.\nl

Examples:\nl

1. Sentence: "\tplVal{Saudi Arabia executes Pakistani man.}"\nl\nl
Reasoning:

{\color{gDarkBlue}
- "Saudi Arabia" is a general country name and hence a location.

- "Pakistani man" refers to an individual, but it's not the name of a specific person, so it's not a person entity.\nl
}

Named Entities: [\tplVal{Saudi Arabia (location)}]\nl

2. Sentence: "\tplVal{BASEBALL - DODGERS WIN FIFTH STRAIGHT.}"\nl\nl
Reasoning:

{\color{gDarkBlue}
- "WIN FIFTH STRAIGHT" likely describes the record of a baseball team, so "DODGERS" should be the name of a sports team. Thus, it's an organization.

- "DODGERS" is a known baseball team name, which qualifies it as an organization entity.\nl
}

Named Entities: [DODGERS (organization)]\nl

3. Sentence: "\tplVal{With 2.5 percent gross domestic product growth expected for 1996, new job growth should slowly lower the unemployment rate over the rest of the year.}"\nl\nl
Reasoning:

{\color{gDarkBlue}
- "2.5 percent" is a percentage. "1996" is a year. The sentence does not mention specific persons, geographical locations, or named organizations.\nl
}

Named Entities: []\nl

4. Sentence: "\tplVal{A.de Silva not out 49}"\nl\nl
Reasoning:

{\color{gDarkBlue}
- "A.de Silva" appears to be an individual's name, hence it's categorized as a person.

- "not out 49" seems to describe a player's score/status in a sports game. Thus, "A.de Silva" is the name of a player, thus a person entity.\nl
}

Named Entities: [\tplVal{A.de Silva (person)}]\nl\nl

\pptSep\nl \\
        \pptHline
        \emph{Input}: Please analyze and annotate the following \tplVal{5 sentences}:\nl

1. Sentence: "\tplVal{Remote village in the Himalayas cut off from aid after heavy snowfall blocks mountain passes.}"\nl

2. Sentence: "\tplVal{The CEO of Amazon, Jeff Bezos, has announced his resignation from the company.}"\nl

3. Sentence: "\tplVal{The United Nations has launched a new initiative to promote gender equality and empower women and girls around the world.}"\nl

4. Sentence: "\tplVal{Russian President Vladimir Putin visits China to discuss economic cooperation.}"\nl

5. Sentence: "\tplVal{Megan Markle and Prince Harry speak out against racism and discrimination in the media.}"\nl

\pptSep\nl

Show your reasoning. Let's think step by step. \\
        \bottomrule
    \end{tabular}
    \vspace{-0.5pc}
    \caption{2-Stage Data Generation step 2: Generate Entity Annotations with CoT prompt example for CoNLL-2003.}
    \vspace{-0.5pc}
    \label{tbl:2-stage-annotate-cot-prompt}
\end{table}

\begin{table}[tb]
    \centering 
    \tiny
    \begin{tabular}{ m{0.94\columnwidth} }
        \toprule        
        \emph{Instruction}: Here are \tplVal{5 sentences from news stories}. For each sentence, please identify all named entities occurred using the format [named entity 1; ...].

Especially identify named entities that belong to one of the following entity types:

[\tplVal{person, location, organization}].

Try to include all possible named entity spans at the expense of potential overlaps.

If no entity is found in the generated sentence, leave the brackets empty. \\
        \pptHline
        \emph{Examples}: Here are some example entity identifications for your reference. Please follow this format.\nl

Examples:\nl

1. Sentence: "\tplVal{With 2.5 percent gross domestic product growth expected for 1996, new job growth should slowly lower the unemployment rate over the rest of the year.}"

Named Entities: []\nl

2. Sentence: "\tplVal{BASEBALL - DODGERS WIN FIFTH STRAIGHT.}"

Named Entities: [\tplVal{DODGERS}]\nl

3. Sentence: "\tplVal{Saudi Arabia executes Pakistani man.}"

Named Entities: [\tplVal{Saudi Arabia; Pakistani; Pakistani man}]\nl

4. Sentence: "A.de Silva not out 49"

Named Entities: [\tplVal{A.de Silva}]\nl\nl

\pptSep \\
        \pptHline
        \emph{Input}: Please identify named entities in the following 5 sentences:

1. Sentence: "\tplVal{Mary Barra, the CEO of General Motors, announced plans to open a new production facility in Tokyo.}"\nl

2. Sentence: "\tplVal{Munich-born CEO of Volkswagen announces plans to expand electric vehicle production in Dubai, United Arab Emirates.}"\nl

3. Sentence: "\tplVal{Renowned scientist, Dr. Jane Smith, awarded prestigious Nobel Prize for groundbreaking research in medical science.}"\nl

4. Sentence: "\tplVal{The United Nations reports a surge in humanitarian aid to Afghanistan as winter approaches.}"\nl

5. Sentence: "\tplVal{Renowned climber John Smith reaches the summit of Mount Everest for the fifth time.}"\nl \\

        \bottomrule
    \end{tabular}
    \vspace{-0.5pc}
    \caption{3-Stage Data Generation step 2: Generate Candidate Entity Spans prompt example for CoNLL-2003.}
    \vspace{-0.5pc}
    \label{tbl:3-stage-gen-step-2-conll2003}
\end{table}


\begin{table}[tb]
    \centering 
    \tiny
    \begin{tabular}{ m{0.94\columnwidth} }
        \toprule
        
        \emph{Instruction}: Here are some \tplVal{sentences from news stories} with potential named entities identified. Your task is to classify whether the spans of text are named entities of the following entity types:

[\tplVal{person, location, organization}].

In particular, for each sentence with a span of text enclosed in double curly braces, please analyze and classify the enclosed span of text in context of the sentence into one of the following categories:

[\tplVal{named person entity, named location entity, named organization entity}, named entity of other type, not a named entity] \\
        \pptHline
        \emph{Examples}: 
Here are some example span analyses and classifications for your reference. Please follow this format.

Examples:

1. Sentence: "\tplVal{A.de Silva not out 49}"

Text Span: "\tplVal{A.de Silva}"\nl
Label: \tplVal{"de Silva" is a first name initial followed by a common last name, so "A.de Silva" should be an individual's name. Additionally, "not out 49" describes a player's score in a sports game. Based on context, "A.de Silva" is the name of a player. A player is a person. A specific person's name is a person entity. The span is a named person entity.}\nl

2. Sentence: "\tplVal{With 2.5 percent gross domestic product growth expected for 1996, new job growth should slowly lower the unemployment rate over the rest of the year.}"

Text Span: "\tplVal{2.5 percent}"\nl
Label: \tplVal{"2.5 percent" is a percentage and a named entity. However, a percentage is not a person, location, or organization. The span is a named entity of other type.}

Text Span: "\tplVal{1996}"\nl
Label: \tplVal{"1996" is a year and a named entity. However, a year is not a person, location, or organization. The span is a named entity of other type.}\nl

3. Sentence: "\tplVal{BASEBALL - DODGERS WIN FIFTH STRAIGHT.}"

Text Span: "\tplVal{DODGERS}"\nl
Label: \tplVal{"WIN FIFTH STRAIGHT" describes the record of a baseball team, so "DODGERS" refers to a sports team by its name. Additionally, "DODGERS" is a known baseball team name. A sports team is an organization. The span is a named organization entity.}\nl

4. Sentence: "\tplVal{Saudi Arabia executes Pakistani man.}"

Text Span: "\tplVal{Saudi Arabia}"\nl
Label: \tplVal{"Saudi Arabia" is the name of a specific country. The name of a country is a location entity. The span is a named location entity.}

Text Span: "\tplVal{Pakistani}"\nl
Label: \tplVal{"Pakistani" is a adjective describing where the "man" is from. The span "Pakistani" in itself is a nationality, not a country name. A adjective describing a nationality is not a named entity. The span is not a named entity.}

Text Span: "\tplVal{Pakistani man}"\nl
Label: \tplVal{"Pakistani man" refers to someone but does not provide the individual's name. A reference to a person not by the name is not a named entity. The span is not a named entity.}\nl

\pptSep \\
        \pptHline
        \emph{Input}: 
Please analyze and classify spans of text in the following 5 sentences:

1. Sentence: "\tplVal{Organization for Economic Cooperation and Development predicts global economic growth to slow down in 2021.}"

Text Span: "\tplVal{Organization for Economic Cooperation and Development}"\nl
Text Span: "\tplVal{2021}"\nl

2. Sentence: "\tplVal{Sahara Desert experiences record high temperatures, posing challenges for local communities and wildlife.}"

Text Span: "\tplVal{Sahara Desert}"\nl

3. Sentence: "\tplVal{Johnny Depp to star in new film directed by Tim Burton.}"

Text Span: "\tplVal{Johnny Depp}"\nl
Text Span: "\tplVal{Tim Burton}"\nl

4. Sentence: "\tplVal{Stanford University appoints former UK prime minister as guest lecturer for political science course.}"

Text Span: "\tplVal{Stanford University}"\nl
Text Span: "\tplVal{UK}"\nl
Text Span: "\tplVal{prime minister}"\nl

5. Sentence: "\tplVal{Department of Justice announces new initiative to combat cybercrime in partnership with global organization.}"

Text Span: "\tplVal{Department of Justice}"\nl
Text Span: "\tplVal{global organization}" \\
        \bottomrule
    \end{tabular}
    \vspace{-0.5pc}
    \caption{3-Stage Data Generation step 3: Generate Candidate Span Labels with CoT prompt example for CoNLL-2003.}
    \vspace{-0.5pc}
    \label{tbl:3-stage-gen-step-3-conll2003}
\end{table}

\begin{table}[tb]
    \centering 
    \tiny
    \begin{tabular}{ m{0.94\columnwidth} }
        \toprule        
        \emph{Instruction}: Here are \tplVal{5 spoken queries to a dialogue system about movies}. Please analyze each query and identify and extract informative keywords in the queries using the format [keyword 1; ...].

Especially identify informative keyword spans that belong to one of the following entity types:

[\tplVal{Title, Viewers' Rating, Year, Genre, Director, MPAA Rating, Plot, Actor, Trailer, Song, Review, Character}].

Try to include all possible keyword spans at the expense of potential overlaps.

\tplVal{Genre keywords should not end in "movie" or "film".} \\
        \pptHline
        \emph{Examples}: Here are some example keyword analyses and identifications for your reference. Please follow this format.\nl

Examples:\nl

1. Query: "\tplVal{what type of movie genre is the perfect weapon}"

Likely Keywords: \tplVal{"the perfect weapon" is a movie name. Likely keywords are [the perfect weapon].}\nl

2. Query: "\tplVal{show me a movie with the song a whole new world}"

Likely Keywords: \tplVal{"a whole new world" is a song name. Likely keywords are [a whole new world].}\nl

3. Query: "\tplVal{how many movies came out in 2004}"

Likely Keywords: \tplVal{"2004" refers to a time period. Likely keywords are [2004].}\nl

4. Query: "\tplVal{when did mark joffe direct the bounty hunter film that is rated pg}"

Likely Keywords: \tplVal{"mark joffe" is a person's name. "bounty hunter" describes a film element. "pg" is a movie rating. Likely keywords are [mark joffe; bounty hunter; pg].}\nl

5. Query: "\tplVal{is cary grant in any historical films that are a must see}"

Likely Keywords: \tplVal{"cary grant" is the name of a person. "historical" defines a genre. "must see" describes a film. Likely keywords are [cary grant; historical; must see].}\nl

6. Query: "\tplVal{could you show me some part of the new indiana jones movie}"

Likely Keywords: \tplVal{"some part" indicates a trailer. "indiana jones" refers to a person. Likely keywords are [some part; indiana jones].}\nl

7. Query: "\tplVal{what movie is considered the funniest of all time}"

Likely Keywords: \tplVal{"funniest of all time" describes a movie. Likely keywords are [funniest of all time].}\nl

\nl\pptSep\nl \\
        \pptHline
        \emph{Input}: Please analyze and identify potential spans in the following 5 queries:\nl

1. Query: "\tplVal{Can you recommend a high-rated sci-fi movie released in the 21st century?}"\nl

2. Query: "\tplVal{Could you show me the trailer for Sing 2 and tell me the main plot of the movie?}"\nl

3. Query: "\tplVal{Could you recommend a good robot movie with a strong female character in a damsel in distress situation?}"\nl

4. Query: "\tplVal{Find me a movie directed by Quentin Tarantino with a strong female lead character}"\nl

5. Query: "\tplVal{I'm looking for the highest rated horror film from the 2000s, preferably directed by a female director like The Babadook.}" \\

        \bottomrule
    \end{tabular}
    \vspace{-0.5pc}
    \caption{3-Stage Data Generation step 2: Generate Candidate Entity Spans with CoT prompt example for MIT-Movie.}
    \vspace{-0.5pc}
    \label{tbl:3-stage-gen-step-2-mit-movie}
\end{table}

\begin{table*}[tb]
    \centering 
    \tiny
    \begin{tabular}{ m{0.95\textwidth} }
        \toprule
        
        \emph{Instruction}: Here are \tplVal{5 spoken queries to a dialogue system about movies}. Your task is to analyze each query and classify whether the spans of text are named entities of the following entity types:

[\tplVal{Title, Viewers' Rating, Year, Genre, Director, MPAA Rating, Plot, Actor, Trailer, Song, Review, Character}].

In particular, for each query with a span of text, please analyze and classify the span of text in context of the query into one of the following categories:

[\tplVal{named Title entity, named Viewers' Rating entity, named Year entity, named Genre entity, named Director entity, named MPAA Rating entity, named Plot entity, named Actor entity, named Trailer entity, named Song entity, named Review entity, named Character entity}, named entity of other type, not a named entity]. \\
        \pptHline
        \emph{Examples}: Here are some example span analyses and classifications for your reference. Please follow this format.\nl

Examples:\nl

1. Query: "\tplVal{could you show me some part of the new indiana jones movie}"

Text Span: "\tplVal{some part}"\nl
Label: \tplVal{"some part" indicates a segment of a movie. General request to a segment or clip from a movie counts as a Trailer entity. The span is a named Trailer entity.}

Text Span: "\tplVal{indiana jones}"\nl
Label: \tplVal{"Indiana Jones" is a well-known character. He's the protagonist in the "Indiana Jones" franchise first appearing in the film "Raiders of the Lost Ark". The span is a named Character entity.}\nl

2. Query: "\tplVal{what type of movie genre is the perfect weapon}"

Text Span: "\tplVal{the perfect weapon}"\nl
Label: \tplVal{"the perfect weapon" is a concrete movie name. That is, "The perfect weapon" is a specific title of a movie. The span is a named Title entity.}\nl

3. Query: "\tplVal{when did mark joffe direct the bounty hunter film that is rated pg}"

Text Span: "\tplVal{mark joffe}"\nl
Label: \tplVal{"Mark Joffe" is a person's name and he directed a film. Thus, "Mark Joffe" is a director's name. A specific director's name is a Director entity. The span is a named Director entity.}

Text Span: "\tplVal{direct}"\nl
Label: \tplVal{"direct" is a verb describing the action of the director. It is the action of a director. The span is a named entity of other type.}

Text Span: "\tplVal{bounty hunter}"\nl
Label: \tplVal{The span "the bounty hunter film" indicates "bounty hunter" is not a movie name. Thus, "bounty hunter" is a specific plot element for movies. A specific movie theme or storyline counts as a Plot entity. The span is a named Plot entity.}

Text Span: "\tplVal{pg}"\nl
Label: \tplVal{"pg" stands for "Parental Guidance Suggested." This rating is a part of the Motion Picture Association of America (MPAA) film rating system. The span is a named MPAA Rating entity.}\nl

4. Query: "\tplVal{is cary grant in any historical films that are a must see}"

Text Span: "\tplVal{cary grant}"\nl
Label: \tplVal{"Cary Grant" is a person's name and he is in films. Thus, "Cary Grant" is an actor's name. A specific actor's name is a named Actor entity. The span is a named Actor entity.}

Text Span: "\tplVal{historical"}\nl
Label: \tplVal{"historical" is a specific category or style of film. A specific movie category or style is a Genre entity. The span is a named Genre entity.}

Text Span: "\tplVal{historical film}"\nl
Label: \tplVal{"historical" is suffice to serve as a Genre keyword. The addition of film is redundant and the entire span "historical film" is not a named entity. The span is not a named entity.}

Text Span: "\tplVal{must see}"\nl
Label: \tplVal{"must see" indicates a specific level of popularity or recommendation. "must see" is a specific assessment of a movie by its audience. A specific movie assessment is a Viewers' Rating entity. The span is a named Viewers' Rating entity.}\nl

5. Query: "\tplVal{show me a movie with the song a whole new world}"

Text Span: \tplVal{"a whole new world"}\nl
Label: \tplVal{"A whole new world" is the name of a song. A song name featured in a movie is a Song entity. The span is a named Song entity.}\nl

6. Query: "\tplVal{show me viewers' rating and plot summary for a good funny movie from a female filmmaker.}"

Text Span: "\tplVal{viewers' rating}"\nl
Label: \tplVal{"viewers' rating" is not an actual movie assessment from viewers. "viewers' rating" merely requests rating information. A flexible request for a movie assessment doesn't count as a Viewers' Rating entity. The span is not a named entity.}

Text Span: "\tplVal{plot summary}"\nl
Label: \tplVal{"plot summary" simply requests storyline information. "plot summary" is not an actual movie theme or plot element. A flexible request for a movie plot doesn't count as a Plot entity. The span is not a named entity.}

Text Span: "\tplVal{good}"\nl
Label: \tplVal{"good" is an adjective that describes "movie". It implies a high quality movie perceived by viewers. The span is a named Viewers' Rating entity.}

Text Span: "\tplVal{funny}"\nl
Label: \tplVal{"movie" is right after "funny", so the span "funny" directly specifies a movie characteristic. It implies a movie in the comedy genre. The span is a named Genre entity.}

Text Span: "\tplVal{female}"\nl
Label: \tplVal{"female" is a adjective describing a characteristic of "filmmaker". It is a preference, not a female filmmaker's  name. The span is not a named entity.}

Text Span: "\tplVal{female filmmaker}"\nl
Label: \tplVal{"female filmmaker" is a category of directors. It is not the name of a specific director. The span is not a named entity.}\nl

7. Query: "\tplVal{how many movies came out in 2004}"

Text Span: "\tplVal{2004}"\nl
Label: \tplVal{"2004" is a year. It specifies a particular time duration when the movies were released. A time duration of movie release counts as a Year entity. The span is a named Year entity.}\nl

8. Query: "\tplVal{what movie is considered the funniest of all time}"

Text Span: "\tplVal{funniest of all time}"\nl
Label: \tplVal{"funniest of all time" is a detailed critique or opinion about a movie. A detailed movie comment from viewers or critics is a Review entity. The span is a named Review entity.}\nl

\nl\pptSep\nl \\
        \pptHline
        \emph{Input}: 
Please analyze and classify spans of text in the following 5 queries:\nl

1. Query: "\tplVal{Show me the trailer of the animated movie Toy Story released in 1995}"\nl
Text Span: "\tplVal{Toy Story}"\nl
Text Span: "\tplVal{1995}"

2. Query: "\tplVal{Show me a trailer for the highest-rated comedy movie of 2020.}"\nl
Text Span: "\tplVal{highest-rated}"\nl
Text Span: "\tplVal{comedy}"\nl
Text Span: "\tplVal{2020}"

3. Query: "\tplVal{What is the viewers' rating for Manchester by the Sea?}"\nl
Text Span: "\tplVal{Manchester by the Sea}"\nl

4. Query: "\tplVal{Show me a trailer for the movie Titanic, which features the song My Heart Will Go On, and provide a cinema evaluation of the film.}"\nl
Text Span: "\tplVal{trailer}"\nl
Text Span: "\tplVal{Titanic}"\nl
Text Span: "\tplVal{My Heart Will Go On}"\nl
Text Span: "\tplVal{cinema evaluation}"\nl

5. Query: "\tplVal{Who directed the supernatural thriller movie 'The Others' that was released in 2001?}"\nl
Text Span: "\tplVal{supernatural thriller}"\nl
Text Span: "\tplVal{The Others}"\nl
Text Span: "\tplVal{2001}" \\
        \bottomrule
    \end{tabular}
    \vspace{-0.5pc}
    \caption{3-Stage Data Generation step 3: Generate Candidate Span Labels with CoT prompt example for MIT-Movie.}
    \vspace{-0.5pc}
    \label{tbl:3-stage-gen-step-3-mit-movie}
\end{table*}

\begin{table}[tb]
    \centering \footnotesize
    \resizebox{0.95\columnwidth}{!}{
    \begin{tabular}{ c l c }
        \toprule
        \thead{Dataset} & \thead{Approach} & \thead{F1} \\ \addlinespace[0.25em]
        
        \midrule
        \multirow{2}{*}[-0.25em]{CoNLL-2003} 
            & 2-Stage (step 2 vanilla) & 74.9 \\ 
            \addlinespace[0.125em] \cline{2-3} \addlinespace[0.25em] 
            & 3-Stage (Step 2 vanilla, Step 3 CoT) & \textbf{76.8 \ng{+1.9}} \\ 

        \midrule
        \multirow{2}{*}[-0.25em]{MIT-Movie} 
            & 2-Stage (step 2 vanilla) & \textbf{70.6} \\ 
            \addlinespace[0.125em] \cline{2-3} \addlinespace[0.25em] 
            & 3-Stage (Step 2 cot, Step 3 CoT) & 70.5 \nr{-0.1} \\ 
        \bottomrule
    \end{tabular}
    }
    \caption{
        3-Stage Data Generation with CoT Performance Comparison. 
        2-Stage and 3-Stage generation uses the same pool of sentences (i.e. same step 1 output).
        The 2-Stage reference setups don't use CoT prompting. 
        \emph{vanilla} and \emph{CoT} refers to without and with chain-of-thought prompting respectively. 
    } \vspace{-0.5pc}
    \label{tbl:compare-3-stage-cot}
\end{table}

\subsection{Step-wise Generation}
\label{apd:step-wise-gen}

The instruction-following capability of ChatGPT3.5 may be limited. 
The complexity of simultaneously generating diverse samples and accurately annotating entities in a single prompt may present a considerable challenge.
Consequently, we investigate whether separate prompts for each aspect (sentence generation \& sentence annotation) leads to 
improved instruction following and thus improved downstream performance. 
In this section, we explore multiple variants of step-wise sample generation, all starting from sentence generation.

\subsubsection{2-Stage Generation}
\label{apd:2-stage-experiment}
Our 2-Stage NER sample generation pipeline is illustrated in Figure \ref{fig:2-stage-pipeline}. 
We first prompt LLMs to generate diverse sentences. Then, the generated sentences are annotated by LLM. 
In the first step (sentence generation), 
we add explicit diversity requirements just like in default NER sample generation 
(\cref{sec:diversity}; note the diversity requirements are for the sentences). 
Then, for all diversity setups, we use the same annotation prompt with temperature $t = 0$. 

We run experiments with setup just like in \cref{sec:experiment-setup}. 
The results are shown in Table \ref{tbl:2-stage-results}\footnote{Experiments were done between Nov. 28th - Dec. 28th, 2023.}. 
We observe that, compared to the vanilla (1-stage) data generation (Appendix \ref{apd:unstable-api-rerun} and Table \ref{tbl:1-stage-early-results}),
2-stage doesn't yield higher F1 scores, even at the expense of additional token querying cost (for prompting the generated sentences again), 
indicating simply re-annotating in an isolated prompt doesn't improve annotation accuracy.

\paragraph{Chain-of-Thought}

We study whether chain-of-thought (CoT) prompting \citep{cot} can help LLMs annotate better where we manually write annotation reasoning for each demo sample. 
We compare generating entity annotations (step 2) naively (\emph{vanilla}) or with manually written chain-of-thought reasoning (\emph{CoT}). 
We experiment on the CoNLL-2003 dataset with Simple Prompt\footnote{Experiments were done between Nov. 19th - 21st, 2023.}.
An example entity annotation prompt is shown in Table \ref{tbl:2-stage-annotate-cot-prompt}. 
We found that CoT results in worse downstream performance (69.6\% vs 72.0\% F1). 
Manual examination reveals that many LLM-generated reasoning chains are not factual, as if copying the formatting from the demo samples.

\subsubsection{3-Stage Generation}
We also experimented with 3-stage generation (pipeline illustrated in Figure \ref{fig:3-stage-pipeline}), 
further splitting the entity-generation step into candidate entity span generation (step 2) and 
entity span classification (step 3). 
In the last step, LLMs can also optionally drop the entity candidate entirely 
to address the common false-positive issues (Detailed discussion in Appendix \ref{apd:analysis-correction-result-with-stats}). 
Intuitively, since step 3 cannot add entity spans, LLMs are instructed to generate more candidate entities in step 2 where we also add non-entities as demos. 
In the last step, we group all candidate spans for the same sentence together to reduce number of prompt tokens and thus token querying costs (c.f. classifying each distinct (sentence, entity) pair). 

We experimented with CoT prompting for both step 2 and 3 on CoNLL-2003 and MIT-Movie to improve annotation accuracy\footnote{Experiments were done between Dec. 27th - Jan. 15th, 2023.}. 
We show example prompts in in Tables \ref{tbl:3-stage-gen-step-2-conll2003}, \ref{tbl:3-stage-gen-step-3-conll2003}, \ref{tbl:3-stage-gen-step-2-mit-movie} and \ref{tbl:3-stage-gen-step-3-mit-movie}. 
For MIT-Movie, to provide additional annotation instructions, we further added a manually crafted NER sample to showcase representative annotation errors. 
Performance comparisons are shown in Table \ref{tbl:compare-3-stage-cot}.
We found that 3-stage generation shows gains for CoNLL-2003, but makes no difference in F1 score for MIT-Movie. 
Manual inspection shows that the main problem for MIT-Movie is incorrect span boundaries as a result of step 2, 
where as step 3 can only correct type errors. 
Thus, we proposed LLM self-correction (\cref{sec:self-correction}) with more explicit feedback and correction mechanisms. 

\subsection{Entity Annotation Uncertainty Ranking}
\label{apd:compare-annotation-uncertainty-ranking-function}
As discussed in \cref{sec:self-correction} and \cref{sec:annotation-uncertainty-ranking}, 
we compare two scoring functions to select entity annotations for LLM Self-Correction (Appendices \ref{apd:log-prob-uncertainty-ranking} \& \ref{apd:bert-uncertainty-ranking}). 
    

\subsubsection{LLM Generation Log-Prob}
\label{apd:log-prob-uncertainty-ranking}
We average the log-probabilities for all tokens w.r.t the entity annotation (i.e. log-probs for \texttt{span (entity type)}), denoted \emph{log-prob}. 
As such, a lower average log-prob means more uncertainty. 
    
\subsubsection{Downstream BERT Loss}
\label{apd:bert-uncertainty-ranking}
We first train a BERT-class model on the generated dataset and compare its prediction signals with the LLM annotations (i.e. the dataset the model was trained on), 
where lower prediction discrepancy means more uncertainty (denoted \emph{loss}). 
Intuitively, we can leverage feedback from the trained downstream model 
since it learns from a larger pool of NER samples and learns to resolve noises in the original LLM annotations in the training process. 
We run inference with the downstream model on the generated sentences to get token-wise cross entropy (CE) loss w.r.t 
the original LLM annotations\footnote{In terms of implementation, we simply use CrossEntropyLoss from PyTorch treating the LLM annotations as ``ground truth''. }, 
where higher loss means higher discrepancy between BERT and LLMs' annotations. 
To get the ranking score for each LLM entity annotation, we average the negated loss values for the subset of tokens in the sentence that correspond to the entity span. 
We further include one more token on both sides of the entity span if possible to better capture discrepancy in span boundaries. 
For example, given a generated NER sample with 
tokens [\texttt{Bob}, \texttt{is}, \texttt{born}, \texttt{in}, \texttt{Athens}, \texttt{.}] 
and BIO tags [\texttt{B-person}, \texttt{O}, \texttt{O}, \texttt{O}, \texttt{B-location}, \texttt{O}], 
to get the uncertainty score for ``Athens``, we average the negated loss for the tokens [\texttt{in}, \texttt{Athens}, \texttt{.}]. 

We note that conceptually, \emph{loss} ranking
can be considered as a log-probability and hence comparable with \emph{log-prob} on the same scale. 
As shown in equation \ref{eqn:ce-loss}, the negated CE loss for token position $i$ with entity class label $y$ is effectively the log of a probability in range $[0, 1]$: 
\begin{equation}
\label{eqn:ce-loss}
\begin{split}
    - \mathrm{CE~Loss}(i, y)
        &= \log \frac{\exp (\mathrm{logits}(i))}{\mathrm{sum}(\exp (\mathrm{logits}(i)))} \\
        &= \log \mathrm{softmax}(\mathrm{logits}(i))[y] \\
        &= \log P(y)
\end{split}
\end{equation}







\begin{table}[tb]
    \centering \footnotesize
    \begin{tabular}{ c c c }
        \toprule
        \multicolumn{2}{c}{\thead{Approach}} & \multirow{2}{*}[-0.8em]{\thead{F1}} \\ 
        \addlinespace[0.25em] \cline{1-2} \addlinespace[0.25em]
        \thead{Correction\\Source} & \thead{Ranking\\Function} & \\ 
        
        \midrule
        \multicolumn{2}{c}{\emph{Diversify Y (vanilla)}} & 72.7 \\
        
        \midrule
        \multirow{2}{*}[-0em]{LLM} 
            & log-prob & \textbf{74.1 \ng{+1.4}} \\
            & loss & 70.7 \nr{-2.0} \\

        \midrule
        \multirow{2}{*}[-0em]{Manual} 
            & log-prob & \textbf{74.9 \ng{+2.2}} \\
            & loss & 73.2 \ng{+0.5} \\
        
        \bottomrule
    \end{tabular}
    
    \caption{
        Entity Annotation Uncertainty Ranking Function performance comparison on MIT-Movie. 
        F1 \textcolor{gGreen}{gains}/\textcolor{gRed}{drops} are w.r.t before annotation corrections. 
    } 
    \label{tbl:compare-ranking-func}
\end{table}

\begin{table}[tb]
    \centering 
    \resizebox{0.95\columnwidth}{!}{
    \begin{tabular}{ *{6}{c} }
        \toprule
        \multirow{2}{*}[-0.8em]{\thead{\#Annot.\\examined}} & \multirow{2}{*}[-0.2em]{\thead{\#Invalid\\LLM\\Correct.}} & \multicolumn{4}{c}{\thead{Manual Correction Type Counts}} \\
        \addlinespace[0.25em] \cline{3-6} \addlinespace[0.225em]
        & & \thead{Wrong\\Span} & \thead{Wrong\\Type} & \thead{NA} & \thead{\emph{total}} \\ 
        
        \midrule
        707 & \textbf{136} & \textbf{59} & \textbf{82} & 137 & 278 \\
        707 & 131 & 52 & 72 & \textbf{254} & \textbf{378} \\
        
        \bottomrule
    \end{tabular}
    }
    \caption{
        Entity Annotation Uncertainty Ranking Function statistics on MIT-Movie. \emph{total} refers to the total number of annotations corrected. 
    } \vspace{-0.5pc}
    \label{tbl:compare-ranking-stats}
\end{table}

\subsubsection{Experiment}
\label{apd:uncertainty-ranking-experiment}
Compare the quality of entity annotations selected via \emph{log-prob} and \emph{loss}, 
we conducted preliminary experiment on data generated for MIT-Movie via \emph{Diversify Y (vanilla)}. 
The experiment setup including annotation-selection hyperparameters for \emph{log-prob} remain the same as in the main experiments (\cref{sec:experiment-setup}). 
For \emph{loss}, we do not perform any filtering ($P_c = 0$) and simply select the top-ranking $T_c = 20\%$ for each entity class normalized by entity proportions. 
In addition to LLM self-correction, one of the authors manually annotated the selected entity annotations. 
The results are shown in Table \ref{tbl:compare-ranking-func}. 
We observe that whether entity annotations are corrected via LLMs or manually, \emph{log-prob}-selected annotations noticeably yield better downstream F1. 

Treating manual corrections as ground truths, we further examine the corrections as shown in Table \ref{tbl:compare-ranking-stats}. 
We observe that, comparing entities annotations selected via \emph{log-prob} and \emph{loss}, 
(1) the ratio of invalid annotation corrections that LLMs make is the same, and 
(2) \emph{loss} selects more wrong entity annotations and identifies significantly more non-named-entities. 
Thus, we conclude that in this case, \emph{loss} scoring selects entity annotations that less significantly contribute to performance gains.
We hypothesize that the downstream model might identify more errors that it already resolved during the training process (Appendix \ref{apd:self-correction-score-drop-reasons}). 
We look forward to more comprehensive investigations on selecting influential entity annotations in future work.

\begin{table}[tb]
    \centering
    \footnotesize
    \resizebox{0.95\columnwidth}{!}{
    \begin{tabular}{ m{0.55\textwidth} }
        \toprule
        \hspace{-0.75em} \emph{CoNLL-2003} \\ 
        \midrule
        President Biden announces new infrastructure plan \\
        President Biden signs new infrastructure bill into law. \\
        President Biden announces new infrastructure plan to rebuild roads and bridges. \\
        
        \pptHline
        Former President Obama to release new memoir. \\
        Former President Obama to release memoir this fall. \\

        \pptHline
        Taylor Swift releases new album \\
        Taylor Swift released a new album, causing a frenzy among fans. \\
        
        \midrule
        \hspace{-0.75em} \emph{WikiGold} \\ 
        \midrule
        The Golden Gate Bridge is a suspension bridge spanning the Golden Gate strait, the mile-wide, three-mile-long channel between San Francisco Bay and the Pacific Ocean. \\
        The Golden Gate Bridge is a suspension bridge spanning the Golden Gate, the one-mile-wide strait connecting San Francisco Bay and the Pacific Ocean. \\
        \bottomrule
    \end{tabular}
    }
    \caption{
        Example groups of LLMs-generated sentences that are syntactically and semantically similar. 
    }
    \label{tbl:similar-gen-sentence}
\end{table}

\begin{table*}[tb]
    \centering
    \resizebox{0.95\textwidth}{!}{
    \footnotesize
    \begin{tabular}{ m{0.8\textwidth} c c }
        \toprule 
        \hfil\multirow{2}{*}[-0.2em]{\thead{Span}}\hfil & \multicolumn{2}{c}{\thead{Type}} \\
        \addlinespace[0.2em] \cline{2-3} \addlinespace[0.2em]
        & \error{\xmark} & \correct{\cmark} \\

        \midrule
        \multicolumn{2}{l}{\hspace{-0.75em} \emph{CoNLL-2003}} \\
        \midrule
        The \error{\textbf{Prime Minister of \ulc{gGreen}{India}}} has announced plans to invest in infrastructure to boost economic growth. & \error{Person} & \correct{Location} \\
        \error{\textbf{\ulc{gGreen}{California} governor}} signs bill to address homelessness crisis. & \error{Person} & \correct{Location} \\
        The \error{\textbf{Secretary-General of the \ulc{gGreen}{United Nations}}} is calling for an immediate ceasefire in the region. & \error{Person} & \correct{Organization} \\
        Former \error{\textbf{CEO of \ulc{gGreen}{Google}}} to testify before Congress. & \error{Person} & \correct{Organization} \\

        \midrule
        \multicolumn{2}{l}{\hspace{-0.75em} \emph{WikiGold}} \\
        \midrule
        The Moscow Kremlin serves as the official residence of the \error{\textbf{President of the \ulc{gGreen}{Russian Federation}}}. & \error{Person} & \correct{Organization} \\
        
        \pptHline
        \error{\textbf{\ulc{gGreen}{Ankara}-based organization}} launches new initiative to combat poverty in the region. & \error{Organization} & \correct{Location} \\
        
        \midrule
        \multicolumn{2}{l}{\hspace{-0.75em} \emph{MIT-Movie}} \\
        \midrule
        Who directed the \error{\textbf{latest \ulc{gGreen}{paranormal thriller}}} film to hit theaters? & \error{Title} & \correct{Genre} \\
        Can I watch a movie snippet of the \error{\textbf{newest \ulc{gGreen}{Marvel}} film}? & \error{Title} & \correct{Genre} \\
        Could you provide me with a trailer for the \error{\textbf{new \ulc{gGreen}{Tarantino}} film}? & \error{Title} & \correct{Director} \\
        
        \pptHline
        Could you recommend a popular \error{\textbf{\ulc{gGreen}{Sports underdog}} story} movie from the 1990s? & \error{Genre} & \correct{Plot} \\

        \pptHline
        What is the \ulc{gGreen}{\error{\textbf{highest}} Viewers' Rating} movie with Olaf in it? & \error{Review} & \correct{Viewers' Rating} \\

        \pptHline
        Who directed the \error{\textbf{\ulc{gGreen}{classic} film}} about unrequited love & \error{Viewers' Rating} & \correct{Genre} \\

        \addlinespace[0.125em] \hline \addlinespace[0.125em]
        I heard there's a new film coming out with \error{\textbf{\ulc{gGreen}{Zendaya} and \ulc{gGreen}{Tom Holland}}}, do you know if it's a PG-13 movie? 
            & \error{Actor} & \correct{Actor}, \correct{Actor} \\
        
        \pptHline
        what are some top-rated \error{\textbf{\ulc{gGreen}{M}/\ulc{gGreen}{PG}}} movies from the 90s & \error{MPAA Rating} & \correct{MPAA Rating}, \correct{MPAA Rating} \\
        
        \midrule
        \multicolumn{2}{l}{\hspace{-0.75em} \emph{MIT-Restaurant}} \\
        \midrule
        What time does the \error{\textbf{\ulc{gGreen}{sushi} place}} close? & \error{Restaurant Name} & \correct{Cuisine} \\

        Tell me a place where I can have a \error{\textbf{nice \ulc{gGreen}{brunch}}} on weekends. & \error{Amenity} & \correct{Hours} \\

        \addlinespace[0.125em] \hline \addlinespace[0.125em]
        What are the operating hours for the \error{\textbf{\ulc{gGreen}{upscale} \ulc{gGreen}{Italian} restaurant}} in the downtown area? 
            & \error{Restaurant Name} & \correct{Price}, \correct{Cuisine} \\

        What are the opening hours of the \error{\textbf{\ulc{gGreen}{nearest} \ulc{gGreen}{Thai} restaurant}}?
            & \error{Location} & \correct{Location}, \correct{Cuisine} \\

        \bottomrule
    \end{tabular}
    }
    \caption{
        Example LLM entity annotation samples that showcase annotation errors that are more malformed/current correction pipeline don't handle. 
        We \error{\textbf{boldface}} LLMs-annotated and wrong entity spans and \ulc{gGreen}{underline} the expected and correct entity spans (\emph{Span}). 
        In \emph{Type}, we show the corresponding entity types. 
        The 2 different annotations are also color coded (\textbf{\error{LLMs/wrong}} (\error{\xmark}) vs \correct{\ul{expected/correct}} (\correct{\cmark})). 
    }
    \label{tbl:annotation-errors-edge-case}
\end{table*}

\section{Future Work}
\label{apd:future-work}

\paragraph{More Analysis}
Given the initial success of ProgGen, it would be interesting to explore how does sample diversity contribute to performance. 
Notably, we hypothesized that distilling entity knowledge is more challenging in specific domains (\cref{sec:specific-domain-results}), 
experiments on how scaling the number of training samples influence downstream performance can provide more insights.

Cost analysis trading downstream performance with token querying cost is also interesting to investigate. 
For example, how would hyperparameters such as batch size (for sample generation and correction generation) influence the quality of the generated dataset? 


\paragraph{ProgGen Extensions}
Cursory manual examinations on the generated datasets found several cases where LLMs generate syntactically and semantically similar sentences 
(examples in Table \ref{tbl:similar-gen-sentence}). 
This implies that 
(1) the number of generated samples may \emph{not correlate linearly} with sample diversity and 
(2) the downstream training process is biased towards the ``duplicated'' groups of samples.
Future work can explore how to select a diverse subset of generated sentences for annotations and downstream training. 

In our main experiments (\cref{sec:experiment-results}), we found that Diversify Y variants work well, 
where the \emph{vanilla} variants perform the best across almost all datasets and the \emph{latent} variant leads to more robust performances. 
One extension can explore merging these variants via sampling to get the best of both worlds and 
further, generating entity pools conditioned on additional attribute dimensions from \emph{Diversify X}. 

Our current self-correction pipeline only handles the most common cases of entity annotation errors. 
More more nuanced annotation errors can be incorporated, including (ordered by occurrence from cursory inspection):
(1) both wrong span and wrong type, and
(2) two valid annotations. 
We show example wrong entity annotations of such case in Table \ref{tbl:annotation-errors-edge-case}. 
Our work mainly focused on reducing false positives since we found this is a more prevalent issue, 
future work can also explore looking at existing annotations given a sample and add additional annotations (reducing false negative). 

\paragraph{Better Annotations}
We note (1) the strong performance of Simple Prompt (Table \ref{tbl:main-results}) and that 
(2) Simple Prompt have relative less number of generated samples in downstream training due to de-duplication (Appendix \ref{apd:gen-data-diversity-stats}). 
Thus, we hypothesize that Simple-Prompt-generated NER samples may have more accurate annotations, despite less diverse, 
possibly due to an implicit self-consistency-like \citep{self-consistency} ``check'' where LLMs generate the same sentence multiple times. 
In particular, one of two cases will happen:
(1) all generated sentences have the same annotations and thus is part of the processed dataset, or 
(2) some of the generated annotations are conflicting and thus such samples are dropped from the processed dataset. 
We also observed that despite many wrong entity annotations GPT-3.5 make, 
GPT-3.5 may have a higher accuracy for easier tasks such as simply checking if terms like ``Nigerian President'' is a named entity. 
Thus, despite incorporating CoT and stage-wise annotation are shown non-trivial (Appendix \ref{apd:step-wise-gen}), 
future directions include exploring more variants, including
(1) combining re-annotation (Appendix \ref{apd:step-wise-gen}) with self-consistency-like \citep{self-consistency} majority voting to reduce noises 
(in entity class, demo \& input ordering) in LLM annotations and 
(2) least-to-most-promoting-like \citep{least2most} step-wise annotation correction to reduce bias. 

Additionally, the task of selecting potentially \emph{all} wrong entity annotations in the generated dataset in a \emph{single trial} may be too challenging. 
In light of the hypothesis that LLMs may not correct enough wrong annotations of the same failure class 
to significantly influence training (Appendix \ref{apd:self-correction-score-drop-reasons}), 
future work can explore propagating ``confident'' LLM corrections to the non-selected pool of entity annotations, leading to better annotated datasets.

\begin{table}[tb]
    \centering
    \scriptsize
    \begin{tabular}{ c *{4}{c} }
        \toprule
        & \thead{CoNLL\\-2003} & \thead{Wiki\\Gold} & \thead{MIT-\\Movie} & \thead{MIT-\\Restaurant} \\
        
        \midrule
        \multicolumn{5}{l}{\hspace{-0.75em} \emph{Duplicate}} \\
        \midrule
            \# & 1,349 / 179 / 147 & 7 & 31 / 1 & 1 / 0 \\
            \% & 9.8 / 5.5 / 4.2 & 0.4  & 0.3 / 0.0 & 0.0 / 0.0 \\
        
        \addlinespace[0.125em] \midrule \addlinespace[0.375em]
        \multicolumn{5}{l}{\hspace{-0.75em} \emph{Conflicting}} \\
        \midrule
            \# & 2 / 2 / 4 & 0 & 16 / 0 & 1 / 0 \\
            \% &  0.0 / 0.1 / 0.1 & 0.0 & 0.2 / 0.0 & 0.0 / 0.0 \\
        \bottomrule
    \end{tabular}
    \caption{Samples dropped from the original datasets due to \emph{duplicate} or \emph{conflicting}. 
        Separated by slash, we show the number of samples dropped for the original train/dev/test or train/test splits respectively if available. } \vspace{-0.5pc}
    \label{tbl:samples-dropped}
\end{table}

\section{Additional Implementation Details}
\label{apd:implementation}

\subsection{NER Sample \& Dataset Processing}
\label{apd:ner-sample-processing}
This section details how LLMs' free-form responses are parsed into structured NER samples for training. 

\paragraph{NER Sample Parsing}
We use regular expression (\texttt{re}) patterns to match and extract the generated sentences and list of entity name-type pairs. 
We drop malformed NER generations. 
Primary reasons for invalid NER samples include
(1) overlapping entity spans annotated, 
(2) unseen entity type annotated, 
and (3) annotated entity spans not found in the generated sentence. 
For the few-shot ICL baseline, to ensure a prediction is made for each sample in the original test set, 
we filter out the subset of malformed entity annotations in the generated annotation list 
(as opposed to filtering out the test sample entirely). 

Despite that LLMs are instructed to 
(1) annotate each multi-occurring named entities multiple times (Appendix \ref{apd:sample-format-search}) and 
(2) order the annotated spans by occurrence,
we found LLMs sometimes don't follow the instructions. 
We show one rare case where LLMs annotate a multi-occurring span with different annotations below: 

\begin{addmargin}[\parindent]{\parindent}
    \setlength\parindent{0pt}
    \small \vspace{0.5em}
    Sentence: \texttt{Can you tell me about the character Harry Potter from the Harry Potter series?}

    Named Entities: [\texttt{Harry Potter (Character), Harry Potter (Title)}]
\end{addmargin}

\noindent Thus, we pragmatically (1) duplicate single annotations for multi-occurring entities as much as it occurs (as a result, they have the same entity type), and 
(2) re-order the entity annotations to match the occurrence order in the generated sentence as needed. 

\paragraph{BIO Format Mapping} 
For downstream model training, we tokenize the sentences and entity spans by splitting on any whitespace and punctuations (and then process NER tags into BIO format). 
Conversely, since the original datasets we studied are in the BIO format and we need the demo samples in a natural-language format in prompts (Appendix \ref{apd:sample-format-search}), 
we detokenize the original samples via the \texttt{sacremoses} package.

\paragraph{Deduplication} 
We observed that the original datasets may contain duplicate samples or conflicting annotations. 
Thus, we filter out these samples. 
In particular, we
(1) keep only 1 copy of duplicate samples (same sentence and same annotations; \emph{duplicate}), and
(2) drop all conflicting samples, i.e. samples with the same sentence but different annotations (\emph{conflicting}). 
The number of dropped samples are summarized in Table \ref{tbl:samples-dropped}.

\subsection{Data Generation}
\label{apd:data-gen-setup}

\subsubsection{Diversify X}
\label{apd:data-gen-diverse-x-setup}

\paragraph{Generate Attribute Values}
Since ChatGPT generates long-form responses, the responses from attribute dimension generation also include additional attribute descriptions and example values. 
We may select high-quality generated descriptions and examples and include them in the attribute value generation prompt (example in Appendix \ref{apd:diverse-x-prompt}).

\paragraph{Construct Diversity Requirement}
The dataset-independent ``topic'' dimension is always included/sampled in the diversity requirements, 
for simplicity of incorporating with the \emph{Diversify X + Y} variant, passing the sampled topic value to the named entity pool. 

During attribute dimension sampling, we note that certain dimensions may have similar or conflicting values. 
For example, for MIT-Restaurant, attribute dimensions ``dietary restriction'' and ``special offering'' may both contain dietary needs. 
Thus, we group such ``conflicting'' attributes together and ensure at most one attribute dimension in the conflicting set is sampled. 
The conflicting sets for each dataset are shown in Table \ref{tbl:diverse-x-summary}. 

Another reason for keeping a small $R_x$ ($\in [1, 2]$; \cref{sec:diverse-x}) is that 
we expect limited instruction-following for GPT3.5 given a larger number of diversity requirements, 
where too many instructions presumably lead to lower-quality samples \citep{instruct-gpt}.

\subsubsection{Diversify Y}
\label{apd:data-gen-diverse-y-setup}
\paragraph{Generate Entities}
We generally use a smaller $M$ for \emph{latent} Diversify Y compared to the \emph{vanilla} variant. 
Intuitively, this is because with a more detailed requirement (conditioned on both an entity class and a domain-specific topic), there tend to be less relevant named entities. 

\paragraph{Construct Diversity Requirement}
For each entity class, we sample a few entities $\{e_i\}_1^{P_c}$ for each entity class $c$, where $P_c$ is sampled uniformly in $[0, 3]$ 
to ensure entity class balance. 
We then union all named entities sampled and takes a subset on the named entities as if in Bernoulli trials where we enforce 
the expected number of remaining entities to be $R_y$. 

In preliminary experiments, we found that including the entity type corresponding to each sampled entity 
in the entity inclusion requirement consistently yield worse performance (\textasciitilde 3\%). 
We hypothesize that 
(1) LLM-generated entity pool may not be accurate, i.e. named entities sampled may not belong to the given type, leading to incorrect requirements, and 
(2) specifying the entity type for spans that may belong to multiple entity types based on context (e.g. Burger as a Dish or a Cuisine entity) 
may limit the LLMs' capability in following the instruction. 
In both cases, LLMs may generate samples with wrong annotations, leading to worse downstream performance. 

\subsubsection{NER Sample Generation}
\label{apd:data-gen-sample-gen-setup}
In our API calls, we use temperature $t=1$ and parameter top $p = 1.0$ unless otherwise specified. 
For a fair comparison, for each approach, we ensure LLMs generated 1.8K samples. 
We filter out malformed samples LLMs generated and de-duplicate the remaining, valid samples (Appendix \ref{apd:ner-sample-processing}),
resulting in different number of NER samples in the NER dataset for training. 
For Simple Prompt, we request ($L = 50$) samples in each prompt (thus \textasciitilde 36 API calls) to further save token querying cost (since less prompt tokens). 
We found that GPT-3.5 frequently fails to generate 50 samples requested so we made additional API calls until 1.8K samples are found (without checking for validity and duplication). 
For all diversity setups, we request 3 samples in each prompt to sample more diverse configurations, making \textasciitilde 600 configurations total for each variant. 
We use $\mathrm{E} \left[ R_y \right] = 4.5$ for MIT-Movie and $\mathrm{E} \left[ R_y \right] = 1.5$ for other datasets in our experiments, which empirically performs well in early experiments. 

Few-shot ICL experiments are greedily decoded ($t=0$). 

\subsubsection{Self-Correction}
For LLM self-correction, each prompt includes 3 samples to correct and we use greedy decoding ($t=0$). 
When crafting the self-correction annotation instructions and demos, 
we generally follow the rule below: 
1) if a failure case appears more than 5 times, we include a demo correction sample, and 
2) if a failure case appears more than 3 times, we summarize the error case in the annotation instruction. 
We provide no more than 6 error cases in the instruction and no more than 6 demos. 
The typical number of wrong-annotation cases in instruction and demos are about 3. 

Less common failure cases are ignored to prevent overly tuning for each generated dataset. 
We further note that 
(1) ChatGPT3.5 can correct many ``easier'' wrong entity annotations 0-shot, i.e. without any instruction and demo, and 
ChatGPT3.5 is biased sometimes and its capacity and instruction following capabilities are limited, 
where (2) a fraction of LLM Self-Corrections are invalid and (3) additional demos may not lead to additional valid corrections.

\subsection{Downstream Model Training}
The maximum token length for each sequence is 144. 
We use the standard training practice and cross entropy loss in PyTorch.
All models are trained with NVIDIA GeForce RTX 2080 Ti GPUs. 

We use DeBERTa and learning rate of $4e^{-5}$ due to cursory hyperparameter search via supervised learning. 
We searched over BERT, RoBERTa and DeBERTa \citep{bert, roberta, deberta-v3} and learning rate in [$1e^{-5}$, $2e^{-5}$, $3e^{-5}$, $4e^{-5}$, $5e^{-5}$, $5e^{-6}$]
and found this setup consistently yield relatively high F1 scores across all 4 datasets studied.

\subsection{Special Handling} 
CoNLL-2003 and WikiGold both contain over 1\% negative samples where as MIT-Movie and MIT-Restaurant don't. 
Thus, we include a negative demo sample for CoNLL-2003 and WikiGold (\cref{sec:ori-dataset}). 
We also add an explicit instruction in the data generation prompt for these datasets in case no named entities is found in the generated sentences (as shown in Table \ref{tbl:prompt}). 
For WikiGold, we used the same train/dev/test split as in \citet{wikigold-split}. 

For MIT-Movie and MIT-Restaurant, 
we lowercase all generated NER samples before (de-duplication and) downstream training, since 
(1) the original dataset samples are all in lowercase, and 
(2) lowercasing leads to noticeable gains (\textasciitilde 10\%). 
For MIT-Movie entity pool generation, we found that GPT-3.5-generated entities don't belong to the respective entity classes for several classes, 
so we additionally provided entity definitions and some demos in the prompt as necessary.

\begin{table}[tb]
    \vspace{-0.5pc}
    \centering 
    \resizebox{0.85\linewidth}{!}{
    \begin{tabular}{ c l c c c }
        \toprule \thead{Dataset} &     \thead{Attr. Dim.} & \thead{$V$} &  \thead{$P$} & \thead{$T$} \\
        \midrule
            \multirow{2}{*}{\makecell{CoNLL\\-2003}} 
                & News Topic* & 44 & - & \multirow{2}{*}{1.0K} \\
                & Writing Style & 22 & 0.4 \\

            \midrule \addlinespace[0.375em]
            \multirow{2}{*}{\makecell{Wiki\\Gold}}
                & Topic*  &  46 &  - & \multirow{2}{*}{1.2K} \\
                & Writing Style & 26 & 0.4 \\

            \midrule \addlinespace[0.375em]
            \multirow{4}{*}[-0.125em]{\makecell{MIT-\\Movie}}
                & Query Category*  & 10 & - & \multirow{4}{*}[-0.125em]{450} \\
                
                \addlinespace[0.25em]
                & User Demographic  & 18 & 0.3 \hspace{-0.25em}\rdelim\}{3}[0em]{0em}[] & \\
                & User Emotion  & 17 & 0.3 & \\
                & Query Language  & 9 & 0.3 & \\

            \midrule \addlinespace[0.375em]
            \multirow{7}{*}[-0.25em]{\makecell{MIT-\\Restaurant}} 
                & Meal Category* & 10 & - & \multirow{7}{*}[-0.25em]{12K} \\
                
                \addlinespace[0.25em]
                & Price Range & 5 & 0.2 \hspace{-0.25em}\rdelim\}{3}[0em]{0em}[] & \\
                & Ambiance & 20 & 0.2  & \\
                & Service Mode & 4 & 0.1 & \\

                \addlinespace[0.25em]
                & User Demographic & 20 & 0.1 \hspace{-0.25em}\rdelim\}{3}[0em]{0em}[] & \\
                & Dietary Restriction & 10 & 0.1 & \\
                & Special Offering & 8 & 0.1 & \\

        \bottomrule
    \end{tabular}
    }
    \caption{
        Diversify X Requirement Configurations Summary Statistics. 
        $V$ and $P$ and refer to the number of unique attribute values and sampling probability for each attribute dimension respectively. 
        $T$ refers to the total number of unique diversity configurations. 
        The dataset-independent ``topic'' attribute dimensions have trailing asterisks* and are always included in the diversity requirements. 
        Conflicting attribute dimensions are grouped with braces$\}$ (Appendix \ref{apd:data-gen-diverse-x-setup}). 
    }
    \label{tbl:diverse-x-summary}
\end{table}

\begin{table}[tb]
    \centering 
    \resizebox{1\columnwidth}{!}{
    \begin{tabular}{ c m{0.48\textwidth} c}
        \toprule \thead{Var.} & \thead{\#unique entities by class} & \thead{$N$} \\
        \midrule
        \multicolumn{3}{l}{\hspace{-0.75em} \emph{CoNLL-2003}} \\
        \midrule
        van. & \{person: 132; location: 86; organization: 132\} & 350 \\
        \addlinespace[0.25em] \arrayrulecolor{gLightGrey} \cline{1-3} \arrayrulecolor{black} \addlinespace[0.5em] 
        lat. & \{person: 45.1, location: 46.7, organization: 43.5\} & 6.0K \\

        \midrule
        \multicolumn{3}{l}{\hspace{-0.75em} \emph{WikiGold}} \\
        \midrule
            van. & \{person: 119; location: 310; organization: 282\} & 711 \\
            \addlinespace[0.25em] \arrayrulecolor{gLightGrey} \cline{1-3} \arrayrulecolor{black} \addlinespace[0.5em] 
            lat. & \{person: 18.1; location: 26.8; organization: 22.0\} & 3.0K \\

        \midrule
        \multicolumn{3}{l}{\hspace{-0.75em} \emph{MIT-Movie}} \\
        \midrule
            van. & \{Title: 184; Viewers' Rating: 257; Year: 98; \par ~~Genre: 133; Director: 120; MPAA Rating: 40; \par ~~Plot: 168; Actor: 110; Trailer: 66; \par ~~Song: 125; Review: 249; Character: 170\} & 1.7K \\
            \addlinespace[0.25em] \arrayrulecolor{gLightGrey} \cline{1-3} \arrayrulecolor{black} \addlinespace[0.5em] 
            lat. & \{Title: 100.9; Viewers' Rating: 127.2; Year: 98.5; \par ~~Genre: 70.2; Director: 92.1; MPAA Rating: 13.3; \par ~~Plot: 90.0; Actor: 84.3; Trailer: 30.0; \par ~~Song: 78.4; Review: 123.7; Character: 124.1\} & 10K \\

        \midrule
        \multicolumn{3}{l}{\hspace{-0.75em} \emph{MIT-Restaurant}} \\
        \midrule
            van. & \{Restaurant Name: 199; Amenity: 170; \par ~~Cuisine: 96; Dish: 188; Hours: 221; \par ~~Location: 221; Price: 284; Rating: 308\} & 1.7K \\
            \addlinespace[0.25em] \arrayrulecolor{gLightGrey} \cline{1-3} \arrayrulecolor{black} \addlinespace[0.5em] 
            lat. & \{Restaurant Name: 35.5; Amenity: 33.6; \par ~~Cuisine: 19.8; Dish: 32.5; Hours: 37.8; \par ~~Location: 26.9; Price: 33.8; Rating: 34.4\} & 3.9K \\
        \bottomrule
    \end{tabular}
    }
    \caption{
        Diversify Y Requirement Configurations Summary Statistics. 
        \emph{Var.} refers to the Diversify Y variants, one of \emph{van.} or \emph{lat.} for \emph{vanilla} and \emph{latent} respectively. 
        $N$ refers to the total number of entities in the entity pool. 
        For the latent variant of Diversity Y, we show number of entities values averaged by the number of ``topic'' dimension values. 
    } 
    \label{tbl:diverse-y-summary}
\end{table}

\begin{table}[tb]
    \centering
    \resizebox{1\columnwidth}{!}{
    \begin{tabular}{ l *{5}{c} }
        \toprule
        \thead{Setup} & $N$ & $|V|$ & $|L|$ & $|E|$ & $\overline{E}$ \\
        
        \midrule
        \multicolumn{6}{l}{\hspace{-0.75em} \emph{CoNLL-2003}} \\
        \midrule
        Simple Prompt & 1.6K & 2.5K & 8.7 & 549 & 1.3 \\
        Diversify X & 1.7K & 3.6K & 11.2 & 972 & 1.5 \\
        Diversify Y (vanilla) & 1.7K & 3.0K & 10.5 & 689 & 1.5 \\
        Diversify Y (latent) & 1.7K & 3.6K & 11.0 & 940 & 1.5 \\
        Diversify X+Y & 1.7K & 4.2K & 13.2 & 1.0K & 1.5 \\

        \midrule
        \multicolumn{6}{l}{\hspace{-0.75em} \emph{WikiGold}} \\
        \midrule
        Simple Prompt & 1.3K & 3.1K & 15.2 & 894 & 2.2 \\
        Diversify X & 1.6K & 5.3K & 20.9 & 975 & 1.7 \\
        Diversify Y (vanilla) &     1.6K & 5.1K & 17.6 & 1.4K & 1.9 \\
        Diversify Y (latent) &      1.7K & 5.7K & 18.4 & 1.6K & 1.9 \\
        Diversify X+Y &             1.7K & 6.3K & 22.7 & 1.4K & 1.7 \\

        \midrule
        \multicolumn{6}{l}{\hspace{-0.75em} \emph{MIT-Movie}} \\
        \midrule
        Simple Prompt       &  1.7K &   618 &   9.7 & 896 &  1.8 \\
        Diverse-X           &  1.4K &   1.3K &   14.1 & 940 &  2.0 \\
        Diverse-Y (vanilla) &  1.7K &   1.9K &   12.3 & 1.5K &  2.1 \\
        Diverse-Y (seeded)  &  1.7K &   2.0K &   12.1 & 1.5K &  2.1 \\
        Diverse-X+Y         &  1.5K &   2.2K &   16.0 & 1.5K &  2.1 \\

        \midrule
        \multicolumn{6}{l}{\hspace{-0.75em} \emph{MIT-Restaurant}} \\
        \midrule
        Simple Prompt       &   1.7K &   704 &  112 & 10.2 &  2.2 \\
        Diverse-X           &   1.6K &   638 &  191 & 12.6 &  2.8 \\
        Diverse-Y (vanilla) &   1.7K &   1.1K &   90 & 12.4 &  2.4 \\
        Diverse-Y (seeded)  &   1.7K &   1.0K &   86 & 12.5 &  2.5 \\
        Diverse-X+Y         &   1.6K &   1.0K &  145 & 13.6 &  2.9 \\

        \bottomrule
    \end{tabular}
    }
    \caption{
        Generated NER Dataset Statistics. 
        $N$ refers to the number of processed NER samples. 
        $|V|$ refers to the vocabulary size. 
        $\overline{L}$ refers to the average token length of generated sentences. 
        $|E|$ refers to number of distinct named entities. 
        $\overline{E}$ refers to the average number of named entities for all sentences with at least 1 named entity annotated. 
        Tokenization via splitting on any whitespace and punctuation. 
    }
    \label{tbl:gen-data-stats}
\end{table}

\begin{table}[tb]
    \centering
    \resizebox{1\columnwidth}{!}{
    \begin{tabular}{ l *{5}{c} }
        \toprule \multirow{2}{*}[-0.2em]{\thead{Setup}} & \multicolumn{3}{c}{\thead{Correction Type}} & \multirow{2}{*}[-0.2em]{\thead{\thead{\%}}} & \multirow{2}{*}[-0.2em]{\thead{\#}} \\
        \addlinespace[0.2em] \cline{2-4} \addlinespace[0.2em] 
        & \thead{NA} & \thead{Span} & \thead{Type} & & \\
        
        \midrule
        \multicolumn{6}{l}{\hspace{-0.75em} \emph{CoNLL-2003}} \\
        \midrule
        Simple Prompt & 41 & 14 & 12 & 17.4 & 386 \\
        Diversify X & 107 & 5 & 15 & 27.3 & 466 \\
        Diversify Y (vanilla) & 42 & 8 & 12 & 15.8 & 392 \\
        Diversify Y (latent) & 53 & 10 & 21 & 19.1 & 439 \\
        Diversify X+Y & 83 & 9 & 11 & 28.1 & 366 \\

        \midrule
        \multicolumn{6}{l}{\hspace{-0.75em} \emph{WikiGold}} \\
        \midrule
        Simple Prompt & 172 & 8 & 21 & 36.8 & 546 \\
        Diversify X & 179 & 8 & 4 & 44.0 & 434 \\
        Diversify Y (vanilla) & 95 & 9 & 17 & 20.1 & 601 \\
        Diversify Y (latent) & 88 & 8 & 18 & 20.6 & 553 \\
        Diversify X+Y & 92 & 4 & 3 & 20.7 & 479 \\

        \midrule
        \multicolumn{6}{l}{\hspace{-0.75em} \emph{MIT-Movie}} \\
        \midrule
        Simple Prompt       &  125 &   18 &   59 & 35.0 &  577 \\
        Diverse-X           &  127 &   34 &   72 & 40.5 &  576 \\
        Diverse-Y (vanilla) &  131 &   47 &   85 & 36.3 &  725 \\
        Diverse-Y (seeded)  &  156 &   37 &   70 & 37.5 &  701 \\
        Diverse-X+Y         &  169 &   27 &   64 & 42.1 &  618 \\

        \midrule
        \multicolumn{6}{l}{\hspace{-0.75em} \emph{MIT-Restaurant}} \\
        \midrule
        Simple Prompt       &   41 &   44 &  112 & 26.5 &  744 \\
        Diverse-X           &   29 &   62 &  191 & 31.0 &  909 \\
        Diverse-Y (vanilla) &   49 &   67 &   90 & 25.0 &  824 \\
        Diverse-Y (seeded)  &   43 &   59 &   86 & 22.7 &  829 \\
        Diverse-X+Y         &   36 &   50 &  145 & 24.7 &  935 \\

        \bottomrule
    \end{tabular}
    }
    \caption{
        Distribution of LLM Correction Types grouped by diversity variant. 
        \crtNA{NA}, \crtSpan{Span}, \crtType{Type} refers to \crtNA{dropping entity annotation}, \crtSpan{correcting entity span}, and \crtType{correcting entity type} respectively. 
        \% refers to the percentage of entity annotations LLMs corrected, among total number of entity annotations prompted. 
        \# refers to the total number of entity annotations prompted. 
    }
    \label{tbl:correction-dist-by-setup}
\end{table}

\begin{table}[tb]
    \centering
    \resizebox{0.95\columnwidth}{!}{
    \begin{tabular}{ l *{5}{c} }
        \toprule \multirow{2}{*}[-0.2em]{\thead{Entity Type}} & \multicolumn{3}{c}{\thead{Correction Type}} & \multirow{2}{*}[-0.2em]{\thead{\thead{\%}}} & \multirow{2}{*}[-0.2em]{\thead{\#}} \\
        \addlinespace[0.2em] \cline{2-4} \addlinespace[0.2em] 
        & \thead{NA} & \thead{Span} & \thead{Type} & & \\

        \midrule
        \multicolumn{6}{l}{\hspace{-0.75em} \emph{CoNLL-2003}} \\
        \midrule
        person & 102 & 30 & 23 & 30.8 & 504 \\
        location & 97 & 12 & 19 & 18.5 & 693 \\
        organization & 127 & 4 & 29 & 18.8 & 852 \\
        \rowcolor{lightCyan} \emph{all} & 326 & 46 & 71 & 21.6 & 2,049 \\

        \midrule
        \multicolumn{6}{l}{\hspace{-0.75em} \emph{WikiGold}} \\
        \midrule
        person & 81 & 14 & 8 & 21.5 & 479 \\
        location & 286 & 17 & 39 & 28.5 & 1,199 \\
        organization & 259 & 6 & 16 & 30.1 & 935 \\
        \rowcolor{lightCyan} \emph{all} & 626 & 37 & 63 & 27.8 & 2,613 \\

        \midrule
        \multicolumn{6}{l}{\hspace{-0.75em} \emph{MIT-Movie}} \\
        \midrule
        Title           &   29 &   14 &   29 & 16.1 &  446 \\
        Viewers' Rating &  105 &    6 &   71 & 52.4 &  347 \\
        Year            &   87 &    6 &    6 & 43.6 &  227 \\
        Genre           &   26 &   69 &   66 & 27.4 &  588 \\
        Director        &   50 &    1 &    4 & 34.8 &  158 \\
        MPAA Rating     &   54 &    7 &   12 & 63.5 &  115 \\
        Plot            &  151 &   10 &   26 & 52.5 &  356 \\
        Actor           &   22 &    0 &    9 & 27.4 &  113 \\
        Trailer         &   29 &   34 &    9 & 34.3 &  210 \\
        Song            &   80 &    4 &   15 & 61.1 &  162 \\
        Review          &   24 &   11 &   94 & 39.4 &  327 \\
        Character       &   51 &    1 &    9 & 41.2 &  148 \\
        \rowcolor{lightCyan} \emph{all}      &  708 &  163 &  350 & 38.2 & 3,197 \\

        \midrule
        \multicolumn{6}{l}{\hspace{-0.75em} \emph{MIT-Restaurant}} \\
        \midrule
        Restaurant Name &    3 &    4 &   33 & 25.2 &  159 \\
        Amenity         &   85 &   92 &  187 & 28.1 & 1,297 \\
        Cuisine         &   20 &   72 &  109 & 24.2 &  830 \\
        Dish            &    3 &   21 &   30 & 21.0 &  257 \\
        Hours           &   32 &    4 &   19 & 13.5 &  408 \\
        Location        &    6 &   70 &   73 & 30.2 &  494 \\
        Price           &   16 &    7 &  102 & 45.6 &  274 \\
        Rating          &   33 &   12 &   71 & 22.2 &  522 \\
        \rowcolor{lightCyan} \emph{all}      &  198 &  282 &  624 & 26.0 & 4,241 \\
        \bottomrule
    \end{tabular}
    }
    \caption{Distribution of LLM Correction Types grouped by entity type.} \vspace{-0.5pc}
    \label{tbl:correction-dist-by-entity-type}
\end{table}

\begin{figure*}[tb]
    \centering

    \begin{minipage}[c]{0.44\textwidth}
        \begin{subfigure}[b]{0.45\textwidth}
            \includegraphics[width=1\textwidth]{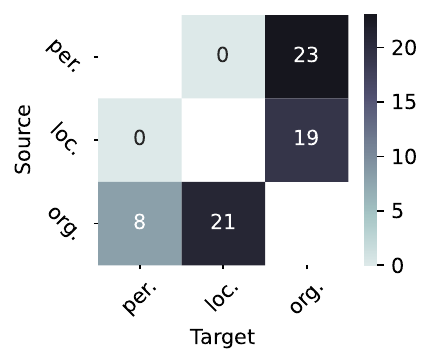} \vspace{-1.5pc}
            \caption{CoNLL2003}
        \end{subfigure}
        \hfill
        \begin{subfigure}[b]{0.45\textwidth}
            \includegraphics[width=1\textwidth]{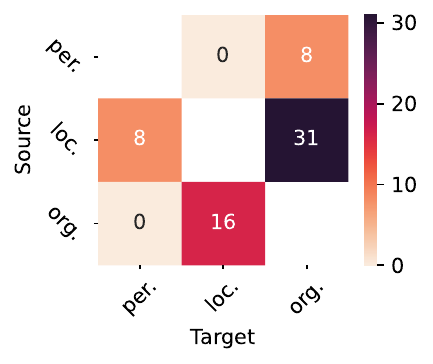} \vspace{-1.5pc}
            \caption{WikiGold} 
        \end{subfigure}

        \vspace{0.5pc}

        \begin{subfigure}[b]{1\textwidth}
            \centering
            \includegraphics[width=0.9\textwidth]{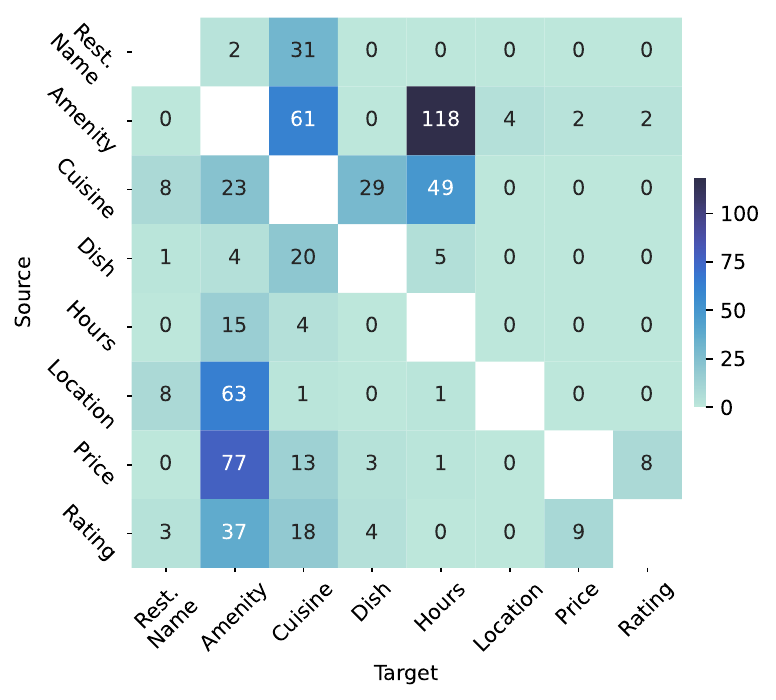} \vspace{-0.5pc}
            \caption{MIT-Restaurant} 
        \end{subfigure}
    \end{minipage}
    \begin{minipage}[c]{0.52\textwidth}
        \begin{subfigure}[c]{1\textwidth}
            \includegraphics[width=1\textwidth]{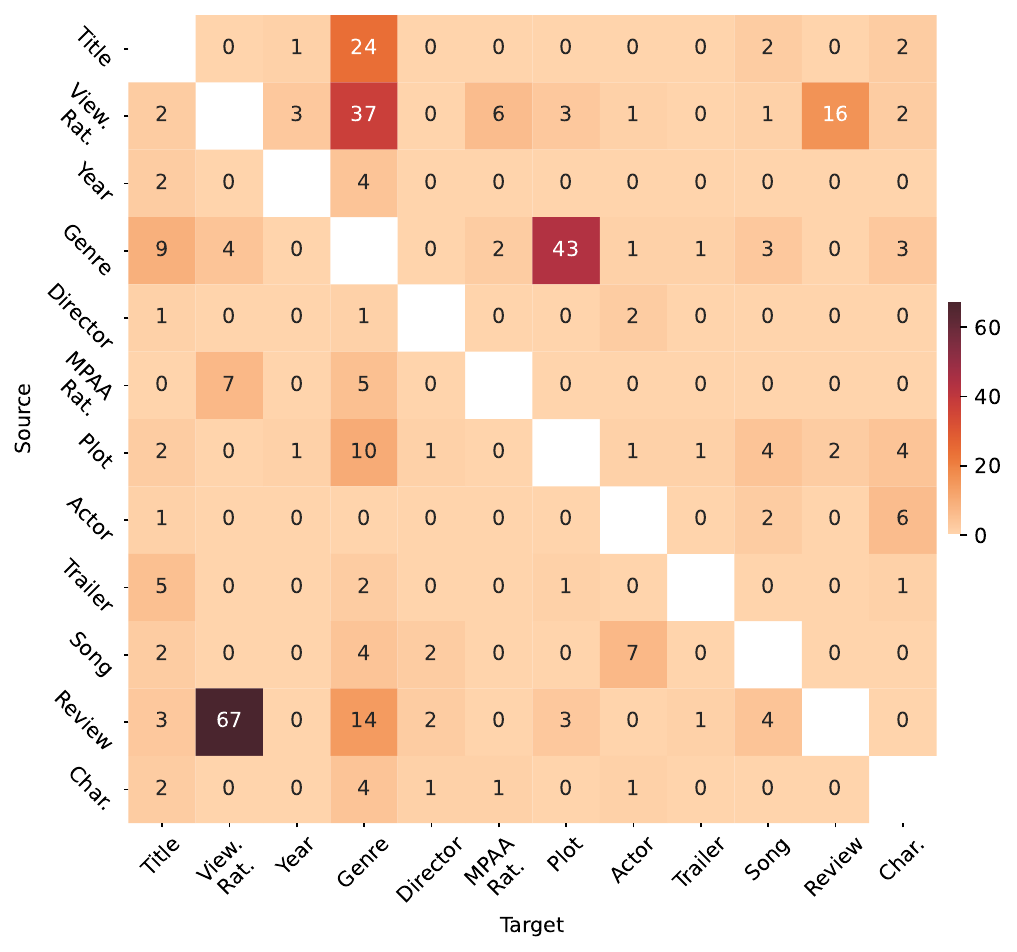} \vspace{-1.5pc}
            \caption{MIT-Movie}
        \end{subfigure}
    \end{minipage}
    \vspace{-0.5pc}
    \caption{LLM Entity Type Self-Correction counts by dataset.}
    \label{fig:type-correction-counts}
\end{figure*}

\section{Additional Data Statistics}
\label{apd:data-stats}

\subsection{Diversity Requirements}
\label{apd:diversity-stats}
We show summary statistics of the diversity requirement configurations for \emph{Diversify X} and \emph{Diversify Y} respectively in Tables \ref{tbl:diverse-x-summary} and \ref{tbl:diverse-y-summary}.

\subsection{Sample Generation}
\label{apd:gen-data-diversity-stats}
We show summary statistics of generated datasets (before LLM Self-Correction) for Simple Prompt and our diversity approaches in Table \ref{tbl:gen-data-stats}. 
We observe consistent trends where we make the following (unsurprising) observations: 
(1) Diversify X leads to longer generated sentences, indicating more tokens are needed to satisfy the semantic requirements, 
(2) Diversify Y shows increased number of distinct named entities, 
(3) Simple Prompt in general domain have less samples processed, mostly due to de-duplication, and 
(4) Explicit diversity requirements yield improved diversity where Diversify X+Y yields the highest diversity as expected.

\subsection{LLM Self-Correction}
\label{apd:correction-stats}
For each dataset, we show summary statistics for distribution of LLM correction types, across different diversity setups and across entity classes, 
in Tables \ref{tbl:correction-dist-by-setup} and \ref{tbl:correction-dist-by-entity-type}.
\crtType{Entity type corrections} to ``other'' are considered as \crtNA{entity annotation drops}. 
For \crtType{entity type corrections} to relevant entity types within the dataset, we show the count of corrections in 
Figure \ref{fig:type-correction-counts}.



\section{Additional Prompts and Samples}
\label{apd:templates-n-examples}

In this section, we show (1) prompt templates in \ref{apd:prompt-templates}, 
(2) some prompt and response samples for the CoNLL-2003 dataset for space constraints in \ref{apd:prompt-eg-conll2003}, 
(3) a small subset of the generated diversity configurations for each dataset in \ref{apd:diversify-config-values}, and
(4) selected 1-shot demo samples and a few generated NER samples for each dataset and each diversity variant in \ref{apd:generated-ner-samples}. 
The full prompts and LLM responses for (1) diverse requirement generation, (2) NER sample generation and (3) entity correction generation, 
and the corresponding processed data are all available in our 
repository\footnote{\url{https://github.com/StefanHeng/ProgGen/reproduce}}.

\subsection{Prompt Templates}
\label{apd:prompt-templates}

\subsubsection{Diversify Sentence}
\label{apd:diverse-x-prompt}

\paragraph{Generate Attribute Dimensions} 

\footnotesize  
\prompt{
    What do you think are important attributes to generate some diverse \tplVar{domain sample description}? 
    Examples: \tplVar{example attributes}... 
    
    Let's think step by step.
}

\paragraph{Generate Attribute Values} 

\prompt{
    Suppose you are \tplVar{domain description}. 
    List \tplVar{\#generate} different \tplVar{attribute dimension} for \tplVar{domain sample description}. 
    Some examples are \tplVar{example attribute values}.
}

\subsubsection{Diversify Entities}
\paragraph{Generate Entities (vanilla)} 

\prompt{
    Suppose you are \tplVar{domain description}. 
    Please generate \tplVar{\#generate} diverse named entities in \tplVar{domain sample description} that can be categorized as \tplVar{entity class}.
}

\paragraph{Generate Entities (latent)} 

\prompt{
    Suppose you are \tplVal{domain + latent topic description}. 
    Please generate \tplVar{\#generate} diverse named entities in \tplVar{domain sample description} that can be categorized as \tplVar{entity class}.}

\subsubsection{Few-Shot ICL baseline}


\footnotesize
\prompt{Suppose you are \tplVar{domain description}. You are given \tplVar{domain sample description}. Please identify all named entities occurred that belong to one of the following entity types:

\tplVar{entity types in the dataset}.

Please list such named entities with the corresponding entity types on the following line, in the order of occurrence.\nl


Here are some example annotations for your reference. Please follow this format.\nl


Examples:\nl


\tplVar{(sentence, entity annotations) demo samples}\nl


\hspace{0.05em}\verb|---|\nl


Please annotate the following \tplVar{domain sample prefix}:\nl

\tplVar{domain sample prefix}: "\tplVar{sample}"}


~\par

\subsection{Prompt Examples for CoNLL-2003}
\label{apd:prompt-eg-conll2003}

\subsubsection{Diversify Sentence}
\paragraph{Generate Attribute Dimensions}
~\par

\Prompt

\footnotesize
\prompt{
    What do you think are important attributes to generate some diverse \tplVal{news from Reuters}? Examples: \tplVal{topic, writing style}...
    
    Let's think step by step. 
}

~\par

\Response

\prompt{

Creating diverse news content, especially for a global news organization like Reuters, involves several key attributes. Let's break them down step by step:

\begin{enumerate}
    \item **Topic Selection**: Covering a wide range of topics is crucial. This includes politics, economics, technology, health, environment, culture, sports, and human interest stories. The goal is to cater to varied interests and provide a holistic view of global events.
    
    \item **Geographical Diversity**: Ensuring that news covers different regions of the world is important. This involves not just focusing on major powers like the US, China, or the EU, but also giving attention to less covered areas like Africa, South America, and smaller Asian and European countries.
    
    \item **Writing Style**: Reuters is known for its straightforward, factual reporting style. However, incorporating different writing styles like in-depth analysis, opinion pieces, and feature stories can enrich the content.
    
    \item **Source Variety**: Utilizing a wide array of sources, from government officials to local activists, experts in various fields, and ordinary citizens, can provide a more comprehensive perspective on issues.
    
    \item **Multimedia Content**: Incorporating videos, podcasts, and interactive graphics can enhance the storytelling and appeal to a broader audience.
    
    \item **Inclusivity and Representation**: Reflecting diverse voices in terms of ethnicity, gender, age, socio-economic status, and cultural backgrounds in both the news stories and among the journalists.
    
    \item **Fact-Checking and Accuracy**: Maintaining high standards of journalism by rigorously fact-checking information and presenting unbiased, accurate news.
    
    \item **Language Accessibility**: Offering news in multiple languages can reach a broader global audience.
    
    \item **Investigative Journalism**: Going beyond surface-level reporting to uncover deeper truths, especially in areas like corruption, human rights abuses, and environmental issues.
    
    \item **Audience Engagement**: Encouraging reader interaction through comments, social media, and possibly reader polls or Q\&A sessions.
    
    \item **Ethical Reporting**: Upholding journalistic ethics by respecting privacy, avoiding sensationalism, and reporting responsibly on sensitive issues.
    
    \item **Technology and Innovation**: Leveraging new technologies like AI, VR, or AR for more immersive and interactive news experiences.
\end{enumerate}

By focusing on these attributes, Reuters or any other news organization can produce diverse, engaging, and high-quality news content that appeals to a global audience.
}

~\par

\paragraph{Generate Attribute Values}
~\par

\Prompt

\prompt{
    Suppose you are \tplVal{a news writer for Reuters}. List \tplVal{30} diverse \tplVal{news categories} for \tplVal{news articles}. Some examples are \tplVal{[health, local stories, international events, environment, business, sports, politics, economics, breaking news, culture, science, technology, human interest stories]}.
}

~\par

\Response

\prompt{






























\begin{enumerate}
    \item World News
    \item Politics
    \item Technology
    \item Sports
    \item Entertainment
    \item Health
    \item Business
    \item Environment
    \item Breaking News
    \item Human Interest Stories
    \item Local Stories
    \item Economics
    \item International Events
    \item Science
    \item Education
    \item Travel
    \item Crime
    \item Fashion
    \item Food
    \item Religion
    \item Art
    \item Music
    \item Film
    \item Literature
    \item Automotive
    \item Lifestyle
    \item Weather
    \item Wildlife
    \item Education
    \item Immigrant Stories
\end{enumerate}}

~\par

\Prompt

\prompt{
    Suppose you are \tplVal{a news writer for Reuters}. List \tplVal{8} different \tplVal{writing styles} for \tplVal{news articles}. Some examples are \tplVal{[op-eds, opinion pieces, feature stories, in-depth analysis]}.
}

~\par

\Response

\prompt{








\begin{enumerate}
    \item Hard news reporting
    \item Investigative journalism
    \item Informative news articles
    \item Human interest stories
    \item Broadcast news
    \item Long-form journalism
    \item Photojournalism
    \item Data-driven journalism
\end{enumerate}}

\footnotesize
~\par

\subsubsection{Diversify Entities}

\paragraph{Generate Entities (vanilla)}
~\par

\Prompt

\prompt{
    Suppose you are \tplVal{a news writer}. Please generate \tplVal{50} diverse named entities in \tplVal{news articles} that can be categorized as \tplVal{location}.
}

~\par

\Response

\prompt{

\begin{enumerate}
    \item New York City
    \item Eiffel Tower
    \item Sahara Desert
    \item Mount Everest
    \item Great Wall of China
    \item Rio de Janeiro
    \item Sydney Opera House
    \item Taj Mahal
    \item Amazon Rainforest
    \item Grand Canyon
    \item Tokyo
    \item Moscow
    \item London
    \item Paris
    \item Rome
    \item Antarctica
    \item Hawaii
    \item Dubai
    \item Bali
    \item Barcelona
    \item African Savannah
    \item Las Vegas
    \item Machu Picchu
    \item Venice
    \item Yellowstone National Park
    \item Cairo
    \item Galapagos Islands
    \item Himalayas
    \item Easter Island
    \item The Bahamas
    \item Mount Kilimanjaro
    \item Norwegian Fjords
    \item Great Barrier Reef
    \item Prague
    \item Istanbul
    \item Vatican City
    \item San Francisco
    \item Cape Town
    \item Athens
    \item Sydney
    \item Bangkok
    \item Buenos Aires
    \item Montreal
    \item Vancouver
    \item Toronto
    \item New Delhi
    \item Cairo
    \item Singapore
    \item Sydney
    \item Shanghai
\end{enumerate}
}

\footnotesize  

~\par

\paragraph{Generate Entities (latent)} 

~\par

\Prompt

\prompt{
    Suppose you are \tplVal{a news writer for Entertainment}. Please generate \tplVal{15} diverse named entities in \tplVal{news articles} that can be categorized as \tplVal{person}.
}

~\par

\Response

\prompt{

\begin{enumerate}
    \item Tom Hanks
    \item Beyoncé
    \item Brad Pitt
    \item Rihanna
    \item Will Smith
    \item Taylor Swift
    \item Leonardo DiCaprio
    \item Oprah Winfrey
    \item Jennifer Lopez
    \item Chris Hemsworth
    \item Meryl Streep
    \item Dwayne "The Rock" Johnson
    \item Kim Kardashian
    \item Hugh Jackman
    \item Lady Gaga
\end{enumerate}
}

~\par

\subsubsection{Few-Shot ICL baseline}


\prompt{Suppose you are \tplVal{a news writer}. You are given \tplVal{a sentence from a news story}. Please identify all named entities occurred that belong to one of the following entity types:

\tplVal{[person, location, organization]}.

Please list such named entities with the corresponding entity types on the following line, in the order of occurrence.\nl

\txtRule

Here are some example annotations for your reference. Please follow this format.\nl


Examples:\nl

    
    
    
    
    
    
    
    
    
    

1. \tplNER{BASEBALL - DODGERS WIN FIFTH STRAIGHT.}{DODGERS (organization)}

2. \tplNER{With 2.5 percent gross domestic product growth expected for 1996, new job growth should slowly lower the unemployment rate over the rest of the year.}{}

3. \tplNER{A.de Silva not out 49}{A.de Silva (person)}

4. \tplNER{Saudi Arabia executes Pakistani man.}{Saudi Arabia (location)}


\pptSep

\txtRule


Please annotate the following sentence:\nl


Sentence: 
"\tplVal{Japan began the defence of their Asian Cup title with a lucky 2-1 win against Syria in a Group C championship match on Friday.}"
}

~\par

\subsubsection{LLM Self-Correction}
\label{apd:correction-prompts}









\normalsize
We show annotation instructions and demo pools for the 3 entity classes: Person, Location and Organization. 
These are for illustrative purposes, encompassing annotation errors in all diversity variants. 
Please refer to our repository for the illustrative prompts for all datasets and the exact prompts used for each generated dataset. 

\paragraph{Person}

~\par

\footnotesize
\prompt{Here are \tplVal{3 sentences from news articles}. Your task is to classify whether the span of text is a named entity of type \tplVal{person}.

In particular, for each query with a span of text enclosed in double braces, please classify the enclosed span of text in context of the query into one of the following categories:

- (A). The span is a named entity of type \tplVal{person}

- (B). The span contains a named \tplVal{person} entity but the span boundary is not precise

- (C). The span is a named entity but the category is not \tplVal{person}

- (D). The span is not a named entity

If the label is (B), you should provide the correct span boundary.

If the label is (C), you should provide the correct entity type. Pick an entity type from the following list:

[\tplVal{location, organization}, other].\nl

\txtRule

{\color{gDarkBlue}
    A named person entity must be the name of a person. Only first names, last names, and full names are considered named person entities.
    
    A named person entity should not have any starting titles such as "President", "Prime Minister", "Mayor", "Professor" and "Dr".
    
    (Governmental, political or executive) titles such as "Prime Minister of Australia", "President of Iceland" and "CEO of Google" are not relevant named entities.
    
    General reference to a person or people such as "actress", "chef", "CEO", "woman", "high school student" are not named entities.\nl
}

\txtRule

Here are some example span classifications for your reference. Please follow this format.\nl

Examples:\nl

    
    
    
    
    
    
    
    
    
    
    

1. \tplCrt{Canadian \emphSpan{Prime Minister Justin Trudeau} visits Indigenous communities.}{
    Prime Minister Justin Trudeau}{(B). Wrong Boundary. The correct span boundary is "Justin Trudeau".}

2. \tplCrt{Newly elected \emphSpan{president} pledges to address climate change.}{president}{(D). Not a Named Entity.}

3. \tplCrt{\emphSpan{CEO of Amazon} steps down from role.}{CEO of Amazon}{(D). Not a Named Entity.}

4. \tplCrt{\emphSpan{CEO} of Dubai-based company arrested for fraud.}{CEO}{(D). Not a Named Entity.}

\pptSep

\txtRule

Please classify the following \tplVal{3} spans of text:\nl

    
    
    
    
    

1. \tplCrtS{\emphSpan{Cindy Gruden} appointed as the new CEO of a major automotive company.}{Cindy Gruden}

2. \tplCrtS{\emphSpan{Justin Trudeau} meets with CEOs of major tech companies to discuss investment opportunities in Canada.}{Justin Trudeau}

3. \tplCrtS{\emphSpan{Angela Merkel} visits Silicon Valley to discuss collaboration with automotive tech companies.}{Angela Merkel}}

~\par

\paragraph{Location}

~\par

\prompt{Here are \tplVal{3 sentences from news articles}. Your task is to classify whether the span of text is a named entity of type \tplVal{location}.

In particular, for each query with a span of text enclosed in double braces, please classify the enclosed span of text in context of the query into one of the following categories:

- (A). The span is a named entity of type \tplVal{location}

- (B). The span contains a named \tplVal{location} entity but the span boundary is not precise

- (C). The span is a named entity but the category is not \tplVal{location}

- (D). The span is not a named entity

If the label is (B), you should provide the correct span boundary.

If the label is (C), you should provide the correct entity type. Pick an entity type from the following list:

[\tplVal{person, organization}, other].\nl

\txtRule

{\color{gDarkBlue}
    A named location entity must be the name of a location.
    
    Generic location references such as "city hall", "community center", "downtown", "major cities" and "bank" are not named entities. Categorical location descriptors such as "major airports" and "arts district" are also not named entities.

    Demonyms such as "American", "Chinese" and "Russian" are not relevant named entities. Events like "Olympics" and "Fashion Week" are also not relevant named entities. Hurricanes and other natural disasters are not relevant named entities.\nl
}

\txtRule

Here are some example span classifications for your reference. Please follow this format.\nl

Examples:\nl

    

    
    

    
    

    
    
    
    
    
    

1. \tplCrt{The Prime Minister of \emphSpan{Japan} meets with world leaders to discuss economic cooperation.}{Japan}{(A). Correct Entity Annotation.}

2. \tplCrt{The European Union imposes sanctions on \emphSpan{Russian} officials over human rights abuses.}{Russian}{(C). Wrong Type. The correct entity type is other.}

3. \tplCrt{German Chancellor Angela Merkel meets with \emphSpan{French} President Macron.}{French}{(C). Wrong Type. The correct entity type is other.}

4. \tplCrt{New restaurant to open in \emphSpan{downtown area}, bringing jobs and economic growth.}{downtown area}{(D). Not a Named Entity.}

5. \tplCrt{Celebrities flock to \emphSpan{Paris Fashion Week} for the latest trends.}{Paris Fashion Week}{(B). Wrong Boundary. The correct span boundary is "Paris".}

\pptSep

\txtRule

Please classify the following \tplVal{3} spans of text:\nl

1. \tplCrtS{Gang-related violence on the rise in \emphSpan{New York City}.}{New York City}

2. \tplCrtS{The \emphSpan{Tokyo} Marathon, one of the largest and most prestigious races in the world, has been postponed due to the ongoing pandemic.}{Tokyo}

3. \tplCrtS{\emphSpan{Detroit} police arrest three suspects in connection with a series of robberies in the downtown area.}{Detroit}} 

~\par

\paragraph{Organization} 

~\par

\prompt{Here are \tplVal{3 sentences from news articles}. Your task is to classify whether the span of text is a named entity of type \tplVal{organization}.

In particular, for each query with a span of text enclosed in double braces, please classify the enclosed span of text in context of the query into one of the following categories:

- (A). The span is a named entity of type \tplVal{organization}

- (B). The span contains a named \tplVal{organization} entity but the span boundary is not precise

- (C). The span is a named entity but the category is not \tplVal{organization}

- (D). The span is not a named entity

If the label is (B), you should provide the correct span boundary.

If the label is (C), you should provide the correct entity type. Pick an entity type from the following list:

[\tplVal{person, location}, other].\nl

\txtRule

{\color{gDarkBlue}
    A named organization entity must be the name of an organization.
    
    Generic organization categories such as "high school", "city council", "local community", "government", "company" are not named entities. Descriptive organization types such as "non-profit organization", "foreign government" and "global technology company" are also not named entities.

    Demonyms such as "Chinese", "Russian", "Indian", "Australian" and "European" are not relevant named entities. Viruses such as "COVID-19" are also not relevant named entities. Products such as "iPhone", "Windows" and "Instagram" are not relevant named entities.\nl
}

\txtRule

Here are some example span classifications for your reference. Please follow this format.\nl

Examples:\nl

1. \tplCrt{Indian government bans \emphSpan{TikTok} and 58 other Chinese apps.}{TikTok}{(A). Correct Entity Annotation.}

2. \tplCrt{\emphSpan{CEO} of Tesla Elon Musk Denies Securities Fraud Allegations}{CEO}{(D). Not a Named Entity.}

3. \tplCrt{\emphSpan{Japanese} automaker Toyota recalls millions of vehicles.}{Japanese}{(D). Not a Named Entity.}

4. \tplCrt{\emphSpan{Russian} cosmonauts and Chinese astronauts conduct joint space mission.}{Russian}{(D). Not a Named Entity.}

5. \tplCrt{Pfizer announces new vaccine efficacy data against \emphSpan{COVID-19} variants.}{COVID-19}{(D). Not a Named Entity.}

6. \tplCrt{Local \emphSpan{non-profit organization} provides free meals to homeless veterans.}{non-profit organization}{(D). Not a Named Entity.}

\pptSep

\txtRule

Please classify the following \tplVal{3} spans of text:\nl

1. \tplCrtS{The pharmaceutical company Pfizer announces a breakthrough in the development of a potential \emphSpan{HIV} vaccine.}{HIV}

2. \tplCrtS{Celebrities flock to \emphSpan{Cape Town Fashion Week} for the latest trends.}{Cape Town Fashion Week}

3. \tplCrtS{\emphSpan{The National Gallery of Art} opens new exhibit featuring works by Monet and Renoir.}{The National Gallery of Art}} 

~\par













\begin{table*}[tb]
    \centering 
    \footnotesize
    \resizebox{0.8\textwidth}{!}{
    \begin{tabular}{ l c l}
        \toprule \thead{Variant} & \thead{Group} & \thead{Sample Values} \\

        \midrule
        \multicolumn{3}{l}{\hspace{-0.75em} \emph{CoNLL-2003}} \\
        \midrule
        \multirow{2}{*}[0em]{\makecell{Diversify X}} 
            & News Topic    & World News, Politics, Technology, ...                                        \\
            & Writing Style & Breaking news reports, Investigative journalism, Human interest stories, ... \\

        \addlinespace[0.25em] \hline \addlinespace[0.5em] 
        \multirow{3}{*}[0em]{\makecell{Diversify Y\\(vanilla)}} 
            & per. & Joe Biden, Angela Merkel, Donald Trump, ...                    \\
            & loc. & New York City, Paris, Tokyo, ...                               \\
            & org. & United Nations, World Health Organization, European Union, ... \\

        \addlinespace[0.25em] \hline \addlinespace[0.5em] 
        \multirow{9}{*}[-0.75em]{\makecell{Diversify Y\\(latent)}} 
            & \emph{World News} - per. & Angela Merkel, Elon Musk, Megan Rapinoe, ...                       \\
            & \emph{World News} - loc. & New York City, London, Sydney, ...                                 \\
            & \emph{World News} - org. & United Nations, World Health Organization, Google, ...             \\
            \lightCline{2-3}{0.125em}{0.25em}
            & \emph{Politics} - per.   & Joe Biden, Kamala Harris, Mitch McConnell, ...                     \\
            & \emph{Politics} - loc.   & Washington, D.C., New York City, Beijing, ...                      \\
            & \emph{Politics} - org.   & United Nations, European Union, Democratic National Committee, ... \\
            \lightCline{2-3}{0.125em}{0.25em}
            & \emph{Technology} - per. & Elon Musk, Tim Cook, Sundar Pichai, ...                            \\
            & \emph{Technology} - loc. & Silicon Valley, Shenzhen, China, Seoul, South Korea, ...           \\
            & \emph{Technology} - org. & Google, Apple, Microsoft, ...                                      \\
            \lightCline{2-3}{0.125em}{0.25em}
            & ...                      & ...                                                                \\

        \midrule
        \multicolumn{3}{l}{\hspace{-0.75em} \emph{WikiGold}} \\
        \midrule
        \multirow{2}{*}[0em]{\makecell{Diversify X}} 
            & Topic         & Geography, Sports, Literature, ...     \\
            & Writing Style & Academic, Persuasive, Explanatory, ... \\

        \addlinespace[0.25em] \hline \addlinespace[0.5em] 
        \multirow{3}{*}[0em]{\makecell{Diversify Y\\(vanilla)}} 
            & per. & Barack Obama, Marilyn Monroe, Albert Einstein, ...                          \\
            & loc. & New York City, Paris, Tokyo, ...                                            \\
            & org. & United Nations, International Monetary Fund, World Health Organization, ... \\
            
        \addlinespace[0.25em] \hline \addlinespace[0.5em] 
        \multirow{10}{*}[-0.25em]{\makecell{Diversify Y\\(latent)}} 
            & \emph{Geography} - per.  & Ferdinand Magellan, Alexander von Humboldt, Marie Tharp, ...                          \\
            & \emph{Geography} - loc.  & Mount Everest, Great Barrier Reef, Amazon Rainforest, ...                             \\
            & \emph{Geography} - org.  & National Geographic Society, National Park Service, World Wildlife Fund, ...          \\
            \lightCline{2-3}{0.125em}{0.25em}
            & \emph{Sports} - per.     & Lionel Messi, Serena Williams, Michael Jordan, ...                                    \\
            & \emph{Sports} - loc.     & Madison Square Garden, Camp Nou, Lambeau Field, ...                                   \\
            & \emph{Sports} - org.     & International Olympic Committee, Major League Baseball, National Football League, ... \\
            \lightCline{2-3}{0.125em}{0.25em}
            & \emph{Literature} - per. & William Shakespeare, Jane Austen, F. Scott Fitzgerald, ...                            \\
            & \emph{Literature} - loc. & Stratford-upon-Avon, The Brontë Parsonage Museum, The Shakespeare's Globe, ...        \\
            & \emph{Literature} - org. & Poetry Foundation, PEN America, National Book Foundation, ...                         \\
            \lightCline{2-3}{0.125em}{0.25em}
            & ...                      & ...                                                                                   \\
        
        \bottomrule
    \end{tabular}
    }
    \caption{Sample generated values for Diversity Requirement Configurations for CoNLL-2003 and WikiGold. } 
    \label{tbl:diverse-config-sample-values-general-domain}
\end{table*}

\begin{table*}[tb]
    \centering 
    \footnotesize
    \resizebox{0.93\textwidth}{!}{
    \begin{tabular}{ l c l}
        \toprule \thead{Variant} & \thead{Group} & \thead{Sample Values} \\

        \midrule
        \multicolumn{3}{l}{\hspace{-0.75em} \emph{MIT-Movie}} \\
        \midrule
        \multirow{4}{*}[0em]{\makecell{Diversify X}} 
            & Query Category   & Plot summary and synopsis, Cast and crew information, Movie reviews and ratings, ... \\
            & User Demographic & Teenage girl, Middle-aged man, Senior woman, ...                                     \\
            & User Emotion     & Excited, Bored, Indifferent, ...                                                     \\
            & Query Language   & Vague, Formal, Indirect, ...                                                         \\
            
        \addlinespace[0.25em] \hline \addlinespace[0.5em] 
        \multirow{12}{*}[0em]{\makecell{Diversify Y\\(vanilla)}} 
            & Title      & The Godfather, Citizen Kane, Pulp Fiction, ...            \\
            & View. Rat. & 5 stars, 4.5 stars, A+, ...                               \\
            & Year       & 1975, 2003, 1999, ...                                     \\
            & Genre      & Action, Romance, Comedy, ...                              \\
            & Director   & Steven Spielberg, Martin Scorsese, Christopher Nolan, ... \\
            & MPAA Rat.  & G, PG, PG-13, ...                                         \\
            & Plot       & Love triangle, Time travel, Revenge, ...                  \\
            & Actor      & Leonardo DiCaprio, Meryl Streep, Tom Hanks, ...           \\
            & Trailer    & Teaser, Sneak peek, Preview, ...                          \\
            & Song       & Bohemian Rhapsody, Purple Haze, Thriller, ...             \\
            & Review     & Gripping, Riveting, Spectacular, ...                      \\
            & Char.      & Harry Potter, Tony Stark, Darth Vader, ...                \\

        \addlinespace[0.25em] \hline \addlinespace[0.5em] 
        \multirow{37}{*}[-0.25em]{\makecell{Diversify Y\\(latent)}} 
            & \emph{Plot summary and synopsis} - Title      & The Lord of the Rings, The Shawshank Redemption, The Matrix, ...         \\
            & \emph{Plot summary and synopsis} - View. Rat. & Gripping, Must see, Riveting, ...                                        \\
            & \emph{Plot summary and synopsis} - Year       & 1980, 1995, 2001, ...                                                    \\
            & \emph{Plot summary and synopsis} - Genre      & Action, Adventure, Animation, ...                                        \\
            & \emph{Plot summary and synopsis} - Director   & Steven Spielberg, Quentin Tarantino, Christopher Nolan, ...              \\
            & \emph{Plot summary and synopsis} - MPAA Rat.  & PG-13, R, G, ...                                                         \\
            & \emph{Plot summary and synopsis} - Plot       & Love triangle, Murder mystery, Time travel, ...                          \\
            & \emph{Plot summary and synopsis} - Actor      & Tom Hanks, Sandra Bullock, Leonardo DiCaprio, ...                        \\
            & \emph{Plot summary and synopsis} - Trailer    & Teaser, Sneak peek, Preview, ...                                         \\
            & \emph{Plot summary and synopsis} - Song       & Purple Rain, Bohemian Rhapsody, Don't Want to Miss a Thing, ...          \\
            & \emph{Plot summary and synopsis} - Review     & Gripping storyline, Intriguing plot, Compelling narrative, ...           \\
            & \emph{Plot summary and synopsis} - Char.      & Harry Potter, Hermione Granger, Ron Weasley, ...                         \\
            
            \lightCline{2-3}{0.125em}{0.25em}
            & \emph{Cast and crew information} - Title      & The Godfather, Pulp Fiction, Titanic, ...                                \\
            & \emph{Cast and crew information} - View. Rat. & Must-see, Excellent, Outstanding, ...                                    \\
            & \emph{Cast and crew information} - Year       & 1995, 2000s, 1987, ...                                                   \\
            & \emph{Cast and crew information} - Genre      & Action, Adventure, Animation, ...                                        \\
            & \emph{Cast and crew information} - Director   & Steven Spielberg, Martin Scorsese, Quentin Tarantino, ...                \\
            & \emph{Cast and crew information} - MPAA Rat.  & PG-13, R, G, ...                                                         \\
            & \emph{Cast and crew information} - Plot       & Time travel, Artificial intelligence, Bank heist, ...                    \\
            & \emph{Cast and crew information} - Actor      & Tom Hanks, Leonardo DiCaprio, Meryl Streep, ...                          \\
            & \emph{Cast and crew information} - Trailer    & Teaser, Sneak peek, Preview, ...                                         \\
            & \emph{Cast and crew information} - Song       & Beyoncé, Elton John, Frank Sinatra, ...                                  \\
            & \emph{Cast and crew information} - Review     & Stellar performances, Chemistry between the actors, Supportive cast, ... \\
            & \emph{Cast and crew information} - Char.      & Luke Skywalker, Harry Potter, Tony Stark, ...                            \\

            \lightCline{2-3}{0.125em}{0.25em}
            & \emph{Movie reviews and ratings} - Title      & The Godfather, Avatar, The Shawshank Redemption, ...                     \\
            & \emph{Movie reviews and ratings} - View. Rat. & Must-see, Highly recommended, Excellent, ...                             \\
            & \emph{Movie reviews and ratings} - Year       & 2020, 1995, 1982, ...                                                    \\
            & \emph{Movie reviews and ratings} - Genre      & Action, Adventure, Comedy, ...                                           \\
            & \emph{Movie reviews and ratings} - Director   & Steven Spielberg, Martin Scorsese, Quentin Tarantino, ...                \\
            & \emph{Movie reviews and ratings} - MPAA Rat.  & G, PG, PG-13, ...                                                        \\
            & \emph{Movie reviews and ratings} - Plot       & Serial killer, Time travel, Alien invasion, ...                          \\
            & \emph{Movie reviews and ratings} - Actor      & Tom Hanks, Meryl Streep, Leonardo DiCaprio, ...                          \\
            & \emph{Movie reviews and ratings} - Trailer    & Teaser, Preview, Sneak peek, ...                                         \\
            & \emph{Movie reviews and ratings} - Song       & Moon River, My Heart Will Go On, Let It Go, ...                          \\
            & \emph{Movie reviews and ratings} - Review     & Gripping, Captivating, Stunning, ...                                     \\
            & \emph{Movie reviews and ratings} - Char.      & Luke Skywalker, Harry Potter, James Bond, ...                            \\
            
            \lightCline{2-3}{0.125em}{0.25em}
            & ...                                           & ...                                                                      \\

        \bottomrule
    \end{tabular}
    }
    \caption{Sample generated values for Diversity Requirement Configurations for MIT-Movie. } 
    \label{tbl:diverse-config-sample-values-mit-movie}
\end{table*}

\begin{table*}[tb]
    \centering 
    \footnotesize
    \resizebox{0.9\textwidth}{!}{
    \begin{tabular}{ l c l}
        \toprule \thead{Variant} & \thead{Group} & \thead{Sample Values} \\

        \midrule
        \multicolumn{3}{l}{\hspace{-0.75em} \emph{MIT-Restaurant}} \\
        \midrule
        \multirow{8}{*}[0em]{\makecell{Diversify X}} 
            & Meal Category       & fine dining, buffet, late-night dining, ...                                              \\
            & User Demographic    & 25-year-old female student, 40-year-old female accountant, 55-year-old male retiree, ... \\
            & Ambiance            & Casual, Outdoor seating, Family-friendly, ...                                            \\
            & Price Range         & Affordable, Budget-friendly, Mid-range, ...                                              \\
            & Dietary Restriction & Vegetarian, Vegan, Gluten-free, ...                                                      \\
            & Special Offering   & Chef's specials, Private Dining, Tasting Menus, ...                                      \\
            & Service Mode        & takeout, delivery, dine-in, ...                                                          \\

        \addlinespace[0.25em] \hline \addlinespace[0.5em] 
        \multirow{8}{*}[0em]{\makecell{Diversify Y\\(vanilla)}} 
            & Rest. Name & La Cucina Italiana, The Steakhouse, Sushi Park, ...     \\
            & Amenity    & WiFi, Outdoor seating, Parking, ...                     \\
            & Cuisine    & Italian, Thai, Chinese, ...                             \\
            & Dish       & Spaghetti carbonara, Pad Thai, Sushi, ...               \\
            & Hours      & Opening hours, Closing time, Operating hours, ...       \\
            & Location   & Near me, Close by, Nearby, ...                          \\
            & Price      & McDonald's, Ruth's Chris Steak House, Olive Garden, ... \\
            & Rating     & Yelp, Google reviews, TripAdvisor, ...                  \\

        \addlinespace[0.25em] \hline \addlinespace[0.5em] 
        \multirow{25}{*}[0em]{\makecell{Diversify Y\\(latent)}} 
            & \emph{fine dining} - Rest. Name       & The French Laundry, Per Se, Alinea, ...                                            \\
            & \emph{fine dining} - Amenity          & Wine cellar, Private dining room, Outdoor terrace, ...                             \\
            & \emph{fine dining} - Cuisine          & Italian, French, Japanese, ...                                                     \\
            & \emph{fine dining} - Dish             & Filet mignon, Lobster bisque, Chicken parmesan, ...                                \\
            & \emph{fine dining} - Hours            & Lunchtime, Dinner, Brunch, ...                                                     \\
            & \emph{fine dining} - Location         & Near me, City, Downtown, ...                                                       \\
            & \emph{fine dining} - Price            & Le Bernardin, Per Se, Eleven Madison Park, ...                                     \\
            & \emph{fine dining} - Rating           & Best, Top, Highest, ...                                                            \\

            \lightCline{2-3}{0.125em}{0.25em}
            & \emph{buffet} - Rest. Name            & Chez Panisse, The French Laundry, Nobu, ...                                        \\
            & \emph{buffet} - Amenity               & WiFi availability, Parking options, Outdoor seating, ...                           \\
            & \emph{buffet} - Cuisine               & Indian, Italian, Chinese, ...                                                      \\
            & \emph{buffet} - Dish                  & Sushi, Pizza, Tacos, ...                                                           \\
            & \emph{buffet} - Hours                 & Sunday Brunch, Breakfast Buffet, Lunch Buffet, ...                                 \\
            & \emph{buffet} - Location              & Nearby, Close, Local, ...                                                          \\
            & \emph{buffet} - Price                 & Budget-friendly, Affordable, Expensive, ...                                        \\
            & \emph{buffet} - Rating                & Best buffet in town, Top-rated buffet restaurants, Highly recommended buffets, ... \\

            \lightCline{2-3}{0.125em}{0.25em}
            & \emph{late-night dining} - Rest. Name & The Midnight Diner, After Hours Bistro, The Night Owl Tavern, ...                  \\
            & \emph{late-night dining} - Amenity    & Outdoor seating, Private dining, Bar service, ...                                  \\
            & \emph{late-night dining} - Cuisine    & Italian, Mexican, Chinese, ...                                                     \\
            & \emph{late-night dining} - Dish       & Pizza, Sushi, Burger, ...                                                          \\
            & \emph{late-night dining} - Hours      & Late-night dining, Open late, Closing time, ...                                    \\
            & \emph{late-night dining} - Location   & Near me, Close by, Nearby, ...                                                     \\
            & \emph{late-night dining} - Price      & Budget-friendly, Affordable, Cheap eats, ...                                       \\
            & \emph{late-night dining} - Rating     & Top rated, Highly recommended, 5-star, ...                                         \\

            \lightCline{2-3}{0.125em}{0.25em}
            & ...                                   & ...                                                                                \\

        \bottomrule
    \end{tabular}
    }
    \caption{Sample generated values for Diversity Requirement Configurations for MIT-Restaurant. } 
    \label{tbl:diverse-config-sample-values-mit-restaurant}
\end{table*}

\subsection{Diversity Requirement Sample Values}
\label{apd:diversify-config-values}
\normalsize
In this section, we show a few of the generated diversity requirement configuration values for illustrative purposes 
in Tables \ref{tbl:diverse-config-sample-values-general-domain}, \ref{tbl:diverse-config-sample-values-mit-movie} and \ref{tbl:diverse-config-sample-values-mit-restaurant}.

\subsection{Generated NER Samples}
\label{apd:generated-ner-samples}
\normalsize
In this section, we show the selected demo samples and 
a few generated NER samples from Simple Prompt and each diversity variant. 
We show the raw LLM-generated samples, which may not follow the diversity instructions and demo formatting. 
Note that the generated samples may contain wrong annotations. 
For diversity approaches, we additionally show the corresponding sampled configurations in the prompts. 

\subsubsection{Demo Samples}
\paragraph{CoNLL-2003}
\footnotesize
~\par

\prompt{1. \tplNER{BASEBALL - DODGERS WIN FIFTH STRAIGHT.}{DODGERS (organization)}
    
2. \tplNER{With 2.5 percent gross domestic product growth expected for 1996, new job growth should slowly lower the unemployment rate over the rest of the year.}{}

3. \tplNER{A.de Silva not out 49}{A.de Silva (person)}

4. \tplNER{Saudi Arabia executes Pakistani man.}{Saudi Arabia (location)}}

~\par

\paragraph{WikiGold}

~\par

\prompt{1. Sentence: `\tplVal{His book, A Biblical Case for an Old Earth was described in a review by Law Professor David W. Opderbeck, in the American Scientific Affiliation's Perspectives on Science and Christian Faith as "succeed[ing] admirably" in "establish[ing] that the `day-age' view is a valid alternative for Christians who hold to biblical inerrancy", but as "less persuasive" at "argu[ing] for a concordist understanding of the Genesis texts and modern science."}'

Named Entities: [\tplVal{David W. Opderbeck (person), American Scientific Affiliation (organization)}]

2. \tplNER{He scored one international goal, against Slovakia in 1994.}{Slovakia (location)}

3. \tplNER{It was built as a self-contained steamship dock facility.}{}}

~\par

\paragraph{MIT-Movie}

~\par

\prompt{1. \tplNERq{what type of movie genre is the perfect weapon}{the perfect weapon (Title)}

2. \tplNERq{how many movies came out in 2004}{2004 (Year)}

3. \tplNERq{could you show me some part of the new indiana jones movie}{some part (Trailer), indiana jones (Character)}

4. \tplNERq{when did mark joffe direct the bounty hunter film that is rated pg}{mark joffe (Director), bounty hunter (Plot), pg (MPAA Rating)}

5. \tplNERq{is cary grant in any historical films that are a must see}{cary grant (Actor), historical (Genre), must see (Viewers' Rating)}

6. \tplNERq{show me a movie with the song a whole new world}{a whole new world (Song)}

7. \tplNERq{what movie is considered the funniest of all time}{funniest of all time (Review)}}

~\par

\paragraph{MIT-Restaurant}

~\par

\prompt{1. \tplNERq{is there any place around here that has a good menu for happy hour}{good (Rating), happy hour (Amenity)}

2. \tplNERq{where can i get afghan cuisine before 9 a m}{afghan (Cuisine), before 9 a m (Hours)}

3. \tplNERq{call mcdonalds}{mcdonalds (Restaurant Name)}

4. \tplNERq{find us a place to eat for under 10 dollars a plate}{under 10 dollars a plate (Price)}

5. \tplNERq{im thirsty take me to the closest slurpee}{closest (Location), slurpee (Dish)}}

~\par

\subsubsection{Simple Prompt}
\paragraph{CoNLL-2003}

~\par

\prompt{1. The European Union is considering imposing tariffs on American products.

Named Entities: [European Union (organization), American (location)]\nl

2. Elon Musk's SpaceX successfully launches another batch of Starlink satellites.

Named Entities: [Elon Musk (person), SpaceX (organization)]\nl

3. The protests in Hong Kong continue to escalate as tensions rise between the government and demonstrators.

Named Entities: [Hong Kong (location)]\nl

...}

~\par

\paragraph{WikiGold}

~\par

\prompt{1. The Eiffel Tower is a wrought iron lattice tower on the Champ de Mars in Paris, France.

Named Entities: [Eiffel Tower (location), Champ de Mars (location), Paris (location), France (location)]

2. John F. Kennedy International Airport is a major international airport in Queens, New York City.

Named Entities: [John F. Kennedy International Airport (location), Queens (location), New York City (location)]

3. The World Health Organization is a specialized agency of the United Nations responsible for international public health.

Named Entities: [World Health Organization (organization), United Nations (organization)]

...}

~\par

\paragraph{MIT-Movie}

~\par

\prompt{1. Can you tell me the plot of the movie Inception?

\quad Named Entities: [Inception (Title)]\nl

2. What is the viewers' rating for The Shawshank Redemption?

\quad Named Entities: [The Shawshank Redemption (Title), viewers' rating (Viewers' Rating)]\nl

3. Show me movies from the 1980s.

\quad Named Entities: [1980s (Year)]\nl

...}

~\par

\paragraph{MIT-Restaurant}

~\par

\prompt{\footnotesize



% ...
1. "Can you recommend a good sushi restaurant nearby?"

Named Entities: [sushi (Cuisine), nearby (Location)]\nl

2. "What Italian restaurant has the best reviews?"

Named Entities: [Italian (Cuisine), best (Rating)]\nl

3. "I want to go to a steakhouse for dinner tonight."

Named Entities: [steakhouse (Cuisine), tonight (Hours)]\nl

...}

~\par

\subsubsection{Diversify X}
\paragraph{CoNLL-2003}

~\par 

\promptNResponse{Additionally, the generated \tplVal{news article sentences} should follow the requirements below:

1. The news should be about \tplVal{Education};

2. The writing style of the news should be \tplVal{human interest stories}.}{1. "Local high school student overcomes homelessness to graduate with honors."

Named Entities: [high school student (person)]\nl

2. "Non-profit organization partners with school district to provide free tutoring services for low-income students."

Named Entities: [non-profit organization (organization), school district (organization)]\nl

3. "Former teacher starts scholarship fund to help underprivileged students pursue higher education."

Named Entities: [teacher (person), scholarship fund (organization)]\nl}

~\par

\paragraph{WikiGold}

~\par 

\promptNResponse{Additionally, the generated \tplVal{Wikipedia article sentences} should follow the requirements below:

1. The topic of the Wikipedia article sentences should be \tplVal{Mathematics};

2. The writing style of the Wikipedia article sentences should be \tplVal{biographical}.}{1. In 1931, mathematician Kurt Gödel published his famous incompleteness theorems, which revolutionized the study of mathematical logic.

Named Entities: [Kurt Gödel (person)]\nl

2. As a pioneer in the field of algebraic geometry, David Mumford was awarded the Fields Medal in 1974 for his groundbreaking work on geometric invariant theory.

Named Entities: [David Mumford (person)]\nl

3. The Clay Mathematics Institute offers a prize of one million dollars for a correct proof of the Riemann hypothesis, one of the most famous unsolved problems in mathematics.

Named Entities: [Clay Mathematics Institute (organization), Riemann (person)]\nl}

~\par

\paragraph{MIT-Movie}

~\par 

\promptNResponse{Additionally, the generated \tplVal{spoken movie queries} should follow the requirements below:

1. The query should inquire about \tplVal{Movie showtimes and ticket availability};

2. The user should be \tplVal{adventurous}.}{1. "Can I buy tickets for the latest James Bond movie?"

Named Entities: [latest James Bond (Title)]\nl

2. "Is there a sci-fi movie with a soundtrack by Hans Zimmer playing at the theater today?"

Named Entities: [sci-fi (Genre), Hans Zimmer (Song)]\nl

3. "I'm looking for a movie directed by Christopher Nolan and rated R. Do you have any suggestions?"

Named Entities: [Christopher Nolan (Director), R (MPAA Rating)]\nl}

~\par

\paragraph{MIT-Restaurant}

~\par 



\promptNResponse{Additionally, the generated \tplVal{restaurant queries} should follow the requirements below:

1. The meal category should be \tplVal{fine dining};

2. The restaurant's price range should be \tplVal{affordable}.}{Query: "{can you recommend a fine dining restaurant with affordable prices}"

Named Entities: [fine dining (Amenity), affordable prices (Price)]\nl

Query: "what are the top-rated fine dining restaurants in the area"

Named Entities: [top-rated (Rating), fine dining (Amenity)]\nl

Query: "where can I find a fine dining restaurant with budget-friendly options"

Named Entities: [fine dining (Amenity), budget-friendly options (Price)]\nl}

~\par

\subsubsection{Diversify Y (vanilla)}
\paragraph{CoNLL-2003}

~\par 

\promptNResponse{Additionally, in the generated \tplVal{sentences}, include the following terms: [\tplVal{United Nations Development Programme, Steven Spielberg}].}{1. "Steven Spielberg to direct new film about the United Nations Development Programme."

Named Entities: [Steven Spielberg (person), United Nations Development Programme (organization)]\nl

2. "United Nations Development Programme launches new initiative to combat climate change in developing countries."

Named Entities: [United Nations Development Programme (organization)]\nl

3. "Steven Spielberg's latest movie receives funding from the United Nations Development Programme."

Named Entities: [Steven Spielberg (person), United Nations Development Programme (organization)]\nl}

~\par

\paragraph{WikiGold}

~\par 

\promptNResponse{Additionally, in the generated \tplVal{sentences}, include the following terms: [\tplVal{Alexander the Great, Angela Merkel}].}{1. "Alexander the Great is often considered one of the most successful military commanders in history for the remarkable achievements of his campaigns."

Named Entities: [Alexander the Great (person)]\nl

2. "Angela Merkel, the former Chancellor of Germany, is known for her leadership during the European financial crisis and her role in the European Union."

Named Entities: [Angela Merkel (person), Germany (location), European Union (organization)]\nl

3. "The Louvre, located in Paris, is the world's largest art museum and a historic monument in France."

Named Entities: [Louvre (location), Paris (location), France (location)]\nl}

~\par

\paragraph{MIT-Movie}

~\par 

\promptNResponse{Additionally, in the generated \tplVal{queries}, include the following terms: [\tplVal{Zooey Deschanel, Cyberpunk, TV-Y, Take Me Out to the Ball Game, 1968}].}{1. "Can you tell me if Zooey Deschanel is starring in any new comedy films"

Named Entities: [Zooey Deschanel (Actor), comedy (Genre)]\nl

2. "I'm looking for a cyberpunk movie released in 1968"

Named Entities: [cyberpunk (Genre), 1968 (Year)]\nl

3. "Is there a family-friendly movie with a TV-Y rating featuring the song 'Take Me Out to the Ball Game'"

Named Entities: [family-friendly (Genre), TV-Y (MPAA Rating), Take Me Out to the Ball Game (Song)]\nl}

~\par

\paragraph{MIT-Restaurant}

~\par 

\promptNResponse{Additionally, in the generated \tplVal{queries}, include the following terms: [\tplVal{Value meal, The Firehouse Grill}].}{1. "I'm hungry, can you recommend a restaurant with a good value meal?"

Named Entities: [good (Rating), Value meal (Amenity)]\nl

2. "What's the address and hours of operation for The Firehouse Grill?"

Named Entities: [The Firehouse Grill (Restaurant Name), address (Location), hours of operation (Hours)]\nl

3. "I'm craving Italian cuisine, do you know a restaurant nearby that serves it?"

Named Entities: [Italian (Cuisine), nearby (Location)]\nl}

~\par

\subsubsection{Diversify Y (latent)}
\label{apd:diverse-y-latent-samples}


\paragraph{CoNLL-2003}

~\par 

\promptNResponse{Additionally, in the generated \tplVal{sentences}, include the following terms: [\tplVal{Pandora Media, Sony Music Entertainment, Sony Music}].}{1. "Pandora Media announces partnership with Sony Music Entertainment."

Named Entities: [Pandora Media (organization), Sony Music Entertainment (organization)]\nl

2. "Sony Music to release new album in collaboration with a popular artist."

Named Entities: [Sony Music (organization)]\nl

3. "Pandora Media's CEO discusses future plans for the company at annual conference."

Named Entities: [Pandora Media (organization)]\nl}

~\par

\paragraph{WikiGold}

~\par 

\promptNResponse{Additionally, in the generated \tplVal{sentences}, include the following terms: [\tplVal{Doctors Without Borders, Centers for Disease Control and Prevention}].}{1. Doctors Without Borders provides medical aid to over 70 countries around the world.

Named Entities: [Doctors Without Borders (organization)]\nl

2. The Centers for Disease Control and Prevention is responsible for monitoring and preventing the spread of infectious diseases in the United States.

Named Entities: [Centers for Disease Control and Prevention (organization)]\nl

3. In 2018, the organization partnered with the local government to provide healthcare services in rural areas.

Named Entities: []\nl}

~\par

\paragraph{MIT-Movie}

~\par 

\promptNResponse{Additionally, in the generated \tplVal{queries}, include the following terms: [\tplVal{Taylor Swift, X, Jessica Chastain, 2011, Patty Jenkins, Sam Mendes, Standout performances}].}{1. "What 2011 movie directed by Patty Jenkins received standout performances by Jessica Chastain?"

\quad Named Entities: [2011 (Year), Patty Jenkins (Director), Jessica Chastain (Actor), Standout performances]\nl

2. "Could you recommend a film with a Taylor Swift song in the trailer?"

\quad Named Entities: [Taylor Swift (Song), Trailer]\nl

3. "Show me a Sam Mendes movie with a suspenseful plot."

\quad Named Entities: [Sam Mendes (Director), suspenseful (Plot)]\nl}

~\par

\paragraph{MIT-Restaurant}

~\par 

\promptNResponse{Additionally, in the generated \tplVal{queries}, include the following terms: [\tplVal{Cheeseburger, Mexican, Discounted items}].}{1. "Can you recommend a place that serves a good cheeseburger?"

Named Entities: [cheeseburger (Dish)]\nl

2. "I'm craving Mexican food, where can I find a Mexican restaurant nearby?"

Named Entities: [Mexican (Cuisine), nearby (Location)]\nl

3. "Do you know of any restaurants that offer discounted items during happy hour?"

Named Entities: [discounted items (Amenity), happy hour (Amenity)]\nl}

~\par

\subsubsection{Diversify X+Y}
\paragraph{CoNLL-2003}

~\par 

\promptNResponse{Additionally, the generated \tplVal{news article sentences} should follow the requirements below:

1. The generated news article sentences should be about \tplVal{Art};

2. The generated news article sentences should include the following terms: [\tplVal{Colosseum, Rome, Public Art Fund, Art Institute of Chicago}].}{1. The Colosseum in Rome to host a new public art exhibition.

Named Entities: [Colosseum (location), Rome (location), Public Art Fund (organization)]\nl

2. Art Institute of Chicago unveils new digital art collection.

Named Entities: [Art Institute of Chicago (organization)]\nl

3. Renowned artist to create a new mural for the Public Art Fund in Rome.

Named Entities: [Public Art Fund (organization), Rome (location)]\nl}

~\par

\paragraph{WikiGold}

~\par 

\promptNResponse{Additionally, the generated \tplVal{Wikipedia article sentences} should follow the requirements below:

1. The topic of the Wikipedia article sentences should be \tplVal{Music};

2. The generated Wikipedia article sentences should include the following terms: [\tplVal{Bob Dylan, Rock and Roll Hall of Fame, Berklee College of Music}].}{1. Her rendition of "Like a Rolling Stone" by Bob Dylan earned her a standing ovation from the audience.

Named Entities: [Bob Dylan (person)]\nl

2. The band was inducted into the Rock and Roll Hall of Fame in 2006, in recognition of their influence on the music industry.

Named Entities: [Rock and Roll Hall of Fame (organization)]\nl

3. Many successful musicians and artists, including Quincy Jones and Melissa Etheridge, have attended the prestigious Berklee College of Music.

Named Entities: [Quincy Jones (person), Melissa Etheridge (person), Berklee College of Music (organization)]\nl}

~\par

\paragraph{MIT-Movie}

~\par 

\promptNResponse{Additionally, the generated \tplVal{spoken movie queries} should follow the requirements below:

1. The query should inquire about \tplVal{Soundtrack and music information};

2. The user should be \tplVal{ambivalent};

3. The generated spoken movie queries should include the following terms: [\tplVal{Thrilling musical arrangements, EDM, Rock and roll legend, Captivating theme music, I Wanna Dance with Somebody, Ska}].}{1. "I'm looking for some thrilling musical arrangements in a movie. Can you recommend anything with EDM or rock and roll legend in the soundtrack?"

Named Entities: [thrilling musical arrangements (Song)]\nl
 
2. "I'm in the mood for some captivating theme music. Do you know any movies featuring 'I Wanna Dance with Somebody' or Ska?"

Named Entities: [captivating theme music (Song), I Wanna Dance with Somebody (Song), Ska (Genre)]\nl

3. "I'm not sure what I'm in the mood for. Can you show me a movie with an awesome EDM soundtrack?"

Named Entities: [EDM (Song)]\nl}

~\par

\paragraph{MIT-Restaurant}

~\par 

\promptNResponse{Additionally, the generated \tplVal{restaurant queries} should follow the requirements below:

1. The meal category should be \tplVal{fine dining};

2. The restaurant's price range should be \tplVal{affordable};

3. The generated restaurant queries should include the following terms: [\tplVal{Near me, Valet parking, Blue Hill at Stone Barns, High-end}].}{Query: "Can you recommend a fine dining restaurant near me with valet parking?"

Named Entities: [fine dining (Amenity), near me (Location), valet parking (Amenity)]\nl

Query: "I'm looking for an affordable high-end restaurant with valet parking, maybe something like Blue Hill at Stone Barns"

Named Entities: [affordable (Price), high-end (Price), valet parking (Amenity), Blue Hill at Stone Barns (Restaurant Name)]\nl

Query: "Where can I find a fine dining restaurant around here with valet parking and a price range that's affordable?"

Named Entities: [fine dining (Amenity), around here (Location), valet parking (Amenity), affordable (Price)]\nl}

~\par

\end{document}